\documentclass{article}

\PassOptionsToPackage{round}{natbib}
\usepackage[final]{neurips_2022}

\usepackage[utf8]{inputenc} 
\usepackage[T1]{fontenc}    
\usepackage{hyperref}       
\usepackage{url}            
\usepackage{booktabs}       
\usepackage{amsfonts}       
\usepackage{nicefrac}       
\usepackage{microtype}      
\usepackage{xcolor}         
\usepackage{todonotes}
\usepackage{amsmath}
\usepackage{caption}
\usepackage{subcaption}
\usepackage{wrapfig}
\usepackage{dsfont}
\usepackage{nccmath}
\usepackage[ruled,vlined]{algorithm2e}
\usepackage{multirow}
\usepackage{graphicx}
\usepackage{listings}
\title{Towards Trustworthy Automatic Diagnosis Systems by Emulating Doctors' Reasoning with Deep Reinforcement Learning}
\author{
    Arsène Fansi Tchango $^{\dagger}$\\
    \texttt{arsene.fansi.tchango@mila.quebec}\\
    \And
    Rishab Goel $^{\ddagger}$\\
    \texttt{rgoel0112@gmail.com}\\
    \And
    Julien Martel $^{\S}$ \\
    \texttt{julien@datadoc.ca} \\
    \AND
    Zhi Wen $^{\dagger}$  \qquad Gaétan Marceau Caron $^{\dagger}$ \qquad Joumana Ghosn $^{\dagger}$ \\
    \texttt{\{zhi.wen,gaetan.marceau.caron,joumana.ghosn\}@mila.quebec}\\
    $^{\dagger}$ Mila-Quebec AI Institute\\
    $^{\ddagger}$ Work done while at Mila-Quebec AI Institute \\
    $^{\S}$ Work done while at Dialogue Health Technologies Inc.\\
}

\newcommand{\CasandeDDx}{{DDXPlus} }

\begin{document}

\maketitle

\begin{abstract}

The automation of the \textit{medical evidence acquisition} and \textit{diagnosis process} has recently attracted increasing attention in order to reduce the workload of doctors and democratize access to medical care. However, most works proposed in the machine learning literature focus solely on improving the prediction accuracy of a patient's pathology. We argue that this objective is insufficient to ensure doctors' acceptability of such systems. In their initial interaction with patients, doctors do not only focus on identifying the pathology a patient is suffering from; they instead generate a differential diagnosis (in the form of a short list of plausible diseases) because the medical evidence collected from patients is often insufficient to establish a final diagnosis. Moreover, doctors explicitly explore severe pathologies before potentially ruling them out from the differential, especially in acute care settings. Finally, for doctors to trust a system's recommendations, they need to understand how the gathered evidences led to the predicted diseases. In particular, interactions between a system and a patient need to emulate the reasoning of doctors. We therefore propose to model the evidence acquisition and automatic diagnosis tasks using a deep reinforcement learning framework that considers three essential aspects of a doctor's reasoning, namely generating a \textit{differential diagnosis} using an \textit{exploration-confirmation} approach while prioritizing \textit{severe pathologies}.
We propose metrics for evaluating interaction quality based on these three aspects. We show that our approach performs better than existing models while maintaining competitive pathology prediction accuracy.
\end{abstract}

\section{Introduction}\label{s:introduction}

In recent years, the digital healthcare industry has grown rapidly, benefiting from advances in machine learning \citep{esteva2019guide,xiao2018opportunities}. In particular, telemedicine, i.e., healthcare services provided via digital means, has received much attention \citep{kichloo2020telemedicine}. Aiming to reduce the workload of human doctors and thereby broaden access to care by automating parts of the interaction with patients, telemedicine applications typically involve two significant components, among others: \textit{evidence acquisition} and \textit{automatic diagnosis}. In a typical interaction, a patient first presents his/her chief complaint\footnote{``A chief complaint is a concise statement in English or other natural language of the symptoms that caused a patient to seek medical care.'' \href{https://www.ncbi.nlm.nih.gov/pmc/articles/PMC7161385/}{https://www.ncbi.nlm.nih.gov/pmc/articles/PMC7161385/}} to the telemedicine system, then the system asks further questions to gather more evidences (i.e., information about additional symptoms the patient might be experiencing and antecedents / risk factors the patient might have), and finally makes a prediction regarding the underlying diseases based on all collected evidences. Importantly, a human doctor would typically review the entire interaction, including the collected evidences and the predicted diseases, before establishing a diagnosis and deciding on next steps (e.g., ordering additional tests, preparing a prescription, etc.).

There are many existing works in the machine learning literature that aim to improve the automated evidence acquisition and disease diagnosis steps, using Reinforcement Learning (RL) \citep{kao2018context,wei2018task, yuan2021efficient}, Bayesian networks \citep{guan2021bayesian, liu2022my}, or supervised approaches \citep{Luo2020KnowledgeGC, chen2021diaformer}. These methods are generally trained to collect relevant evidences and to predict the pathology the patient is suffering from, while minimizing the number of questions asked to the patient.

However, these works overlook several crucial aspects of medical history taking, which is the process used by a doctor to interact with a patient with the goal of building the patient's medical history. Medical history is a broad concept encompassing various kinds of information relevant to a patient's current concerns \citep{nichol2018medical}. It includes past illnesses, current symptoms, demographics, etc. In taking medical history, a doctor organizes the questions asked to a patient under an \textit{exploration-confirmation} framework. At any point during the interaction, the doctor has in their mind a \textit{differential diagnosis}, corresponding to a list of diseases worth considering, which might be modified throughout the interaction, based on the answers provided  by the patient. In forming the differential diagnosis, the doctor treats \textit{severe pathologies} differently by prioritizing questions that could help ruling them out even if they are less likely. In a usual in-person clinical setting, a physical examination will then be conducted to seek specific physical signs and improve the sensitivity and specificity of the medical history. A visual examination and/or supervised self-examination can also be conducted in a virtual care setting. After the medical history and physical examination are completed, the physician will decide if additional tests are necessary to establish the final diagnosis. Finally, a care plan will be established with the patient and proper follow-up, if needed. We next provide more details about the main concepts.

\paragraph{Differential diagnosis}

During the interaction with a patient, a medical doctor considers a set of plausible diseases, known as the \textit{differential diagnosis} or simply \textit{differential}, which is refined throughout the interaction based on the information provided by the patient \citep{henderson2012patient,guyatt2002users,rhoads2017formulating}. The final differential often contains several diseases because the patient's symptoms and antecedents are insufficient to pinpoint a single pathology. As a result, the doctor might order follow up exams such as imagery and blood works to collect additional information to refine the differential and identify the pathology the patient is suffering from.
Prior work on automatic diagnosis primarily focused on predicting the ground truth pathology, i.e., the disease causing the patient's symptoms \citep{chen2021diaformer,zhao2021weighted,yuan2021efficient,guan2021bayesian,wei2018task,xu2019end,kao2018context}.  \citet{richens2020improving} considers the differential in their approach; however it is unclear whether all evidences are provided to the model at the same time or if the model is part of the acquisition process. Our approach considers the differential diagnosis as an essential part of the model. Considering differentials instead of a single pathology has the added benefit of accounting for uncertainty and errors inherent in the diagnosis, especially when decisions are made solely based on interactions with patients and without medical exams or tests. For this reason, being able to predict the differential rather than the ground truth pathology is an important part of gaining the trust of doctors in a model.

\paragraph{Exploration-confirmation}

In addition to considering differentials, inquiring about evidences in a manner similar to the way a doctor would engage with patients is another important factor in gaining the trust of doctors in automated systems, and this is largely overlooked in prior works as well.  
Doctors generally proceed according to a diagnostic framework that provides an organized and logical approach for building a differential. More specifically, the interactions conducted by a doctor generally consist of two distinct phases: \textit{exploration} followed by \textit{confirmation}. During exploration, the doctor mainly asks questions that keep multiple pathologies under evaluation. In that phase, pathologies can be removed from or added to the differential depending on the patient's answers. During the confirmation phase, the doctor inquires about evidences that strengthen the actual differential they are considering \citep{mansoor2018frameworks,richardson2015the,rhoads2017formulating}. This two-phase property of the interaction of doctors with patients is a separate dimension of the acquisition process that differs from the objective of predicting differentials, and should therefore be considered and measured separately from the common paradigm of optimizing prediction accuracy.

\paragraph{Severe pathologies}

Among all possible diseases, severe pathologies receive more attention than others. Generally, a doctor does not want to miss out on them and would therefore explicitly gather evidences to rule them out from the differential as soon as they become plausible, even if they are less unlikely than other diseases \citep{rhoads2017formulating, Ramanayake2018EvaluationOR}. Emulating this behavior within automated systems would contribute to increasing the trust of doctors in these systems.

In this work, we make a case for explicitly modelling the reasoning of doctors in designing evidence acquisition and automated diagnosis models using RL. More specifically, we focus on (1) using the \textbf{differential diagnosis}, rather than the ground truth pathology, as the training target of models, (2) modulating the evidence acquisition process to mimic a doctor' \textbf{exploration-confirmation} approach, and (3) prioritizing \textbf{severe pathologies}. 

In the remaining sections, we review existing works and their shortcomings (Section \ref{s:related_work}); we 
describe our method that improves upon prior works especially in mimicking the reasoning of doctors (Section  \ref{s:method}); we provide empirical results demonstrating the benefits of the proposed approach (Sections \ref{s:experiments} and \ref{s:results}); finally, we discuss the limitations (Section \ref{s:limitation}) before concluding this work (Section \ref{s:conclusion}). Our main contributions are: (1) we reformulate the task of evidence acquisition and automated diagnosis by introducing doctors' trust as a desideratum, and argue for explicitly designing models towards this goal; (2) we propose a RL agent, CASANDE, that promotes the desired behavior in addition to accurate predictions, by means of predicting differentials and reward shaping; (3) we empirically show that existing strong models on this task, even with modifications, are not sufficient for the proposed goal of mimicking the reasoning of human doctors, and that CASANDE improves over existing models while being competitive on conventional metrics.

\section{Related work}\label{s:related_work}

There have been a variety of prior works on evidence acquisition and automatic diagnosis. For example, \citet{wei2018task} proposes to use DQN \citep{mnih2015human} to tackle this problem. \citet{kao2018context} additionally proposes to create a hierarchy of symptoms and diseases based on anatomical parts and train an agent using Hierarchical Reinforcement Learning \citep{sutton1999between}. \citet{xu2019end}, \citet{zhao2021weighted}, and \citet{liu2022my} propose to encode relations among different evidences and evidence-disease pairs to enhance the efficiency of an agent trained using DQN. \citet{peng2018refuel} uses a policy gradient algorithm \citep{William12Reinforce} and relies on reward shaping functions \citep{ng1999policy,wiewiora2003principled, devlin2012dynamic} defined in the state space to favour positive evidence acquisition. Our proposed approach also makes use of reward shaping functions, but unlike the work done in \citep{peng2018refuel}, those functions are defined in the action space and are used to induce a bias in our agent reflecting the reasoning of doctors.

In the aforementioned approaches, the space of the evidences to be collected and the one of the diseases to be predicted are merged together. \citet{janisch2019classification} considers dealing with those spaces separately and proposes an agent which has two branches, one that decides which evidence to inquire about, trained using RL, and one that predicts the disease, trained using supervised learning. Other approaches enhance the training of the evidence inquiry branch by using information from the classifier branch. For example, \citet{kachuee2019opportunistic} relies on Monte-Carlo dropout \citep{gal2016dropout} to estimate the certainty improvement from the classifier branch output and uses that information as an evidence acquisition reward to train the acquisition branch. \citet{yuan2021efficient} proposes an adaptive method to align the tasks performed by the two branches. More specifically, the acquisition branch receives an extra reward of the change in entropy of the disease prediction, and the acquisition process ends when the entropy of the prediction is below a dynamically learned, disease-specific threshold. Our approach also uses the two-branches setting and designs a mechanism to inform the acquisition process when to stop based on information from the classifier branch. Unlike \citet{yuan2021efficient}, we explicitly introduce a \textit{stop action} whose Q-value is trained to replicate a reward derived from the classification branch.

The evidence acquisition branch can be trained with learning paradigms different from RL. \citet{chen2021diaformer} proposes to model the evidence acquisition process as a sequential generation task and trains the agent in a similar manner to BERT's Masked Language Modelling (MLM) objective \citep{devlin2019bert}. Similarly, \citet{Luo2020KnowledgeGC} relies on randomly generated trajectories to train a system to collect evidences from patients in a supervised way. Finally, \citet{guan2021bayesian} makes use of the Quick Medical Reference belief network \citep{milner86QMR}, and applies a Bayesian experimental design \citep{Chaloner95Bed} in the inquiry phase while relying on Bayes rule to infer the corresponding disease distribution.

However, most of these approaches are primarily focused on predicting the single ground truth pathology, as opposed to the differential diagnosis in our work. Also, they have rarely made specific efforts to shape or guide the interaction to resemble how doctors would interact with patients. As a result, it is debatable how much doctors could trust such systems, even though some may perform well on benchmark datasets, therefore casting doubt on their applicability in real-world applications.

\section{Method}\label{s:method}
Let $E$ and $D$ be the number of all evidences (i.e., symptoms and antecedents) and diseases under consideration. Evidences can be \textit{binary} (e.g., \textit{are you coughing?}), \textit{categorical} (e.g., \textit{what is the pain intensity on a scale of 1 to 10?}), or \textit{multi-choice} (e.g., \textit{where is your pain located?}).
The task of automated evidence collection and diagnosis can be viewed as a sequential decision process where each interaction with a patient is formalized using a finite-horizon Markov Decision Process $ \mathcal{M} = (\mathcal{S}, \mathcal{A}, {P}, {R}, \gamma, T) $ with a state space $\mathcal{S}=\mathcal{S'}\cup\{\mathbf{s}_{\perp}\}$ (with $\mathbf{s}_{\perp}$ being the terminal state), an action space $\mathcal{A}$, a dynamics ${P}$, a reward function ${R}$, a discount factor $\gamma$, and a maximum episode length $T$. A state $\mathbf{s} \in \mathcal{S'}$ encodes socio-demographic data regarding the patient (e.g., age, sex) as well as the evidences provided by the patient so far. $\mathcal{A}$ is defined as $\{1, \cdots, E, E+1\}$ where the first $E$ elements, referred to as \textit{acquisition actions}, are used to inquire about corresponding evidences, and the last element is the \textit{exit action} that is used to explicitly terminate the interaction. At any point in time, only actions not yet selected in the episode are available for future selection. ${P}$ is deterministic and updates the current state based on the current action and the patient response if the current turn is less than $T$, otherwise, the next state is set to $\mathbf{s}_{\perp}$. Finally, regarding ${R}$ which is defined on the space $\mathcal{S} \times \mathcal{A} \times \mathcal{S}$ like ${P}$, each acquisition action is characterized by an inquiry cost $r_i$, and depending on whether the underlying patient is experiencing the corresponding evidence, it may additionally incur a retrieval reward $r_{p}$ or a missing penalty $r_{n}$. 
Upon the termination of the interaction, a prediction regarding the patient's differential is made based on the collected evidences.

\begin{figure}
    \centering
    \includegraphics[width=\textwidth]{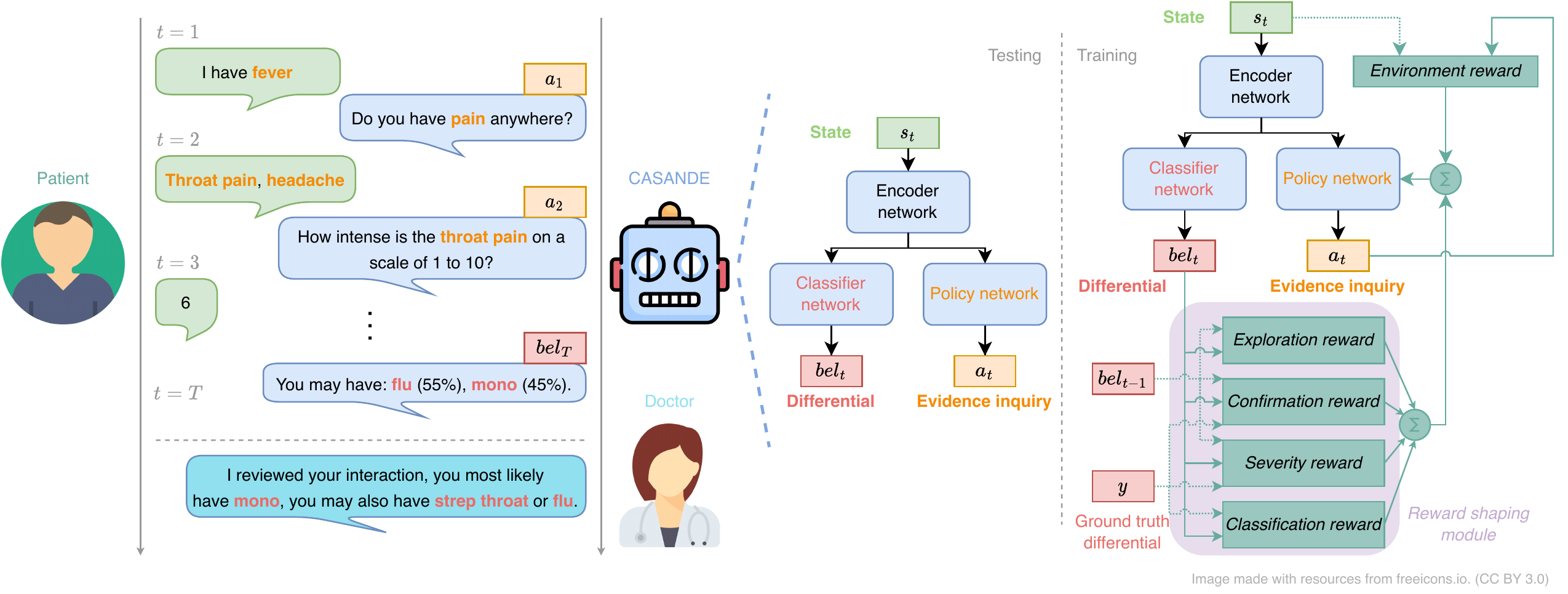}
    \caption{\textbf{Left}: A synthetic example of a patient's interaction with our proposed model. Importantly, the role of human doctors as the final decision maker is largely overlooked in prior works. The example is for illustration purposes only and is not meant as medical advice. \textbf{Middle}: The agent consists of two parts: an evidence acquisition module and a classifier responsible for generating the differential. \textbf{Right}: The reward shaping module leverages the output of the classifier to define auxiliary rewards that help induce desired features of the reasoning of medical doctors in the agent.}
    \label{fig:architecture}
\end{figure}

An overview of the proposed approach is depicted in Figure~\ref{fig:architecture}. Our model is made of two branches: an evidence acquisition branch in charge of the policy for interacting with patients and a classifier branch in charge of predicting the differential at each turn. Both the classifier and the policy networks rely on a latent representation computed by an encoder from the evidences collected so far from the patient. Importantly, we introduce a \textit{reward shaping module} which leverages the output of the classifier to define auxiliary rewards that help induce in the agent, during training, the desired properties of doctor reasoning identified in Section \ref{s:introduction} . More precisely, the \textit{exploration} and \textit{confirmation} rewards are designed to encourage the two-phase based interactions, whereas the \textit{severity} reward is used to explicitly handle severe pathologies. Finally, the \textit{classification} reward is used to train the agent to predict the right differential at the end of the interaction process. 
The proposed agent predicts at each turn $t$ the next action $\mathbf{a}_{t}$ as well as the belief ${bel}_{t}$ regarding the patient's differential from the current state $\mathbf{s}_{t}$.  We rely on a DQN variant algorithm, Rainbow~\citep{hessel2018rainbow} (without noisy networks~\cite{fortunato2018noisy}), to search for the  optimal policy and the objective of the proposed auxiliary rewards is to encourage trajectories similar to those of experienced medical practitioners. In the next sections, we thoroughly describe each of these auxiliary rewards.

\subsection{Exploration reward: encouraging the evaluation of multiple pathologies}
We want our agent to undertake actions that favour considering several pathologies in the differential diagnosis during the first phase of the interaction. In practice, this means that the agent's belief is constantly changing after inquiring about a new evidence. As more evidences are collected, this fluctuation should gradually reduce until it becomes insignificant. Based on this intuition, we derive the exploration reward $R_{\text{Ex}}$ as
\begin{equation}
    R_{\text{Ex}}(\mathbf{s}_t, \mathbf{a}_t, \mathbf{s}_{t+1}) = 
    \mathds{1}_{\mathbf{s}_{t+1} \neq \mathbf{s}_{\perp}} w_{\text{Ex}}(t) \times JSD(bel_{t}, bel_{t+1}),
    \label{eq:paper_JSDIV}
\end{equation}
where $JSD$ is the Jensen-Shannon divergence that computes the dissimilarity between consecutive beliefs $bel_t$ and $bel_{t+1}$,  and $w_{\text{Ex}}(t) \in [0, 1]$ is a dynamic weight which aims at controlling the importance of this shaping component as a function of time. Inspired from the shape of the \textit{sigmoid} function, we design $w_{\text{Ex}}(t)$ as a translated version in $[0, T]$ of the ``flipped'' sigmoid function within the interval $[\bar{x}_{min}, \bar{x}_{max}]$. In other words, using $trans(t) = \frac{t * (\bar{x}_{max} - \bar{x}_{min})}{T} + \bar{x}_{min}$, we have 
\begin{equation}
    w_{\text{Ex}}(t) = sigmoid(-\left(trans(t) + \Delta_{\text{Ex}}\right)),
    \label{eq:paper_div}
\end{equation}
where $\Delta_{\text{Ex}}$ is a hyper-parameter that controls how fast the function saturates at 0 (cf. Appendix~\ref{appx:paper_scheduler}).

The beliefs $bel_t$ and $bel_{t+1}$ are generated by the classifier branch based respectively on states $s_t$ and $s_{t+1}$, with the latter state resulting from the execution of action $a_t$ when in state $s_t$.

\subsection{Confirmation reward: Strengthening the agent's belief in the pathology}

In the second phase of the interaction with the patient, the agent should inquire about evidences that help strengthen its belief regarding the differential diagnosis. This means that the more the agent collects evidences, the closer its belief should be to the ground truth distribution $y$. The importance of this shaping component should gradually increase as the agent moves towards the end of the interaction. From this intuition and inspired from potential-based reward shaping \cite{ng1999policy}, we derive the confirmation reward $R_{\text{Co}}$ as
\begin{equation}
    R_{\text{Co}}(\mathbf{s}_t, \mathbf{a}_t, \mathbf{s}_{t+1}) = 
    - \mathds{1}_{\mathbf{s}_{t+1} \neq \mathbf{s}_{\perp}} w_{\text{Co}}(t) \times (\gamma \  CE(bel_{t+1}, y) - CE(bel_{t}, y)), 
    \label{eq:paper_CEDIV}
\end{equation}
where $CE$ is the cross-entropy function, and $w_{\text{Co}}(t) \in [0, 1]$ controls the importance of this shaping component as a function of time. Like $w_{\text{Ex}}(t)$, $w_{\text{Co}}(t)$ is a translated version of the sigmoid function with the difference that it increases over time. Thus, we have:
\begin{equation}
    w_{\text{Co}}(t) = sigmoid(\left(trans(t) + \Delta_{\text{Co}}\right)),
    \label{eq:paper_enf}
\end{equation}
where $\Delta_{\text{Co}}$ is a hyper-parameter controlling how fast the function saturates at 1 (cf. Appendix~\ref{appx:paper_scheduler}).

\subsection{Severity reward: evidence gathering for ruling out severe pathologies}

We want our agent to collect evidences that help rule out severe pathologies that are not part of the ground truth differential as soon as they become plausible. One proxy to achieve this for the agent is to behave in such a way to monotonically increase the number $SevOut$ of severe pathologies that are not in both the ground truth differential and its predictions through time. As such, we define the severity reward $R_{\text{Sev}}$ as 
\begin{equation}
    R_{\text{Sev}}(\mathbf{s}_t, \mathbf{a}_t, \mathbf{s}_{t+1}) = 
    \mathds{1}_{\mathbf{s}_{t+1} \neq \mathbf{s}_{\perp} \& SevOut_{t+1} \neq SevOut_t} (\gamma  SevOut_{t+1} - SevOut_{t}). 
    \label{eq:paper_SevDIV}
\end{equation}
$SevOut_t$ and $SevOut_{t+1}$ correspond to the number of severe pathologies which are not in both the ground truth differential and the differentials respectively predicted at time $t$ and $t+1$, i.e., the beliefs $bel_t$ and $bel_{t+1}$. Here, a pathology is not part of a predicted distribution if its probability is below a given threshold.

\subsection{Classification reward}

This reward is designed to provide feedback to the agent describing how good its final predicted differential is with respect to the ground truth differential $y$ when the interaction process is over. Let $SevIn_t$ be the number of severe pathologies that are part of both the ground truth differential $y$ and the predicted belief $bel_t$, and let $Sev_y$ be the number of severe pathologies in $y$. As noted previously, a pathology is not part of a predicted distribution if its probability is below a given threshold. We define $\mathcal{V}(\mathbf{s}_t, y) = -CE(bel_t, y) + w_{{\text{si}}} \frac{SevIn_t}{Sev_y}$ as a measure of the quality of the belief prediction from state $\mathbf{s}_t$ where $CE$ is the cross-entropy function, and the second term, controlled by the hyper-parameter $w_{\text{si}}$, measures the inclusion rate of the relevant severe pathologies in $bel_t$. We thus define the classification reward $R_{\text{Cl}}$ as 
\begin{equation}
    R_{{\text{Cl}}}(\mathbf{s}_t, \mathbf{a}_t, \mathbf{s}_{t+1}) = 
    \mathds{1}_{\mathbf{s}_{t+1} = \mathbf{s}_{\perp}} \mathcal{V}(\mathbf{s}_t, y). 
    \label{eq:paper_ClfDIV}
\end{equation}

\subsection{Training Method}

The above mentioned auxiliary rewards are combined together as
\begin{equation}
    F = \alpha_{{\text{Ex}}} R_{{\text{Ex}}}  +  \alpha_{{\text{Co}}} R_{{\text{Co}}}  +  \alpha_{{\text{Sev}}} R_{{\text{Sev}}}  +  \alpha_{{\text{Cl}}} R_{{\text{Cl}}},
    \label{eq:shapingFunc}
\end{equation}
where the $\alpha$ weights are hyper-parameters, and the $(\mathbf{s}_t, \mathbf{a}_t, \mathbf{s}_{t+1})$ arguments for $F$ and all reward functions were dropped for clarity. The resulting function $F$ is then combined with the environment reward $R$, leading to a new reward function $R' = R + F$ that is used to train the policy network.

For simplicity, this section relies on DQN to present the loss for training each part of our agent, with the extension to Rainbow easy to do. Let $\theta$ and $\phi$ be the agent and target network parameters.

\paragraph{Policy network loss function:}
The Q-value of the \textit{exit} action does not depend on any following state. Also, the reward received when explicitly exiting is the classification reward. Therefore, to synchronize the agent branches, the \textit{exit} action Q-value needs to be reconciled with the classification reward. This should be done at each interaction time step during the training, given that the agent needs to have a good estimate of the expected Q-value of the \textit{exit} action to select this action. Thus,
\begin{equation}
    Loss_{Q}(B) = \frac{1}{2} \mathop{\mathbb{E}}_{(\mathbf{s}_t, \mathbf{a}_t, r_t, \mathbf{s}_{t+1}, y) \in B} \left[\medmath{\left[Q_t - Q_{\theta}({\mathbf{s}_t}, {\mathbf{a}_t}) \right]^2 + \mathds{1}_{{\mathbf{s}_{t+1}} \neq {\mathbf{s}_{\perp}}} \left[Q_{\theta}({\mathbf{s}_t}, {\mathbf{a}_{exit}}) - \mathcal{V}({\mathbf{s}_{t}}, y) \right]^2}\right]
    \label{eq:DQNlossComponent}
\end{equation}
where $B$ is a batch from the replay buffer, $r_t = R({\mathbf{s}_t}, {\mathbf{a}_t}, {\mathbf{s}_{t+1}})$ is the $R$ reward at time $t$, $Q_{\theta}({\mathbf{s}_t}, {\mathbf{a}_{exit}})$ is the predicted Q-value for the pair $({\mathbf{s}_t, \mathbf{a}_{exit}})$, $Q_t = {r'_t} + \mathds{1}_{{\mathbf{s}_{t+1}} \neq {\mathbf{s}_{\perp}}} \gamma \max_{\mathbf{a} \in \mathcal{A}} Q_{\phi}({\mathbf{s}_{t+1}}, \mathbf{a})$ is the target Q-value for the pair $({\mathbf{s}_t, \mathbf{a}_t})$ with $r'_t = R'({\mathbf{s}_t}, {\mathbf{a}_t}, {\mathbf{s}_{t+1}}) = r_t + F(\mathbf{s}_t, \mathbf{a}_t,\mathbf{s}_{t+1})$.

\paragraph{Classifier network loss function:} The classifier is updated using the following loss at the end of an interaction:
\begin{equation}
    Loss_{C}(B) = \frac{1}{2} \mathop{\mathbb{E}}\limits_{({\mathbf{s}_t, \mathbf{a}_t, r_t, \mathbf{s}_{t+1}, y}) \in B} \left[\medmath{\mathds{1}_{{\mathbf{s}_{t+1}} = {\mathbf{s}_{\perp}}} CE(bel_{t}, y)}\right].
    \label{eq:ClassifierlossComponent}
\end{equation}

\paragraph{Training process: } Using the loss functions defined in Equations~\ref{eq:DQNlossComponent} and~\ref{eq:ClassifierlossComponent}, the two branches are alternatively updated until the stopping criteria are satisfied. Both updates share the same batch of data $B$ which is sampled from the replay buffer. The training process is described in Appendix~\ref{appx:paper_algo}.

\section{Experiments}\label{s:experiments}


\paragraph{Datasets}\label{ss:dataset}

The datasets used in prior works such as DX \citep{wei2018task}, Muzhi \citep{xu2019end}, SymCAT \citep{peng2018refuel}, HPO \citep{guan2021bayesian}, and MedlinePlus \citep{yuan2021efficient} do not provide differential diagnosis information, and are therefore ineligible for validating the approach proposed in this work. Besides, as supported by \citet{yuan2021efficient}, the patients simulated using SymCAT are not sufficiently realistic for testing automatic diagnosis systems. We instead use the \CasandeDDx dataset \citep{Fansi_Tchango_DDXPlus_A_new_2022} for that purpose. In addition to providing differential diagnosis and pathology severity information, the \CasandeDDx dataset, unlike the above mentioned datasets, is not restricted to binary evidences, but also has categorical and multi-choice evidences, allowing for more efficient interactions with patients. Furthermore, the dataset contains 49 pathologies and 223 evidences (corresponding to 110 symptoms and 113 antecedents). Finally, the dataset is split into 3 subsets: a training subset with more than $10^6$ synthetic patients, and validation and test subsets containing roughly $1.4 \times 10^5$ synthetic patients each.

\paragraph{Baselines}

We consider 4 baselines, AARLC (short for Adaptive Alignment of Reinforcement Learning and Classification) \citep{yuan2021efficient}, Diaformer \citep{chen2021diaformer}, BED \citep{guan2021bayesian}, and BASD  which is an extension on \citet{Luo2020KnowledgeGC}. AARLC demonstrates SOTA results on the SymCAT dataset while Diaformer shows competitive results on the Muzhi and DX datasets. BED, an approach that does not require training, has impressive results on the HPO dataset. All these methods are designed to only handle binary evidences and had to be modified to deal with the different evidence types in the \CasandeDDx dataset. Moreover, those  methods were designed to predict the ground truth pathology, and had to be modified to handle differentials. Additional details are provided in Appendix~\ref{appx:paper_baselines}.

\paragraph{Experimental settings}

Each patient in the \CasandeDDx dataset is characterized by a chief complaint, which is presented to the models at the beginning of the interaction, a set of additional evidences that the models need to discover through inquiry, a differential diagnosis, and a ground truth pathology. We allow interactions to have a maximum of $T = 30$ turns. We tune the hyper-parameters for each model, including the baselines, separately on the validation set and use the resulting optimal set of parameters to report the performances. For more details, see  Appendix~\ref{app:training_details}.

\paragraph{Evaluation metrics}
We report on the interaction length (IL), the recall of the inquired evidences (PER), the ground truth pathology accuracy when only considering the top entry of the  differential (GTPA@1) and when considering all entries (GTPA), the F1 score of the predicted differential diagnosis (DDF1), and the harmonic mean of the rule-in and rule-out rates of severe pathologies (DSHM). See Appendix~\ref{appx:metric} for the definition of these metrics. The GTPA@1 metric is only relevant for models trained to predict the patient's ground truth pathology. It is not relevant for models trained to predict the differential as the ground truth pathology is not always the top pathology in the differential \citep{Fansi_Tchango_DDXPlus_A_new_2022}. We do not report the precision of the inquired evidences as it can be useful and sometimes necessary to ask questions that lead to negative answers from patients. For example, such questions need to be asked in order to safely rule out severe pathologies. Measuring the quality of the collected negative evidence is left as future work. 

\section{Results}\label{s:results}

\begin{table}[tb!]
\centering
\caption{Evaluation performance on the test set. Values indicate the average of 3 runs, and numbers in brackets indicate 95\% confidence intervals. Values in \textbf{bold} indicate the best performance for a column, and values in \textit{italic} indicate those that are not statistically significantly worse than the best ($p > 0.05$). BED is deterministic and related results are from 1 run. $\downarrow$ indicates lower is better and $\uparrow$ indicates higher is better.}
\vspace{1mm}
\scriptsize
\begin{tabular}{cccccccc}
\cline{4-8}
\multicolumn{1}{l}{}                                                                                           & \multicolumn{1}{l}{} & \multicolumn{1}{l}{}  & \multicolumn{5}{c}{Values expressed in percentage (\%) $\uparrow$}                                                        \\ \cline{4-8} 
                                                                                                               & \textbf{Method}      & \textbf{IL $\downarrow$}           & \textbf{GTPA@1} & \textbf{DDF1}         & \textbf{DSHM}         & \textbf{PER}          & \textbf{GTPA}         \\ \hline
\multicolumn{1}{c|}{\multirow{3}{*}{\begin{tabular}[c]{@{}c@{}}Trained \\ With \\ Differential\end{tabular}}}  & \textbf{CASANDE}     & 19.71 (0.46)          & 69.7 (3.54)     & \textbf{94.24 (0.55)} & \textbf{73.88 (0.34)} & \textbf{98.39 (0.86)} & {99.77 (0.03)}  \\ \cline{2-8} 
\multicolumn{1}{c|}{}                                                                                          &
\textbf{Diaformer}     & 18.41 (0.07)          & 73.62 (0.60)     & 83.3 (2.10)          & 69.32 (0.73)          & {92.92 (0.30)} & {99.01 (0.33)} \\ \cline{2-8} 
\multicolumn{1}{c|}{}                                                                                          & \textbf{AARLC}       & 25.75 (2.75) & 75.39 (5.53)     & 78.24 (6.82)           & 69.43 (2.01)          & 54.55 (14.73)           & 99.92 (0.03)          \\ \cline{2-8} 
\multicolumn{1}{c|}{}                                                                                          & \textbf{BASD}         & 17.86 (0.88)          & 67.71 (1.19)    & 83.69 (1.57)          & 65.06 (2.30)          & 88.18 (1.12)          & 99.30 (0.27)          \\ \hline
\multicolumn{1}{c|}{\multirow{4}{*}{\begin{tabular}[c]{@{}c@{}}Trained\\ Without\\ Differential\end{tabular}}} & \textbf{CASANDE}     & 19.53 (0.87)          & 98.8 (0.54)     & 30.84 (0.32)          & 10.62 (0.22)          & \textit{98.07 (1.91)} & {99.46 (0.46)} \\ \cline{2-8} 
\multicolumn{1}{c|}{}                                                                                          &
\textbf{Diaformer}     & 18.45 (0.33)          & 91.81 (1.95)     & 30.38 (14.72)          & 19.90 (31.07)         & {92.61 (1.12)} & {96.28 (6.21)} \\ \cline{2-8} 
\multicolumn{1}{c|}{}                                                                                          &
\textbf{AARLC}       & 6.73 (1.35)           & \textbf{99.21 (0.78)}    & 31.28 (0.38)          & 10.96 (0.26)          & 32.78 (13.92)         & \textbf{99.97 (0.01)}          \\ \cline{2-8} 
\multicolumn{1}{c|}{}                                                                                          & \textbf{BASD}         & 17.99 (3.57)          & 97.15 (1.70)    & 31.31 (0.29)          & 10.81 (0.29)          & 88.45 (5.78)          & 98.82 (1.03)          \\ \cline{2-8} 
\multicolumn{1}{c|}{}                                                                                          & \textbf{BED}         & \textbf{5.47}         & 99.47           & 31.01                 & 10.78                 & 18.62                 & 99.76                 \\ \hline
\end{tabular}
\label{tbl:main_results}
\end{table}

\paragraph{Disease prediction and evidence acquisition}\label{ss:results_prediction}

Table \ref{tbl:main_results} depicts the performance of each model at the end of the interaction process. We first focus on metrics measuring the quality of the predicted diseases. All models are on par with regard to the recovery of the ground truth pathology (GTPA), even when they are trained to predict the differential and are not given any indication about the pathology in the ground truth differential that corresponds to the disease the patient is suffering from. As for recovering the right differential (DDF1), models perform poorly when trained to predict the ground truth pathology even if some of those models generate a posterior pathology distribution. Performance greatly improves when models are trained to predict the differential, with CASANDE doing significantly better than all other models, outperforming the second-best model by more than $10\%$ absolute on average. Likewise, CASANDE outperforms the baselines in handling the severe pathologies (DSHM) with an absolute margin of more than $4\%$ when trained with the differential. Those results indicate that training on the differential is essential and that CASANDE significantly outperforms the baselines.

CASANDE achieves the best performance on the PER metric demonstrating its ability to collect relevant evidences. We observe that the PER score also improves for AARLC when trained on the differential, suggesting that differentials are beneficial for collecting evidences. This improvement is not observed for BASD because its differential classifier branch only operates at the end of the interaction process, after the evidence collection. Similarly, Diaformer generates the differential at the end of the evidence collection and doesn't benefit from it to improve the collection process.

Finally, CASANDE has the longest interaction length. This is unsurprising, as CASANDE is trained to emulate the reasoning of doctors. In particular, the exploration and confirmation phases may lead to the model asking more questions. Additionally, CASANDE's interaction length is still considerably lower than the maximal value  (30), which indicates that CASANDE is capable of terminating an interaction rather than always continuing to the end. Together, this shows that the slight increase in CASANDE's interaction length (less than 2 turns compared to the model with the shortest length) may be due to the different priorities of CASANDE and the baselines.

A qualitative evaluation of CASANDE was performed by a doctor who analyzed CASANDE's predictions for 20 patients randomly selected from the test set. The profiles of those patients, the predictions of CASANDE as well as the doctor's evaluation are presented in Appendix \ref{sec:qualEval}. The doctor concluded that the evidences collected by CASANDE are indeed helpful for establishing a differential. Meanwhile, he cannot definitively assess CASANDE's disease predictions since, unlike CASANDE, the scope of diseases he considers is not limited to the 49 pathologies in DDXPlus. 

\paragraph{Trajectory quality evaluation}\label{ss:results_traj_quality}
\begin{wrapfigure}[22]{R}{0.55\textwidth}
    \centering
    \includegraphics[width=0.55\textwidth]{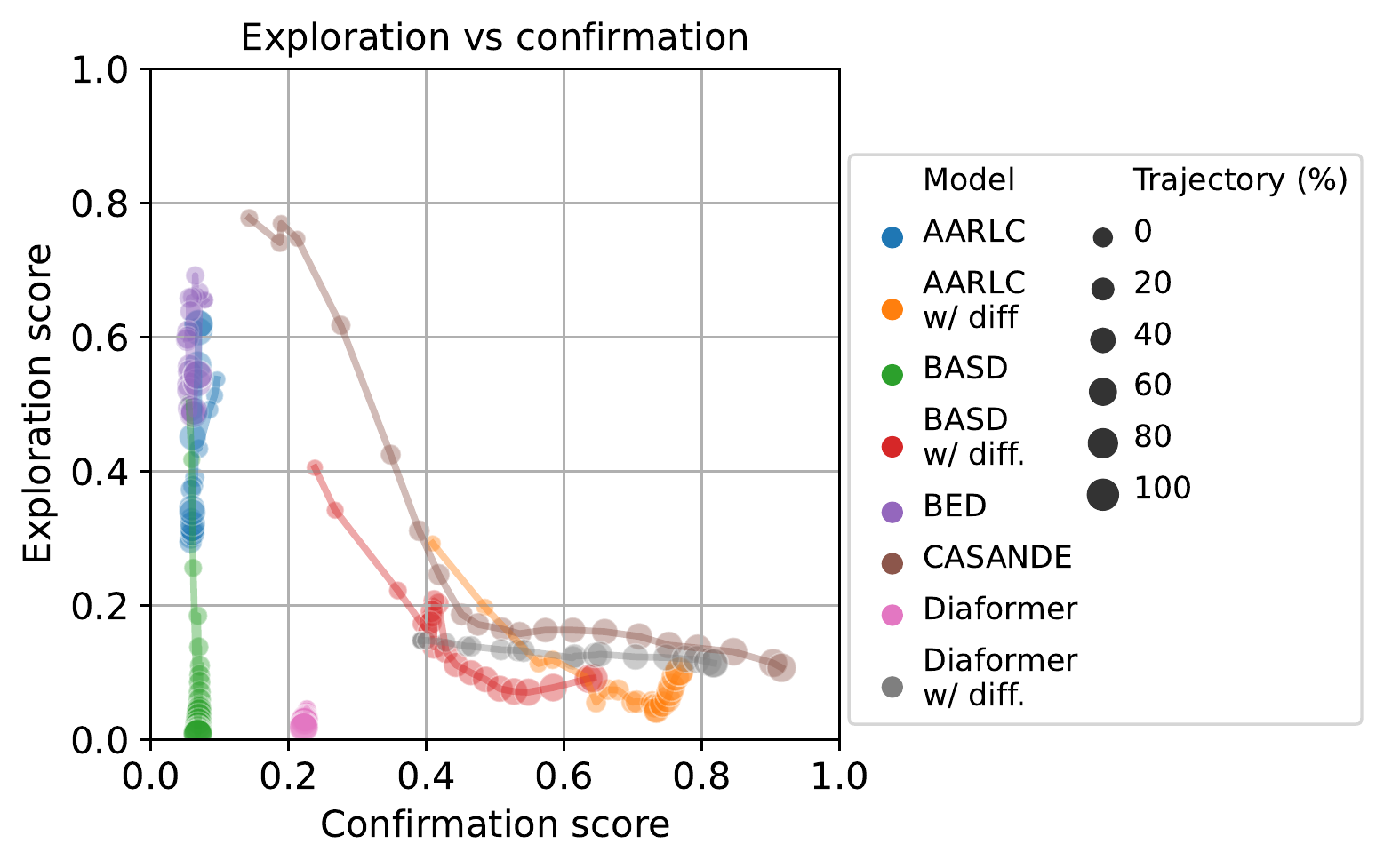}
    \caption{Confirmation score vs exploration score throughout trajectories on average for each model on the test set. For each trajectory, the scores are computed for each turn. We then select a fixed number of 21 equally spaced turns, which represent 0\%, 5\%, $\cdots$, 100\% of the trajectory, and visualize the scores at these turns. Finally, we average these 21 turns across trajectories for each method.}
    \label{fig:explore_confirm}
\end{wrapfigure}
We plot in Figure~\ref{fig:explore_confirm} the \textit{confirmation score} versus the \textit{exploration score} throughout the trajectories for the different models. The exploration score captures how distant two consecutive agent predictions are, while the confirmation score captures how close the agent prediction is with respect to the target distribution (see Appendix \ref{app:explore-confirm} for the definitions). Ideally, an agent would start with a high exploration score and a low confirmation score (upper-left corner), gradually decreases the former and increases the latter, until the exploration score reaches a low value and the confirmation score reaches a high value at the end of the interaction (lower-right corner). As can be observed, this trend is highly followed by CASANDE which exhibits the highest exploration score at the beginning of the interaction and the highest confirmation score at the end of the interaction while, unlike other models, it is consistently moving towards the lower-right corner of the chart.
Appendix \ref{sec:exampleInteractions} presents the differentials predicted by CASANDE at each interaction turn for 3 patients from the test set. We clearly observe CASANDE exploring various differentials at the beginning of the interaction and then focusing on a differential towards the end of the interaction.

\begin{figure}[b]
    \centering
    \includegraphics[width=\textwidth]{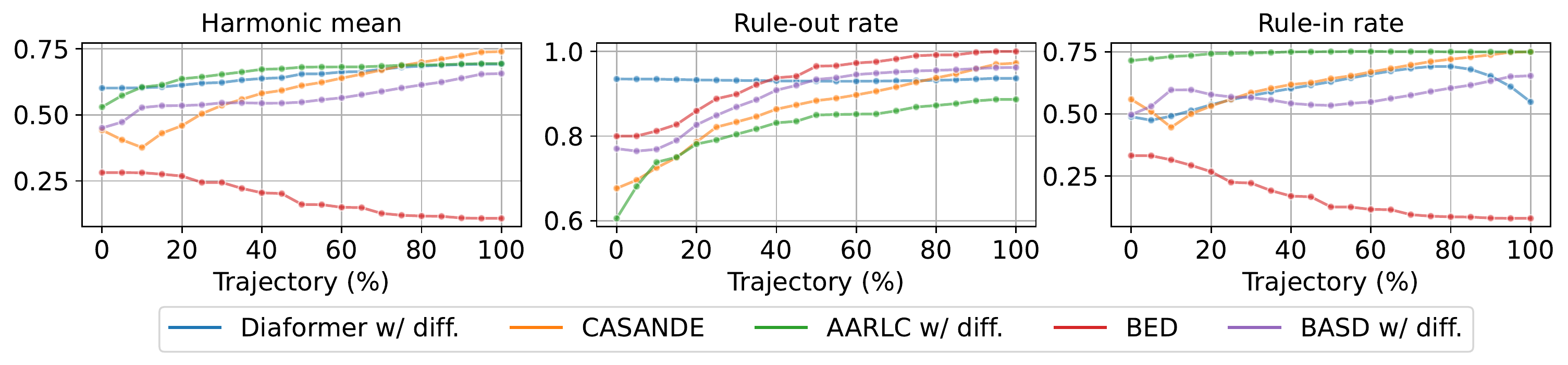}
    \caption{Average harmonic mean of the rule-out and rule-in rates, rule-out rate and rule-in rate on severe pathologies throughout trajectories. For each trajectory, their values are calculated at each turn.  We then select a fixed number of 21 equally spaced turns, which represent 0\%, 5\%, $\cdots$, 100\% of the trajectory, and visualize the scores at these turns. Finally, we average these 21 turns across trajectories for each method. The results are on the test set. 
    }
    \label{fig:severe_metrics}
\end{figure}

\paragraph{Handling of severe pathologies}
\label{ss:results_severe}
We are interested in analysing the pace at which severe pathologies are ruled out from or ruled in within the predicted differentials throughout the trajectories. We show in Figure \ref{fig:severe_metrics} the curves corresponding to the average values of the harmonic mean scores of the rule-in and rule-out rates (i.e., DSHM), the \textit{rule-out} rate, and the \textit{rule-in} rate of severe pathologies throughout the trajectories (see Appendix \ref{app:end_performance_metrics} for the definitions).
As observed, CASANDE improves significantly on the harmonic mean throughout a trajectory, and eventually performs the best among all models. This shows that the capability of CASANDE for handling severe pathologies is heavily grounded on the evidence it gradually collects. This evidence-based behavior is further confirmed by the rule-out and rule-in rates as CASANDE is the only method that significantly improves on both rates throughout the trajectories.

Interestingly, when focusing on the rule-out curve, it is noticeable that CASANDE performs the second-worst at the beginning, trailing the best performing model by more than $20\%$, but it improves to the second-best at the end, with the gap smaller than $3\%$. This suggests that at the beginning, CASANDE is more lenient on including severe diseases in its differentials, resulting in more false positives than others. However, this is a desired behavior of CASANDE, as during the exploration phase, it is expected to consider unlikely but severe diseases before gathering enough evidences to rule them out. In other words, at the beginning of a trajectory where few evidences are gathered, a model should err on the side of caution by keeping unlikely but severe diseases into consideration.

\paragraph{Ablation studies}

We perform an ablation analysis to illustrate the contribution of each auxiliary reward component of CASANDE to the overall performance. Table~\ref{tbl:ablations} shows the results when some reward components are deactivated. We observe a drop in DDF1 and DSHM when most rewards are disabled. This indicates that each reward function helps the agent capture complementary information that is useful for predicting the differential. Also, as expected, the exploration and confirmation scores contribute in increasing the interaction length. Finally, we observe a decrease of the DSHM metric when the severity reward is not used. Additional studies are provided in Appendix~\ref{app:ablations}.

\begin{table}[]
\centering
\caption{Impact of the auxiliary reward components on the differential prediction and evidence collection, as measured on the validation set. All experiments are repeated 3 times with different random seeds. Values are in \% except for IL. Values are expressed as an average $\pm$ a standard deviation. Values in brackets indicate 95\% confidence intervals. Values in \textbf{bold} indicate the best performance for a column, and values in \textit{italic} indicate those that are not statistically significantly worse than the best ($p > 0.05$).}
\label{tbl:ablations}
\resizebox{\textwidth}{!}{%
\begin{tabular}{cccc|cccc}
$R_{\text{Ex}}$ & $R_{\text{Co}}$ & $R_{\text{Sev}}$ & $R_{\text{Cl}}$ & \textbf{IL} & \textbf{DDF1} & \textbf{DSHM} & \textbf{PER} \\ \hline
$\times$ & $\times$ & $\times$ & $\times$ & \textbf{17.56 (0.25) $\pm$ 0.06} & 93.11 (0.70) $\pm$ 0.16 & 73.55 (0.27) $\pm$ 0.06 & \textit{98.24 (0.36) $\pm$ 0.08} \\ \hline
$\times$ & $\times$ & $\times$ & $\checkmark$ & 17.79 (0.13) $\pm$ 0.03 & 93.48 (1.31) $\pm$ 0.30 & 73.33 (0.28) $\pm$ 0.06 & \textit{98.32 (0.88) $\pm$ 0.20} \\ \hline
$\checkmark$ & $\checkmark$ & $\checkmark$ & $\times$ & 19.09 (0.38) $\pm$ 0.09 & \textit{94.26 (2.82) $\pm$ 0.65} & \textbf{74.19 (0.41) $\pm$ 0.10} & \textit{98.34 (1.13) $\pm$ 0.26} \\ \hline
$\times$ & $\checkmark$ & $\checkmark$ & $\checkmark$ & 18.01 (0.36) $\pm$ 0.08 & \textbf{94.49 (0.27) $\pm$ 0.06} & \textit{73.92 (0.38) $\pm$ 0.09} & \textit{98.48 (1.10) $\pm$ 0.26} \\ \hline
$\checkmark$ & $\times$ & $\checkmark$ & $\checkmark$ & 18.91 (1.36) $\pm$ 0.32 & \textit{94.16 (0.69) $\pm$ 0.16} & \textit{73.86 (0.34) $\pm$ 0.08} & \textbf{98.54 (0.68) $\pm$ 0.16} \\ \hline
$\checkmark$ & $\checkmark$ & $\checkmark$ & $\checkmark$ & 19.71 (0.46) $\pm$ 0.11 & \textit{94.24 (0.55) $\pm$ 0.13} & \textit{73.88 (0.34) $\pm$ 0.08} & \textit{98.39 (0.86) $\pm$ 0.20}
\end{tabular}%
}
\end{table}

\section{Limitations and potential negative social impact}\label{s:limitation}

\paragraph{Limitations} 

In this work, we set out to include doctors' trust as part of the desiderata in building evidence acquisition and automated diagnosis systems. In doing so, we focus our efforts on three essential aspects of the reasoning of doctors, namely the generation of a differential diagnosis, the exploration-confirmation approach, and the prioritized handling of severe pathologies. We chose to focus on them as advised by our collaborating physician. However, to most accurately evaluate doctors' trust, the proposed method has to be deployed so as to allow doctors different from the one we collaborated with to judge our approach. Additionally, it is reasonable to assume that doctors in other locations and/or specialties may have different ways of engaging with patients. Therefore, while we provide evidence in the medical literature (Section \ref{s:introduction}) to show the identified desiderata are widely applicable, we recognize that there may be cases where this work does not apply. In this work, we conduct experiments on synthetic patients, mainly due to the lack of real-patient datasets that contain differential diagnoses. We recognize that synthetic patients can be different from real patients in various and important ways, and that therefore results reported on synthetic patients may not extend to real patients.

\paragraph{Potential negative social impact}

As we have made clear from the beginning, evidence acquisition and automated diagnosis systems are not substitutes for human doctors, but rather they are supportive tools for doctors, who should make the final decisions. However, the predictions of such systems potentially might be provided to patients as the final medical advice, without the intervention of human doctors. In such cases, the instructions that patients receive may be misleading or erroneous, and thus can do more harm than good to patients' health.

\section{Conclusion}\label{s:conclusion}

We reflected on the task formulation of evidence acquisition and automatic diagnosis that are essential for telemedicine services, and introduced doctors' trust as an additional desideratum. Concretely, we argued that emulating the reasoning of doctors is critical for gaining their trust, and we identified three essential doctor reasoning features that models can mimic. We proposed a novel RL agent with these features built in. We showed empirically existing models are insufficient for imitating the reasoning of doctors. We then demonstrated the importance of the explicit modelling of the differential diagnosis, and the efficacy of our model in emulating doctors, while being competitive on conventional metrics. This work is a first step towards reshaping the research in automatic diagnosis systems, and there is abundant potential for future work to explore. First, it is important to continue working with doctors to determine whether additional medical elements need to be considered when building machine learning approaches. Second, we need to build datasets that cover a wide spectrum of pathologies and train agents using this extensive diagnostic space along with all the corresponding evidences to uncover if the learned strategies are similar to expert doctors and how well machine learning approaches scale when the action space and the pathology space become much larger. Third, it is important to find a way of measuring the quality of collected negative evidences. Finally, it would be useful to consider online learning methods where doctors could identify missing evidences and give their overall feedback on the collected medical history.

\begin{ack}
We would like to thank Dialogue Health Technologies Inc.  for providing us access to the physician who supported us throughout this work and to some of the computational resources used to run the experiments. We would also like to thank Quebec's Ministry of Economy and Innovation and Invest AI for their financial support.
\end{ack}

\clearpage

\bibliographystyle{plainnat}
\bibliography{refs}

\begin{thebibliography}{38}
\providecommand{\natexlab}[1]{#1}
\providecommand{\url}[1]{\texttt{#1}}
\expandafter\ifx\csname urlstyle\endcsname\relax
  \providecommand{\doi}[1]{doi: #1}\else
  \providecommand{\doi}{doi: \begingroup \urlstyle{rm}\Url}\fi

\bibitem[Chaloner and Verdinelli(1995)]{Chaloner95Bed}
Kathryn Chaloner and Isabella Verdinelli.
\newblock {Bayesian Experimental Design: A Review}.
\newblock \emph{Statistical Science}, 10\penalty0 (3):\penalty0 273 -- 304,
  1995.

\bibitem[Chen et~al.(2022)Chen, Li, Chen, Zhou, and Liu]{chen2021diaformer}
Junying Chen, Dongfang Li, Qingcai Chen, Wenxiu Zhou, and Xin Liu.
\newblock {Diaformer: Automatic Diagnosis via Symptoms Sequence Generation}.
\newblock In \emph{Proceedings of the AAAI Conference on Artificial
  Intelligence}, volume~36, pages 4432--4440, 2022.

\bibitem[Devlin et~al.(2019)Devlin, Chang, Lee, and Toutanova]{devlin2019bert}
Jacob Devlin, Ming-Wei Chang, Kenton Lee, and Kristina Toutanova.
\newblock {BERT}: Pre-training of deep bidirectional {T}ransformers for
  language understanding.
\newblock In \emph{Proceedings of the 2019 Conference of the North {A}merican
  Chapter of the Association for Computational Linguistics: Human Language
  Technologies, Volume 1 (Long and Short Papers)}, pages 4171--4186,
  Minneapolis, Minnesota, June 2019. Association for Computational Linguistics.
\newblock \doi{10.18653/v1/N19-1423}.
\newblock URL \url{https://aclanthology.org/N19-1423}.

\bibitem[Devlin and Kudenko(2012)]{devlin2012dynamic}
Sam~Michael Devlin and Daniel Kudenko.
\newblock Dynamic potential-based reward shaping.
\newblock In \emph{Proceedings of the 11th International Conference on
  Autonomous Agents and Multiagent Systems}, pages 433--440. IFAAMAS, 2012.

\bibitem[Esteva et~al.(2019)Esteva, Robicquet, Ramsundar, Kuleshov, DePristo,
  Chou, Cui, Corrado, Thrun, and Dean]{esteva2019guide}
Andre Esteva, Alexandre Robicquet, Bharath Ramsundar, Volodymyr Kuleshov, Mark
  DePristo, Katherine Chou, Claire Cui, Greg Corrado, Sebastian Thrun, and Jeff
  Dean.
\newblock {A Guide to Deep Learning in Healthcare}.
\newblock \emph{Nature Medicine}, 25\penalty0 (1):\penalty0 24--29, 2019.

\bibitem[Fansi~Tchango et~al.(2022)Fansi~Tchango, Goel, Wen, Martel, and
  Ghosn]{Fansi_Tchango_DDXPlus_A_new_2022}
Arsene Fansi~Tchango, Rishab Goel, Zhi Wen, Julien Martel, and Joumana Ghosn.
\newblock {DDXPlus: A New Dataset For Automatic Medical Diagnosis}.
\newblock In \emph{Proceedings of the Neural Information Processing Systems -
  Track on Datasets and Benchmarks}, volume~2, 2022.

\bibitem[Fortunato et~al.(2018)Fortunato, Azar, Piot, Menick, Hessel, Osband,
  Graves, Mnih, Munos, Hassabis, et~al.]{fortunato2018noisy}
Meire Fortunato, Mohammad~Gheshlaghi Azar, Bilal Piot, Jacob Menick, Matteo
  Hessel, Ian Osband, Alex Graves, Volodymyr Mnih, Remi Munos, Demis Hassabis,
  et~al.
\newblock {Noisy Networks For Exploration}.
\newblock In \emph{Proceedings of the International Conference on Learning
  Representations}, 2018.

\bibitem[Gal and Ghahramani(2016)]{gal2016dropout}
Yarin Gal and Zoubin Ghahramani.
\newblock {Dropout as a Bayesian Approximation: Representing Model Uncertainty
  in Deep Learning}.
\newblock In \emph{Proceedings of the International Conference on Machine
  Learning}, pages 1050--1059. PMLR, 2016.

\bibitem[Guan and Baral(2021)]{guan2021bayesian}
Hong Guan and Chitta Baral.
\newblock {A Bayesian Approach for Medical Inquiry and Disease Inference in
  Automated Differential Diagnosis}.
\newblock \emph{arXiv preprint arXiv:2110.08393}, 2021.

\bibitem[Guyatt et~al.(2002)Guyatt, Rennie, Meade, Cook,
  et~al.]{guyatt2002users}
Gordon Guyatt, Drummond Rennie, Maureen Meade, Deborah Cook, et~al.
\newblock \emph{Users' guides to the medical literature: a manual for
  evidence-based clinical practice}, volume 706.
\newblock AMA press Chicago, 2002.

\bibitem[Henderson et~al.(2012)Henderson, Tierney, and
  Smetana]{henderson2012patient}
Mark Henderson, Lawrence~M Tierney, and Gerald~W Smetana.
\newblock \emph{The patient history: Evidence-based approach}.
\newblock McGraw Hill Professional, 2012.

\bibitem[Hessel et~al.(2018)Hessel, Modayil, Van~Hasselt, Schaul, Ostrovski,
  Dabney, Horgan, Piot, Azar, and Silver]{hessel2018rainbow}
Matteo Hessel, Joseph Modayil, Hado Van~Hasselt, Tom Schaul, Georg Ostrovski,
  Will Dabney, Dan Horgan, Bilal Piot, Mohammad Azar, and David Silver.
\newblock Rainbow: Combining improvements in deep reinforcement learning.
\newblock In \emph{Proceedings of the AAAI Conference on Artificial
  Intelligence}, volume~32, 2018.

\bibitem[Janisch et~al.(2019)Janisch, Pevn{\`y}, and
  Lis{\`y}]{janisch2019classification}
Jarom{\'\i}r Janisch, Tom{\'a}{\v{s}} Pevn{\`y}, and Viliam Lis{\`y}.
\newblock Classification with costly features using deep reinforcement
  learning.
\newblock In \emph{Proceedings of the AAAI Conference on Artificial
  Intelligence}, volume~33, pages 3959--3966, 2019.

\bibitem[Kachuee et~al.(2019)Kachuee, Goldstein, K{\"a}rkk{\"a}inen, Darabi,
  and Sarrafzadeh]{kachuee2019opportunistic}
Mohammad Kachuee, Orpaz Goldstein, Kimmo K{\"a}rkk{\"a}inen, Sajad Darabi, and
  Majid Sarrafzadeh.
\newblock Opportunistic learning: Budgeted cost-sensitive learning from data
  streams.
\newblock In \emph{Proceedings of the International Conference on Learning
  Representations}, 2019.

\bibitem[Kao et~al.(2018)Kao, Tang, and Chang]{kao2018context}
Hao-Cheng Kao, Kai-Fu Tang, and Edward Chang.
\newblock Context-aware symptom checking for disease diagnosis using
  hierarchical reinforcement learning.
\newblock In \emph{Proceedings of the AAAI Conference on Artificial
  Intelligence}, volume~32, 2018.

\bibitem[Kichloo et~al.(2020)Kichloo, Albosta, Dettloff, Wani, El-Amir, Singh,
  Aljadah, Chakinala, Kanugula, Solanki, et~al.]{kichloo2020telemedicine}
Asim Kichloo, Michael Albosta, Kirk Dettloff, Farah Wani, Zain El-Amir, Jagmeet
  Singh, Michael Aljadah, Raja~Chandra Chakinala, Ashok~Kumar Kanugula,
  Shantanu Solanki, et~al.
\newblock Telemedicine, the current {COVID-19} pandemic and the future: a
  narrative review and perspectives moving forward in the {USA}.
\newblock \emph{Family Medicine and Community Health}, 8\penalty0 (3), 2020.

\bibitem[Kingma and Ba(2015)]{KingmaB14Adam}
Diederik~P. Kingma and Jimmy Ba.
\newblock {Adam: A Method for Stochastic Optimization}.
\newblock In \emph{Proceedings of the 3rd International Conference on Learning
  Representations, {ICLR} 2015, San Diego, CA, USA, May 7-9}, 2015.

\bibitem[Liu et~al.(2022)Liu, Cheng, Wang, Tangi, Liu, Zhao, Li, Zheng, and
  Liang]{liu2022my}
Wenge Liu, Yi~Cheng, Hao Wang, Jianheng Tangi, Yafei Liu, Ruihui Zhao, Wenjie
  Li, Yefeng Zheng, and Xiaodan Liang.
\newblock "{M}y nose is running. {Are} you also coughing?": Building a medical
  diagnosis agent with interpretable inquiry logics.
\newblock In \emph{Proceedings of the Thirty-First International Joint
  Conference on Artificial Intelligence}, pages 4266--4272, 2022.

\bibitem[Luo et~al.(2020)Luo, Li, and Glass]{Luo2020KnowledgeGC}
Hongyin Luo, Shang-Wen Li, and James~R. Glass.
\newblock Knowledge grounded conversational symptom detection with graph memory
  networks.
\newblock In \emph{Proceedings of the Clinical Natural Language Processing
  Workshop, ClinicalNLP}, 2020.

\bibitem[Mansoor(2018)]{mansoor2018frameworks}
Andre Mansoor.
\newblock \emph{{Frameworks for Internal Medicine}}.
\newblock Lippincott Williams \& Wilkins, 2018.

\bibitem[Miller et~al.(1986)Miller, Masarie, and Myers]{milner86QMR}
R~Miller, FE~Masarie, and JD~Myers.
\newblock {Quick Medical Reference (QMR) for Diagnostic Assistance}.
\newblock \emph{M.D. Computing : Computers in Medical Practice}, 3:\penalty0
  34--48, 1986.

\bibitem[Mnih et~al.(2015)Mnih, Kavukcuoglu, Silver, Rusu, Veness, Bellemare,
  Graves, Riedmiller, Fidjeland, Ostrovski, et~al.]{mnih2015human}
Volodymyr Mnih, Koray Kavukcuoglu, David Silver, Andrei~A Rusu, Joel Veness,
  Marc~G Bellemare, Alex Graves, Martin Riedmiller, Andreas~K Fidjeland, Georg
  Ostrovski, et~al.
\newblock Human-level control through deep reinforcement learning.
\newblock \emph{Nature}, 518\penalty0 (7540):\penalty0 529--533, 2015.

\bibitem[Ng et~al.(1999)Ng, Harada, and Russell]{ng1999policy}
Andrew~Y Ng, Daishi Harada, and Stuart Russell.
\newblock Policy invariance under reward transformations: Theory and
  application to reward shaping.
\newblock In \emph{Proceedings of the International Conference on Machine
  Learning}, volume~99, pages 278--287, 1999.

\bibitem[Nichol et~al.(2018)Nichol, Sundjaja, and Nelson]{nichol2018medical}
Jonathan~R Nichol, Joshua~Henrina Sundjaja, and Grant Nelson.
\newblock {Medical History}.
\newblock \emph{StatPearls}, 2018.

\bibitem[Peng et~al.(2018)Peng, Tang, Lin, and Chang]{peng2018refuel}
Yu-Shao Peng, Kai-Fu Tang, Hsuan-Tien Lin, and Edward Chang.
\newblock {REFUEL}: Exploring sparse features in deep reinforcement learning
  for fast disease diagnosis.
\newblock \emph{Advances in Neural Information Processing Systems},
  31:\penalty0 7322--7331, 2018.

\bibitem[Ramanayake and Basnayake(2018)]{Ramanayake2018EvaluationOR}
R.P.J.C. Ramanayake and Basnayake Mudiyanselage Duminda~Bandara Basnayake.
\newblock Evaluation of red flags minimizes missing serious diseases in primary
  care.
\newblock \emph{Journal of Family Medicine and Primary Care}, 7:\penalty0 315
  -- 318, 2018.

\bibitem[Rhoads et~al.(2017)Rhoads, Penick, et~al.]{rhoads2017formulating}
Jacqueline Rhoads, Julie~C Penick, et~al.
\newblock \emph{Formulating a Differential Diagnosis for the Advanced Practice
  Provider}.
\newblock Springer Publishing Company, 2017.

\bibitem[Richardson and Wilson(2015)]{richardson2015the}
W.~Scott Richardson and Mark~C. Wilson.
\newblock {The Process of Diagnosis}.
\newblock In Gordon Guyatt, Drummond Rennie, Maureen~O. Meade, and Deborah~J.
  Cook, editors, \emph{Users' Guides to the Medical Literature: A Manual for
  Evidence-Based Clinical Practice, 3rd edition}. McGraw-Hill Education, New
  York, NY, 2015.

\bibitem[Richens et~al.(2020)Richens, Lee, and Johri]{richens2020improving}
Jonathan~G Richens, Ciar{\'a}n~M Lee, and Saurabh Johri.
\newblock Improving the accuracy of medical diagnosis with causal machine
  learning.
\newblock \emph{Nature communications}, 11\penalty0 (1):\penalty0 1--9, 2020.

\bibitem[Stooke and Abbeel(2019)]{Stooke19Rlpyt}
Adam Stooke and Pieter Abbeel.
\newblock {rlpyt: A Research Code Base for Deep Reinforcement Learning in
  PyTorch}.
\newblock \emph{arXiv preprint arXiv:1909.01500}, 2019.

\bibitem[Sutton et~al.(1999)Sutton, Precup, and Singh]{sutton1999between}
Richard~S Sutton, Doina Precup, and Satinder Singh.
\newblock Between {MDP}s and semi-{MDP}s: A framework for temporal abstraction
  in reinforcement learning.
\newblock \emph{Artificial intelligence}, 112\penalty0 (1-2):\penalty0
  181--211, 1999.

\bibitem[Wei et~al.(2018)Wei, Liu, Peng, Tou, Chen, Huang, Wong, and
  Dai]{wei2018task}
Zhongyu Wei, Qianlong Liu, Baolin Peng, Huaixiao Tou, Ting Chen, Xuanjing
  Huang, Kam-fai Wong, and Xiangying Dai.
\newblock Task-oriented dialogue system for automatic diagnosis.
\newblock In \emph{Proceedings of the 56th Annual Meeting of the Association
  for Computational Linguistics (Volume 2: Short Papers)}, pages 201--207,
  Melbourne, Australia, July 2018. Association for Computational Linguistics.
\newblock \doi{10.18653/v1/P18-2033}.
\newblock URL \url{https://aclanthology.org/P18-2033}.

\bibitem[Wiewiora et~al.(2003)Wiewiora, Cottrell, and
  Elkan]{wiewiora2003principled}
Eric Wiewiora, Garrison~W Cottrell, and Charles Elkan.
\newblock Principled methods for advising reinforcement learning agents.
\newblock In \emph{Proceedings of the 20th International Conference on Machine
  Learning}, pages 792--799, 2003.

\bibitem[Williams(1992)]{William12Reinforce}
Ronald~J. Williams.
\newblock {Simple Statistical Gradient-Following Algorithms for Connectionist
  Reinforcement Learning}.
\newblock \emph{Machine Learning}, 8\penalty0 (3–4):\penalty0 229–256, May
  1992.

\bibitem[Xiao et~al.(2018)Xiao, Choi, and Sun]{xiao2018opportunities}
Cao Xiao, Edward Choi, and Jimeng Sun.
\newblock Opportunities and challenges in developing deep learning models using
  electronic health records data: a systematic review.
\newblock \emph{Journal of the American Medical Informatics Association},
  25\penalty0 (10):\penalty0 1419--1428, 2018.

\bibitem[Xu et~al.(2019)Xu, Zhou, Gong, Liang, Tang, and Lin]{xu2019end}
Lin Xu, Qixian Zhou, Ke~Gong, Xiaodan Liang, Jianheng Tang, and Liang Lin.
\newblock End-to-end knowledge-routed relational dialogue system for automatic
  diagnosis.
\newblock In \emph{Proceedings of the Thirty-Third AAAI Conference on
  Artificial Intelligence and Thirty-First Innovative Applications of
  Artificial Intelligence Conference and Ninth AAAI Symposium on Educational
  Advances in Artificial Intelligence}, AAAI'19/IAAI'19/EAAI'19. AAAI Press,
  2019.
\newblock ISBN 978-1-57735-809-1.
\newblock \doi{10.1609/aaai.v33i01.33017346}.
\newblock URL \url{https://doi.org/10.1609/aaai.v33i01.33017346}.

\bibitem[Yuan and Yu(2021)]{yuan2021efficient}
Hongyi Yuan and Sheng Yu.
\newblock {Efficient Symptom Inquiring and Diagnosis via Adaptive Alignment of
  Reinforcement Learning and Classification}.
\newblock \emph{arXiv preprint arXiv:2112.00733}, 2021.

\bibitem[Zhao et~al.(2021)Zhao, Chen, and Chen]{zhao2021weighted}
Xinyan Zhao, Liangwei Chen, and Huanhuan Chen.
\newblock {A Weighted Heterogeneous Graph-Based Dialog System}.
\newblock \emph{IEEE Transactions on Neural Networks and Learning Systems},
  2021.

\end{thebibliography}






\clearpage
\section*{Checklist}


\begin{enumerate}

\item For all authors...
\begin{enumerate}
  \item Do the main claims made in the abstract and introduction accurately reflect the paper's contributions and scope?
    \answerYes{See Sections~\ref{s:method},~\ref{s:experiments},~\ref{s:results}.}
  \item Did you describe the limitations of your work?
    \answerYes{See Section~\ref{s:limitation}.}
  \item Did you discuss any potential negative societal impacts of your work?
    \answerYes{See Section~\ref{s:limitation}.}
  \item Have you read the ethics review guidelines and ensured that your paper conforms to them?
    \answerYes{See Section~\ref{s:limitation}.}
\end{enumerate}

\item If you are including theoretical results...
\begin{enumerate}
  \item Did you state the full set of assumptions of all theoretical results?
    \answerNA{}
        \item Did you include complete proofs of all theoretical results?
    \answerNA{}
\end{enumerate}

\item If you ran experiments...
\begin{enumerate}
  \item Did you include the code, data, and instructions needed to reproduce the main experimental results (either in the supplemental material or as a URL)?
    \answerYes{the code is provided in the supplementary and it contains a readme with instructions needed to reproduce the main experimental results.}
  \item Did you specify all the training details (e.g., data splits, hyperparameters, how they were chosen)?
    \answerYes{Data splits are described in Section~\ref{ss:dataset}. Other algorithm-related and training details are described in Appendix \ref{appx:casande_details}, \ref{appx:paper_baselines} \ref{app:training_details}.}
        \item Did you report error bars (e.g., with respect to the random seed after running experiments multiple times)?
    \answerYes{See Section~\ref{ss:results_prediction}}
        \item Did you include the total amount of compute and the type of resources used (e.g., type of GPUs, internal cluster, or cloud provider)?
    \answerYes{See Appendix \ref{app:training_details}.}
\end{enumerate}

\item If you are using existing assets (e.g., code, data, models) or curating/releasing new assets...
\begin{enumerate}
  \item If your work uses existing assets, did you cite the creators?
    \answerYes{See Section~\ref{s:experiments}}.
  \item Did you mention the license of the assets?
    \answerYes{The license for the code is attached with the code in the supplementary.}
  \item Did you include any new assets either in the supplemental material or as a URL?
    \answerYes{We include code in the supplementary.}
  \item Did you discuss whether and how consent was obtained from people whose data you're using/curating?
    \answerYes{The data used in this project is publicly available under the CC-BY licence.}
  \item Did you discuss whether the data you are using/curating contains personally identifiable information or offensive content?
    \answerYes{The data is synthetic and does not contain personally identifiable information (See Section~\ref{ss:dataset}).}
\end{enumerate}

\item If you used crowdsourcing or conducted research with human subjects...
\begin{enumerate}
  \item Did you include the full text of instructions given to participants and screenshots, if applicable?
    \answerNA{}
  \item Did you describe any potential participant risks, with links to Institutional Review Board (IRB) approvals, if applicable?
    \answerNA{}
  \item Did you include the estimated hourly wage paid to participants and the total amount spent on participant compensation?
    \answerNA{}
\end{enumerate}

\end{enumerate}


\appendix
\normalsize
\section{CASANDE details}
\label{appx:casande_details}

CASANDE's code can be found at \url{https://github.com/mila-iqia/Casande-RL}.

\subsection{Reward shaping schedulers}\label{appx:paper_scheduler}
\begin{figure}[ht]
\centering
\begin{subfigure}[b]{.5\textwidth}
  \centering
  \includegraphics[width=.98\linewidth]{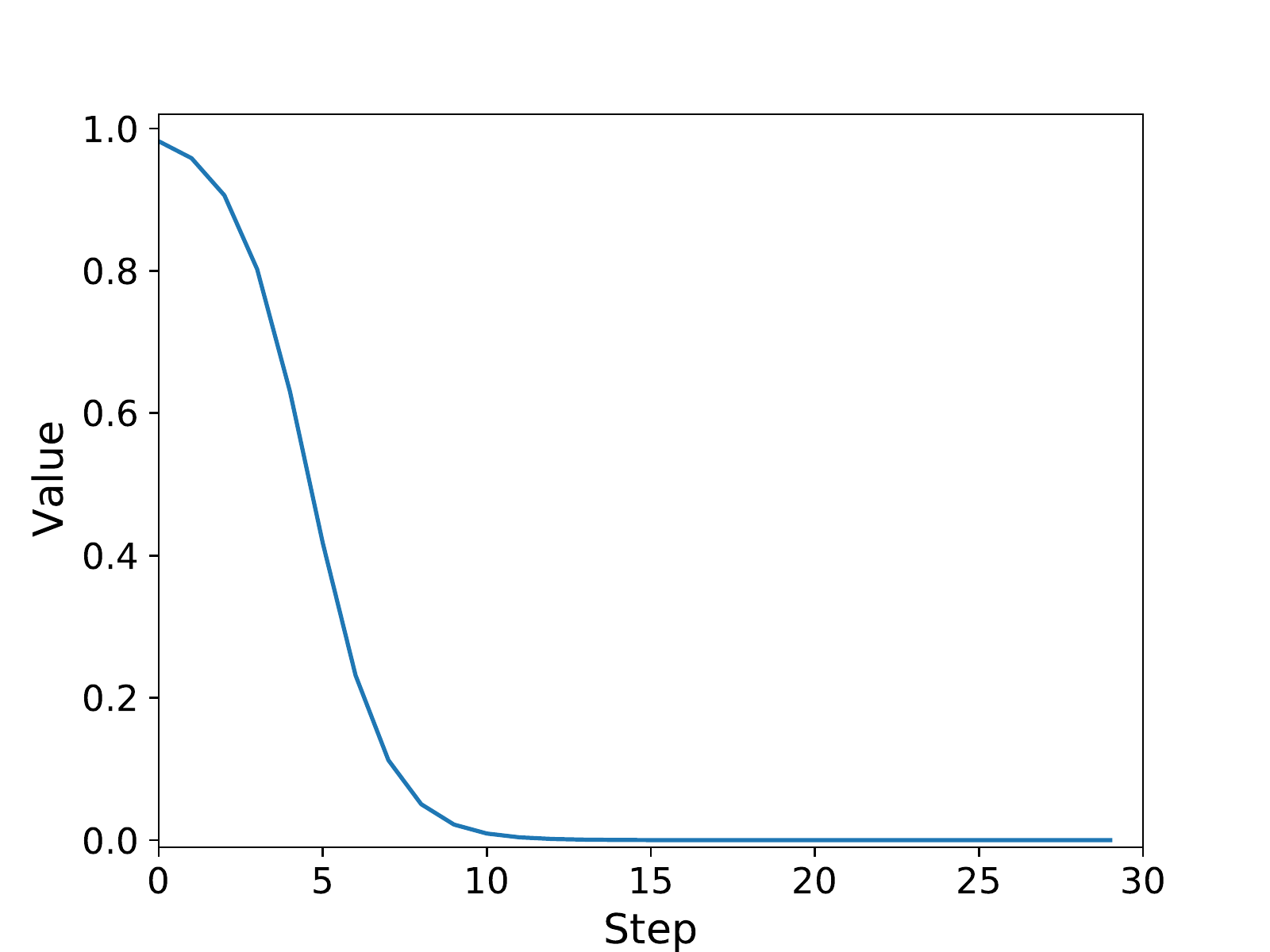}
  \caption{$w_{\text{Ex}}(t)$}
  \label{fig:paper_wdiv}
\end{subfigure}%
\begin{subfigure}[b]{.5\textwidth}
  \centering
  \includegraphics[width=.98\linewidth]{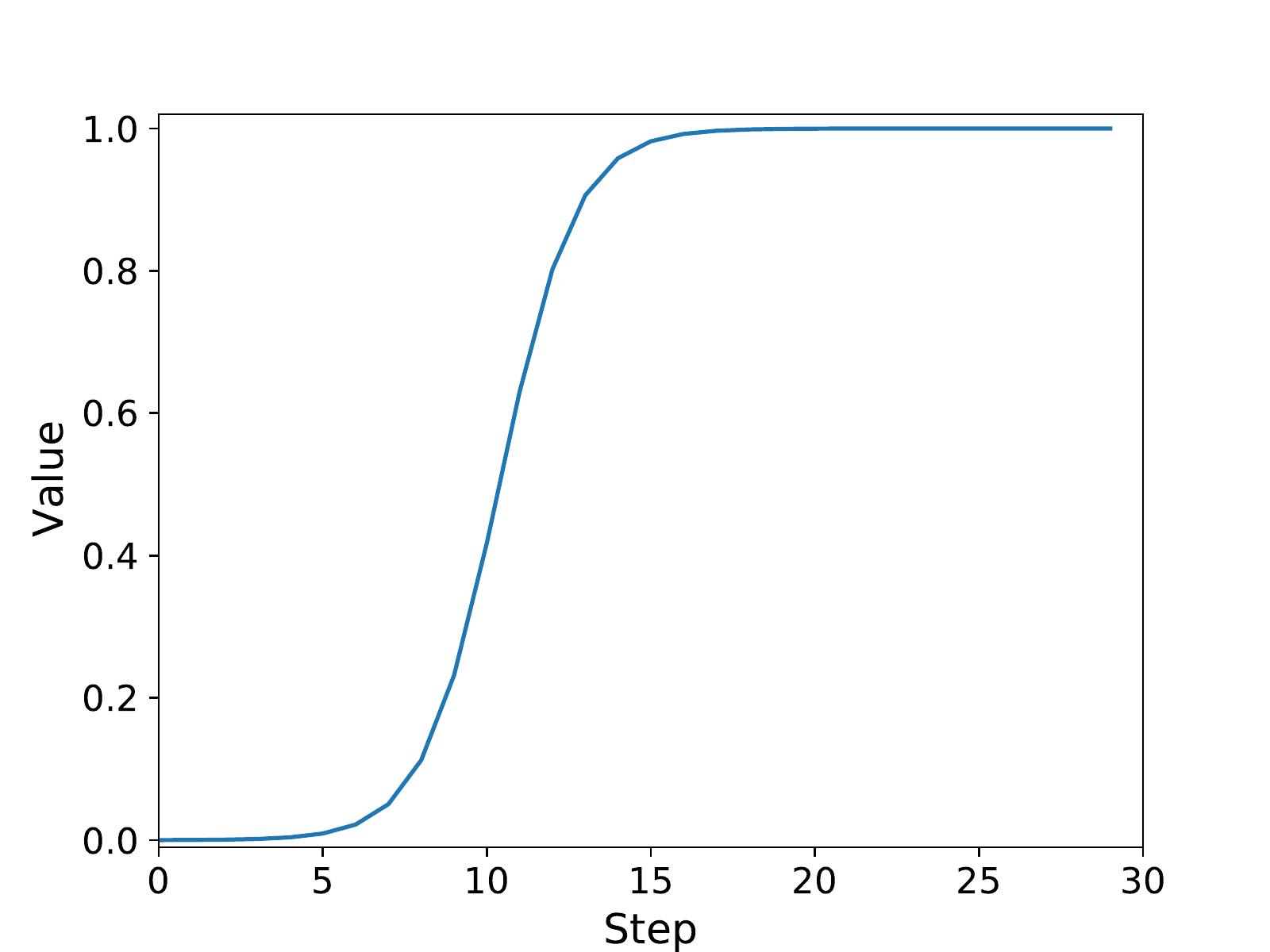}
  \caption{$w_{\text{Co}}(t)$}
  \label{fig:paper_wenf}
\end{subfigure}
\caption{Examples of the time-dependent weights $w_{\text{Ex}}(t)$ and $w_{\text{Co}}(t)$ that control the importance of the exploration and confirmation reward components. Here, $\bar{x}_{min} = -13$, $\bar{x}_{max} = 13$, $T=30$, $\Delta_{\text{Ex}} = 9$, and $\Delta_{\text{Co}} = 4$.}
\label{fig:paper_scheduler}
\end{figure}

 We introduced time-dependent coefficients $w_{\text{Ex}}(t)$ and $w_{\text{Co}}(t)$ to control the importance of the exploration and confirmation reward components through time (see Equations~\ref{eq:paper_div} and~\ref{eq:paper_enf}). Figure~\ref{fig:paper_scheduler} shows an example of those coefficients, based on some fixed parameters.

\subsection{Algorithm}\label{appx:paper_algo}
The two branches of our agent are updated alternatively using the loss functions introduced in Equations~\ref{eq:DQNlossComponent} and~\ref{eq:ClassifierlossComponent}. Let $\theta$ and $\phi$ be the agent and target network parameters. Let $\pi_{\theta}$ be the agent policy characterized by $\theta$. And let $\rho$ be the target network soft update rate. Then, Algorithm~\ref{algo:paper_training} depicts a pseudo-code of the process used to train the agent.

\begin{algorithm}[htb!]
\SetAlgoLined
 Randomly initialize $\theta$ and $\phi$\;
 Initialize the set $\mathbf{E}$ of all the environments\;
 Initialize the replay buffer $\mathbf{Mem}$\;
 \While{not converged}{
  \For{$env \in \mathbf{E}$}{
  Simulate one step with $\epsilon$-greedy policy based on $\pi_{\theta}$\;
  Add the resulting transition in the replay buffer $\mathbf{Mem}$\;
  }
  Sample a batch $B$ from the replay buffer $\mathbf{Mem}$\;
  Compute the policy loss $Loss_{Q}(B)$ and update $\theta$\;
  Compute the classifier loss $Loss_{C}(B)$ and update $\theta$\;
  Update parameters $\phi \leftarrow (1-\rho) \phi + \rho \theta$\;
 }
 \caption{Training Algorithm}
 \label{algo:paper_training}
\end{algorithm}

\subsection{Input data representation}\label{appx:exp_input_data}
In this section, we describe how we encode input data to handle different types of evidences, namely binary, categorical, and multi-choice evidences from the \CasandeDDx dataset.

Let $u_e$ be the dimension induced by the $e^{th}$ evidence in the state space. We have $u_e$ equals $1$ for binary evidences or numerical categorical evidences (i.e., categorical evidences whose options are numbers), while $u_e$ corresponds to the number of available options $n_e$ for  categorical evidences that are not numerical or for multi-choice evidences. Let $w$ be the dimension induced by socio-demographic data such as the age and the sex of the patient. We assume in our state representation that the socio-demographic data are encoded before the evidences. Let $v_e = w +  \sum\limits_{t=1}^{e-1}u_t$ be the cumulative dimension induced by the set of evidences indexed before the $e^{th}$ evidence, with $v_1 = w$. Let $f_{\text{Pres}}(e)$ be the function which returns the evidence  $e$'s options that are experienced by the underlying patient, or, in case $e$ is binary, whether or not it is experienced by the patient. The evidence $e$ is then encoded as follows in the state  $\mathbf{s}_{t}$ based on its type:
    \begin{itemize}
        \item \textbf{Binary evidences}
        \begin{equation*}
          {\mathbf{s}_{t, v_e + 1}} =
          \begin{cases}
          0 & \text{if $e$ is not yet inquired},\\
          1 & \text{if $f_{\text{Pres}}(e)$,}\\
          -1 & \text{otherwise.}
        \end{cases}       
       \end{equation*}
        \item \textbf{Numerical categorical evidences}
        \begin{equation*}
          {\mathbf{s}_{t, v_e + 1}} =
          \begin{cases}
          0 & \text{if $e$ is not yet inquired},\\
          \frac{f_{\text{Pres}}(e) + 1}{M_e + 1} & \text{otherwise, where $M_e$ is the maximum value for evidence $e$.}
        \end{cases}       
       \end{equation*}
       \item \textbf{Non-numerical categorical evidences}
        \begin{equation*}
          \forall j \in [1, u_e], \; {\mathbf{s}_{t, v_e + j}} =
          \begin{cases}
          0 & \text{$e$ is not yet inquired,}\\
          1 & \text{if $j \in f_{\text{Pres}}(e)$,}\\
          -1 & \text{otherwise.}
        \end{cases}       
       \end{equation*}
       \item \textbf{Multi-choice evidences}
        \begin{equation*}
          \forall j \in [1, u_e], \; {\mathbf{s}_{t, v_e + j}} =
          \begin{cases}
          0 & \text{$e$ is not yet inquired,}\\
          1 & \text{if $j \in f_{\text{Pres}}(e)$,}\\
          -1 & \text{otherwise.}
        \end{cases}       
       \end{equation*}
    \end{itemize}

\section{Baseline details}\label{appx:paper_baselines}

\subsection{BASD}\label{appx:paper_baseline_asd}

BASD (short for Baseline ASD) is inspired by the work done by \citet{Luo2020KnowledgeGC}. In that work, the authors propose to build an Automatic Symptom Detection (ASD) module to collect evidences from patients using supervised learning while leveraging a knowledge graph which encodes relations among symptoms and diseases. We follow the setup introduced in \citep{Fansi_Tchango_DDXPlus_A_new_2022}. More specifically, we attach to the evidence acquisition module a classifier network whose goal is to predict the patient's underlying disease at the end of the acquisition process based on the collected evidences. More specifically, the BASD agent consists of an MLP network with 2 prediction branches:
\begin{itemize}
    \item a policy branch whose role is to predict whether to stop or continue the interaction, and if the latter, which evidence to inquire about next;
    \item a classifier branch whose role is to predict the underlying patient disease.
\end{itemize}
The knowledge graph is not used in BASD unlike the work done by \citet{Luo2020KnowledgeGC}.

To train the network, we simulate dialogue states together with their target values. In other words, let us assume a given patient has $n$ evidences that he/she is experiencing. We simulate a dialogue state as follows:
\begin{enumerate}
    \item Randomly select $p \in [1, n]$ representing the number of positive evidences already acquired. Sample $p$ evidences from the ones experienced by the patient and set them in the simulated dialog state.
    \item Randomly select $q \in [0, T - p)$ representing the number of negative evidences already inquired where T is the maximum number of allowed dialog turns. Sample $q$ evidences from the ones not experienced by the patient and set them in the simulated dialog state.
    \item If $p = n$, set the target of the policy branch to "stop"; otherwise set the target to be one of the experienced evidences that was not sampled at step 1).
    \item Set the classifier branch target to be the ground truth pathology.
\end{enumerate}

Both branches are trained using the cross-entropy loss function and  the classifier branch is only updated when the target of the policy branch is set to "stop". We use the same input data representation described in Appendix~\ref{appx:exp_input_data} for this baseline.

\subsection{Diaformer}

Diaformer \citep{chen2021diaformer} is a recent Transformer-based model that models the symptom acquisition process as a generation task. It is a supervised model, and its training objective is similar to BERT's Masked Language Modelling (MLM) \citep{devlin2019bert}. The model is trained to maximize the likelihood of synthesized trajectories, each consisting of a patient's disease, initial complaints (called explicit evidences), and evidences that need to be collected through the interaction with the patient (called implicit evidences). At test time, the model is provided with explicit evidences only, and needs to iteratively inquire about implicit ones, until it decides to end the interaction and predict the disease.

\begin{figure}
    \centering
    \includegraphics[width=0.95\textwidth]{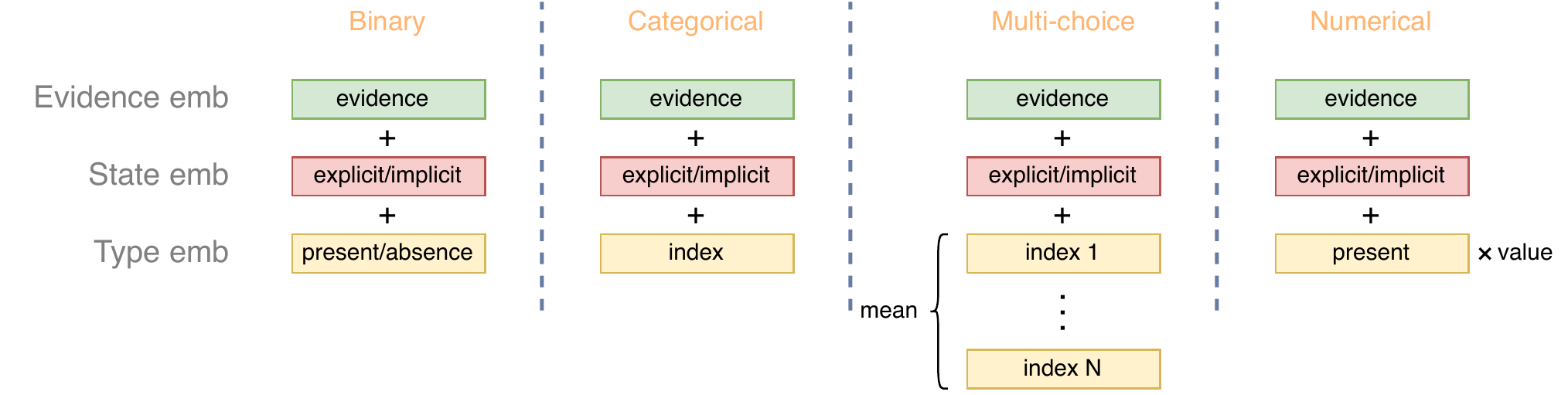}
    \caption{Illustration of Diaformer's input representations}
    \label{fig:diaformer_embs}
\end{figure}

Diaformer represents each evidence with three types of embeddings, which respectively indicate the \textit{evidence Id}, the evidence's \textit{state} (i.e., whether it is explicit or implicit), and the evidence's \textit{type} (i.e., whether it is present or not). All embeddings are learnable, and the overall evidence's representation is the sum of its three embeddings. To represent the data in the \CasandeDDx dataset, in particular categorical, multi-choice, and numerical evidences, we modify Diaformer's input representations such that different types of evidences are represented differently. Figure \ref{fig:diaformer_embs} illustrates the modified input representations. Specifically, we  define a learnable embedding for each possible option defined for non-binary evidences. Then, for categorical evidences, we use the embedding corresponding to the evidence option experienced by the underlying patient as the type embedding. Similarly, for multi-choice evidences, we take the average of the embeddings corresponding to all the options experienced by the patient for that evidence. Finally, for numerical evidences, we multiply the embedding corresponding to the ``present" type with the numerical evidence value.

\subsection{BED}

BED \citep{guan2021bayesian} is an approach that does not need training and which exploits prior knowledge regarding evidence-disease relationships (i.e., the Quick Medical Reference belief network) to decide which evidence to inquire about next and to update its disease prediction. More specifically, the objective of BED at each turn is to inquire about the evidence $e$ that maximizes a utility function $U(e)$ defined for binary evidences as:
\begin{equation}
    U(e) = \sum\limits_{v \in \{0, 1\}} p(e=v | \mathbf{e}^{+}, \mathbf{e}^{-}) \times D_{\text{KL}}\left(p(d | \mathbf{e}^{+}, \mathbf{e}^{-}, e=v) || p(d | \mathbf{e}^{+}, \mathbf{e}^{-})\right).
    \label{eq:BedUtilityBinary}
\end{equation}
$\mathbf{e}^{+}$ is the set of already inquired evidences which are experienced by the patient. Similarly, $\mathbf{e}^{-}$ is the set of already inquired evidences which are not experienced by the patient. $p(e=v | \mathbf{e}^{+}, \mathbf{e}^{-})$ captures the probability that the outcome of inquiry about $e$ results in $v$ given the sets $\mathbf{e}^{+}$ and $\mathbf{e}^{-}$. Finally, $p(d | \mathbf{e})$ represents the probability of the diseases given the set of inquired evidences $\mathbf{e}$. The inquiry process ends when the utility of all the remaining evidences is below a predefined threshold.

To be able to use BED on the \CasandeDDx dataset, we extend the notion of utility function on categorical and multi-choice evidences.

For categorical evidences, the extension is easy and we simply have:
\begin{equation}
    U(e) = \sum\limits_{v \in \mathcal{C}_e} p(e=v | \mathbf{e}^{+}, \mathbf{e}^{-}) \times D_{\text{KL}}\left(p(d | \mathbf{e}^{+}, \mathbf{e}^{-}, e=v) || p(d | \mathbf{e}^{+}, \mathbf{e}^{-})\right),
    \label{eq:BedUtilityCategorical}
\end{equation}
where $\mathcal{C}_e$ is the set of possible options for the categorical evidence $e$.

For multi-choice evidences, since each option can independently be experienced by the patient, we treat each such option as a binary evidence and define $U(e)$ as the maximum of the utility of the different options. Thus, we have
\begin{equation}
    U(e) = \max\limits_{v \in \mathcal{C}_e} U(v),
    \label{eq:BedUtilityMultiChoice}
\end{equation}
where $\mathcal{C}_e$ is the set of possible options for the multi-choice evidence $e$ and $U(v)$ is defined according to Equation~\ref{eq:BedUtilityBinary}. We chose to use the maximum value instead of the sum of all values to keep the magnitude of the utility values comparable among the different evidence types. The mean value wasn't considered as it might have diluted the impact of options with high utility value.  

\subsection{AARLC}

AARLC \citep{yuan2021efficient} is an RL-based approach consisting of two branches, just like our approach, which proposes an adaptive method to align the tasks performed by the two branches using the entropy of the distributions predicted by the classifier branch. To use this baseline with the \CasandeDDx dataset, we use the input data representation described in Appendix~\ref{appx:exp_input_data}.

To train AARLC with differentials, we make several changes, in addition to replacing the ground truth pathology with the ground truth differential probabilities as the classifier's training objective: 

\begin{itemize}
    \item Instead of updating the stopping threshold $K$ when the predicted pathology matches the ground truth pathology, we update it when the set $D^i$ of diseases in the ground truth differential is identical to the set of top-$|D^i|$ predicted diseases. We make this change because AARLC designs the threshold to be updated when the predicted disease is correct, and therefore if the differential is replacing the ground truth pathology as the target, it should also replace it as the standard of correctness.
    \item Second, now that the agent does not focus on predicting a single disease, it is no longer reasonable to only update the threshold associated with one disease. Therefore, we instead use one global threshold that is not associated with any particular disease, and update it every time the aforementioned condition is met.
    \item Similar to the condition of updating the threshold, we change the condition under which a positive reward is given to the agent, as part of $r_p$, for making the correct diagnosis. We give the positive reward when the set $D^i$ is identical to the set of top-$|D^i|$ predicted diseases.
\end{itemize}

There are several differences between AARLC and CASANDE:
\begin{itemize}
\item CASANDE has the exploration, confirmation and severity reward functions that integrate in the training process elements that mimic the reasoning of doctors. Those functions do not exist in AARLC.
\item AARLC uses separate models for the classifier and policy networks while in CASANDE, those 2 networks share the same encoder and exchange information.
\item The classifier network is updated at each interaction turn in AARLC. In CASANDE, it is only updated at the end of the interaction as this is when the differential diagnosis is needed and can be accurately predicted. Forcing the classifier to predict the ground truth differential at each turn can confuse the classifier, in particular at the start of the interaction, when the number of collected evidences is small.
\item AARLC provides a small positive reward when the agent asks questions about evidence that the patient doesn’t experience. CASANDE doesn’t because it isn’t clear when questions about negative evidence are useful: some questions are useful because they can help rule out pathologies from the differential; others are not informative. It might be worth exploring in future work a reward about negative evidence that is based on the impact of this information on the differential.
\item AARLC has the $r_H$ reward which encourages the model to reduce the entropy of the differential from one interaction turn to the next, as more information is accumulated. In CASANDE, we use another strategy (in the consolidation phase), which consists in ensuring that the differential at the next interaction turn is closer to the ground truth differential than at the previous turn. We therefore consider that it is not sufficient to reduce the entropy of the differential if the differential is not being pushed in the right direction.
\item AARLC’s policy network doesn’t predict an \textit{exit} action; instead, AARLC compares the entropy of the differential at each interaction turn to a learnable threshold and stops the interaction when the entropy is smaller than the threshold, irrespective of whether the differential is correct. CASANDE’s policy network can directly predict the \textit{exit} action, and its loss (Equation \ref{eq:DQNlossComponent}) provides feedback to the network at each interaction turn about the value of predicting this action; it also has a reward at the end of the interaction that depends on the quality of the predicted differential diagnosis.
\end{itemize}

\section{Training details}\label{app:training_details}


The cluster used for training purposes is a mixture of NVIDIA A100, K80, M40, RTX 8000, TITAN RTX and V100 GPUs. Except for BED which does not require training, each training session is conducted with one GPU allocated by the scheduler of the cluster.

\subsection{CASANDE}

We rely on the rlpyt framework \citep{Stooke19Rlpyt} and use the Rainbow algorithm \citep{hessel2018rainbow} to train our agent. We allow interactions to have a maximum of $T = 30$ turns and set $\gamma$ to 0.99. For the reward, we set $r_i$ to 0.5, $r_p$ to 2, and $r_n$ to 0. This means that we do not penalize the agent when it asks about evidence not being experienced by the patient as it is sometimes necessary to inquire about such evidence; given that we do not know which negative evidence is pertinent, we neither penalize nor reward the agent when inquiring about such evidence. We use 16 environment instances to collect data during training. We use Adam \citep{KingmaB14Adam} as an optimizer with a learning rate of $6.25 \times 10^{-5}$. $[\bar{x}_{min}, \bar{x}_{max}]$ is set to $[-13, 13]$ whereas $\Delta_{\text{Ex}}$ and $\Delta_{\text{Co}}$ are respectively set to 9 and 4, unless stated otherwise. The probability threshold used to decide if a pathology is part of a differential when computing $SevOut_t$, $SevIn_t$ and $Sev_y$ is 0.01. We perform hyper-parameter tuning and the following values were selected based on the performance on the validation dataset: $w_{{\text{si}}} = 1$, $\alpha_{{\text{Ex}}} = 12$, $\alpha_{{\text{Co}}} = 1$, $\alpha_{{\text{Sev}}} = 0.75$, and $\alpha_{{\text{Cl}}} = 1$.

The architectural details of our agent together with the parameters of the Rainbow algorithm are described in Table~\ref{tbl:paper_archiparams}.

\begin{table}[htb!]
\centering
\caption{The details of the network architecture and the parameters of the Rainbow algorithm used in the CASANDE experiments. The numbers in brackets represent MLP layer sizes. For all MLPs, we use the ReLU activation function. The dataset has 49 pathologies and 223 evidences.}
\vspace{1mm}
\small
\begin{tabular}{lr}
\hline
\textbf{Components} & \textbf{Description} \\ \hline
\multicolumn{2}{c}{\textbf{Encoder}}                                                             \\ \hline
MLP                                                                       & [4096, 2048, 2048]     \\ \hline
\multicolumn{2}{c}{\textbf{Classifier}}                                                          \\ \hline
MLP                                                                       & [1024, 512, 49]        \\ \hline
\multicolumn{2}{c}{\textbf{Policy}}                                                              \\ \hline
Number of atoms                                                           & 51                     \\ \hline
\begin{tabular}[c]{@{}l@{}}Dueling Advantage\\ Network (MLP)\end{tabular} & [1024, 512,  223 x 51] \\ \hline
\begin{tabular}[c]{@{}l@{}}Dueling Value\\ Network (MLP)\end{tabular}     & [1024, 512, 51]        \\ \hline
\multicolumn{2}{c}{\textbf{Rainbow Algorithm}}                                                              \\ \hline
Number of atoms                                                           & 51                     \\ \hline
$V_{min}$                                                           & -90                     \\ \hline
$V_{max}$                                                           & 70                     \\ \hline
N-step Q-learning                                                           & 3                     \\ \hline
\end{tabular}
\label{tbl:paper_archiparams}
\end{table}

\subsection{BASD}

The models used are MLPs with hidden layers of size 2048. We use a batch size of $2657$, a patience of $20$, and we tune the number of layers as well as the learning rate. For the model trained to predict the ground truth pathology, the number of hidden layers and the learning rate leading to the optimal validation performance are respectively $2$ and $2.801 \times 10^{-4}$. For the model trained to predict the differential diagnosis, the optimal set of parameters corresponds to a number of hidden layers of $3$ and a learning rate of $3.469 \times 10^{-4}$.

\subsection{Diaformer}

We reuse the same setup as in \citet{chen2021diaformer} for the DX dataset, except for the batch size and the learning rate. We use a batch size of $256$ subject to GPU memory, and we tune the learning rate. 
The optimal learning rate is $1.0136\times 10^{-5}$ for Diaformer trained to predict the ground truth pathology, and $5.0257\times 10^{-4}$ for Diaformer trained to predict the differential.

\subsection{BED}

BED is deterministic and the only parameter to be set in addition to the maximum number of turns is the utility threshold. We follow \citep{guan2021bayesian} and set it to $10^{-2}$.

\subsection{AARLC}

We use the same setup as in \citep{yuan2021efficient} with a batch size of $2657$. We tune the $\nu$ and $\lambda$ parameters together with the learning rate. The optimal set of parameters for the model trained to predict the ground truth pathology is $\left\{ \nu=3.337, \lambda=0.97, \text{lr}=5.175 \times 10^{-4} \right\}$. For the model trained to predict the differential, we obtain $\left\{ \nu=2.826, \lambda=0.99, \text{lr}=3.52 \times 10^{-4} \right\}$.

\section{Evaluation metrics}
\label{appx:metric}

This section describes the metrics used to evaluate the different agents. Let $N$ be the number of patients, $\mathcal{D}$ the set of diseases, and $S \subseteq \mathcal{D}$ be the set of severe pathologies. Also, let $E^{i}$ be the set of evidences (i.e., symptoms and antecedents) experienced by the $i^{th}$ patient, and  $\hat{E}^{i}$ be the set of evidences an agent inquired about when interacting with that patient. Additionally, let $y^i$ be the ground truth differential, $d^{i}$ be the ground truth pathology, and $\hat{y}^i$ be the last pathology distribution (i.e., belief) generated by the agent for that patient. Besides, let $\text{Traj}^i=\{bel_1^i,bel_2^i,\cdots,bel_{|\hat{E}^{i}|}^i = \hat{y}^i\}$ be the predicted differentials made by the agent throughout the interaction process. We further post-process both the ground truth differentials and the predicted ones to remove pathologies whose mass is less than or equal to 0.01. This threshold is selected to reduce the size of the differentials by removing highly unlikely pathologies. Let $\mathcal{D}^{i}$ and $\hat{\mathcal{D}}^i$ be the resulting set of pathologies after applying the post-processing on $y^i$ and $\hat{y}^i$ respectively. In what follows, we use $|.|$ to denote the size of a set.

\subsection{End-performance metrics}
\label{app:end_performance_metrics}

\paragraph{Interaction length (IL)} The average interaction length is defined as:
\begin{equation}
    IL =  \frac{1}{N} \Sigma_{i=1}^{N}|\hat{E}^{i}|.
\end{equation}

\paragraph{Differential diagnosis F1 score (DDF1) } This metric measures how aligned the predicted differential is with respect to the ground truth differential. We first define the differential diagnosis recall (DDR) and precision (DDP) for the $i^{th}$ patient as
\begin{equation}
    DDR^i = \frac{|\hat{\mathcal{D}}^{i} \cap \mathcal{D}^{i}|}{|\mathcal{D}^{i}|},\label{eq:DDRecallP}
\end{equation}
and
\begin{equation}
    DDP^i = \frac{|\hat{\mathcal{D}}^{i} \cap \mathcal{D}^{i}|}{|\hat{\mathcal{D}}^{i}|}.\label{eq:DDPrecisionP}
\end{equation}
Finally, DDF1 is defined as
\begin{equation}
    DDF1 =  \frac{1}{N} \Sigma_{i=1}^{N} DDF^{i},
\end{equation}
where
\begin{equation}
    DDF^i = 2 \frac{DDR^i \times DDP^i}{DDR^i + DDP^i}.\label{eq:DDF1P}
\end{equation}

\paragraph{Harmonic mean score of severe pathologies (DSHM) } This metric corresponds to the harmonic mean of the rule-in and rule-out rates of severe pathologies. We first compute the final rule-in rate and rule-out rate of severe pathologies for the $i^{th}$ patient as
\begin{equation}
    \text{rule-in}^i = \frac{|\hat{\mathcal{D}}^{i} \cap \mathcal{D}^{i} \cap S|}{|\mathcal{D}^{i}\cap S|},\label{eq:DSRecallP}
\end{equation}
and
\begin{equation}
    \text{rule-out}^i = \frac{|S \cap (\mathcal{D}\setminus(\mathcal{D}^{i} \cup \hat{\mathcal{D}}^{i}))|}{|S \cap (\mathcal{D}\setminus \mathcal{D}^{i})|}.\label{eq:DSPrecisionP}
\end{equation}
In other words, the final rule-in rate captures the ratio of severe pathologies that are rightfully included in the final predicted differential while the final rule-out rate measures the ratio of severe pathologies that are rightfully excluded from the final predicted differential.

DSHM is then defined as
\begin{equation}
    DSHM =  \frac{1}{N} \Sigma_{i=1}^{N} DSH^{i},
\end{equation}
where
\begin{equation}
    DSH^i = 2 \frac{\text{rule-in}^i \times \text{rule-out}^i}{\text{rule-in}^i + \text{rule-out}^i}.\label{eq:DSF1P}
\end{equation}

\paragraph{Ground truth pathology accuracy at $k$ (GTPA@k):} This metric measures whether the top $k$ entries of the differential diagnosis predicted by an agent contain the patient's ground truth pathology:
\begin{equation}
    GTPA@k = \frac{1}{N}\Sigma_{i=1}^{N} \mathds{1}_{d^{i} \in \hat{D}_k^{i}},\label{eq:GTPA@k}
\end{equation}
where $\hat{D}_k^{i}$ is the set of top-$k$ pathologies extracted from $\hat{y}^i$. 

\paragraph{Ground truth pathology accuracy (GTPA):} This metric measures whether the patient's ground truth pathology is part of the differential diagnosis predicted by an agent:
\begin{equation}
    GTPA = \frac{1}{N}\Sigma_{i=1}^{N} \mathds{1}_{d^{i} \in \hat{\mathcal{D}}^i}.\label{eq:GTPA}
\end{equation} 

\paragraph{Positive evidence recall (PER)} The average recall of the evidences experienced by patients is computed as:
\begin{equation}
    PER = \frac{1}{N} \Sigma_{i=1}^{N} PER^i, \qquad\text{ where  } PER^i = \frac{|E^{i} \cap \hat{E}^{i} |}{|E^{i}|}.
\end{equation}




\subsection{Trajectory quality metrics}\label{app:explore-confirm}
In this section, we introduce the \textit{confirmation score} and the \textit{exploration score} used to assess the quality of a trajectory. The exploration score captures how distant two consecutive agent predictions are, while the confirmation score measures how close the agent prediction is to the ground truth differential. Thus, given two consecutive agent predictions $bel_{t-1}$ and $bel_{t}$ together with the target differential diagnosis $y$, we have:
\begin{equation}
    \text{confirmation\_score}_t = \exp(-\text{KL}(y || bel_t)),\qquad\text{ and  }
\end{equation}

\begin{equation}
    \text{exploration\_score}_t = 1 - \exp(-\text{KL}(bel_{t-1} || bel_t)),
\end{equation}
where $\text{KL}(. || .)$ is the Kullback–Leibler divergence. The more $bel_t$ deviates from the prediction at the previous time step $bel_{t-1}$, the higher the exploration score. Also, the closer $bel_t$ to $y$, the higher the confirmation score. Finally, both scores are within the range of $0$ to $1$.

\section{Ablation studies}
\label{app:ablations}

In this section, we present additional ablation analyses to further demonstrate the properties of the proposed approach.

\subsection{Exploration and confirmation schedulers}

\begin{figure}[htb!]
\centering
\begin{subfigure}[b]{.5\textwidth}
  \centering
  \includegraphics[width=.98\linewidth]{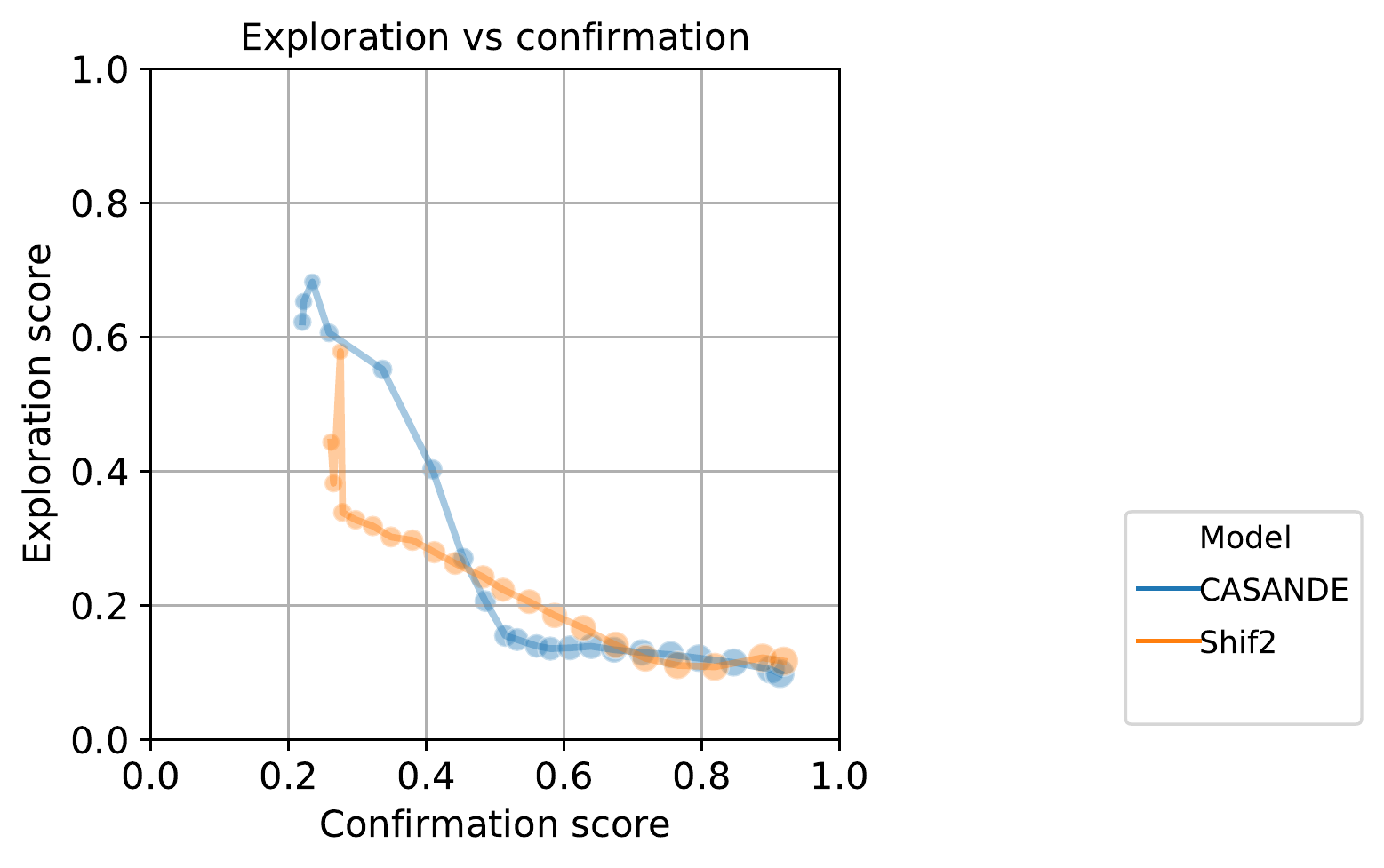}
  \caption{Shif2 vs. CASANDE}
  \label{fig:shif2vscasande}
\end{subfigure}%
\begin{subfigure}[b]{.5\textwidth}
  \centering
  \includegraphics[width=.98\linewidth]{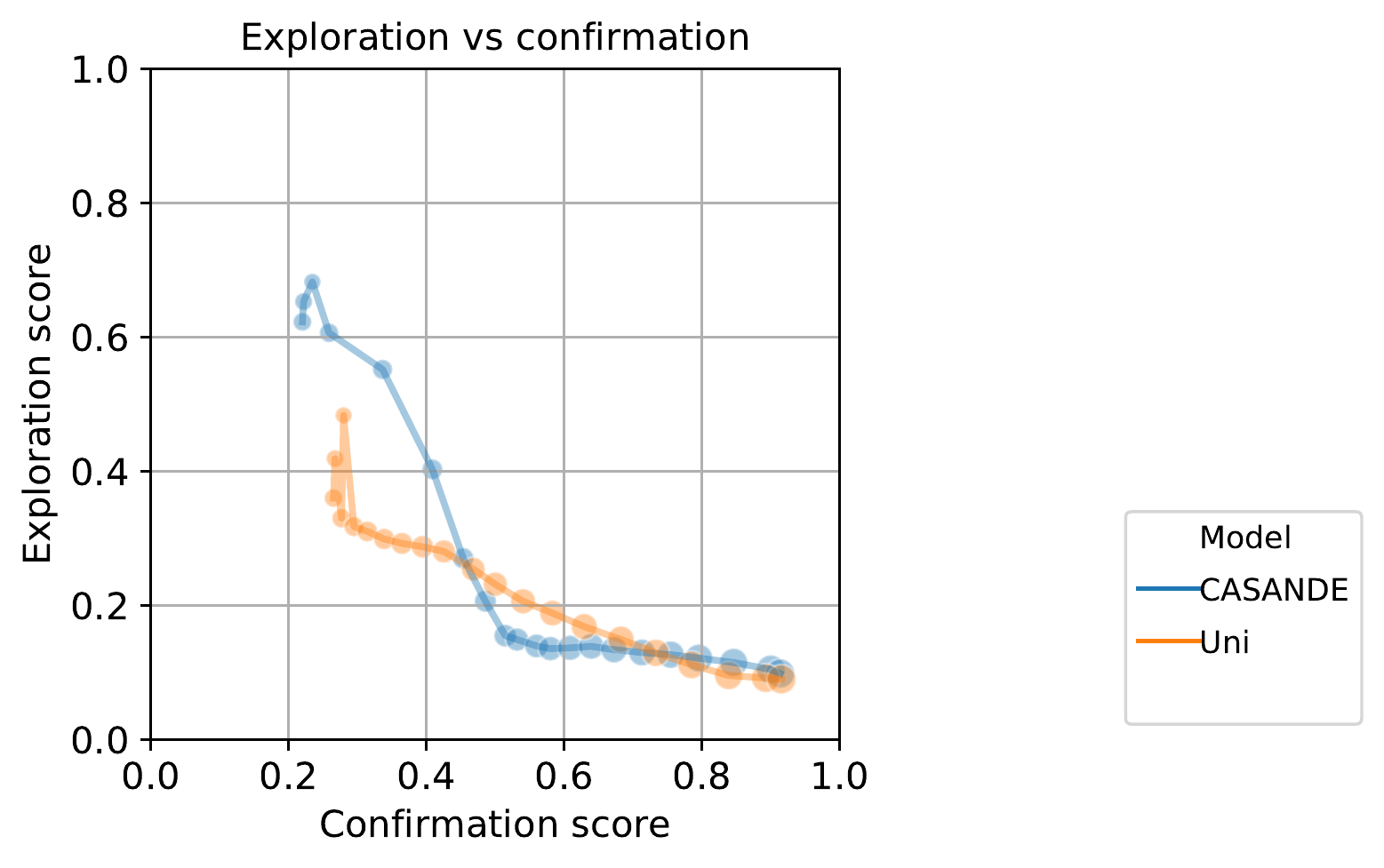}
  \caption{Uni vs. CASANDE}
  \label{fig:univscasande}
\end{subfigure}
\caption{Plots of confirmation score vs. exploration score for different schedulers on the validation set.}
\label{fig:scheduler}
\end{figure}

To analyse the impact of the exploration and confirmation schedulers on the performance of the proposed approach, we consider two additional settings with different scheduling parameters. The first one, referred to as ``Shif2'', uses $\Delta_{\text{Co}}=2$ and $\Delta_{\text{Ex}}=7$. Basically, this shifts the initial schedule forward by two steps. The second setting, referred to as ``Uni'', uses $\Delta_{\text{Co}}=20$ and $\Delta_{\text{Ex}}=-20$, and is designed in such a way that both auxiliary rewards are active during the entire interaction process. As shown in Table~\ref{tbl:scheduler}, the ``Shif2'' setting results in an agent having a slightly better performance in terms of DDF1 when compared to our original setting.  However, this improvement comes at the expense of an increase of the interaction length. Also, the exploration score decreases slowly for the trajectories followed by the "Shif2" agent when compared to the original setting (see Figure~\ref{fig:shif2vscasande}). In the ``Uni'' setting, the agent is tasked with simultaneously optimizing contradictory rewards, one encouraging it to explore different differentials, and one encouraging it to confirm a differential. This leads to higher interaction length, worse evidence recall, and a smaller exploration score at the beginning of the interaction (see Figure~\ref{fig:univscasande}).

\begin{table}[tbh!]
\centering
\caption{Impact of the exploration and confirmation schedulers on interaction length, differential prediction,  severe pathology handling, and evidence collection, as measured on the validation set. Values are in \% except for IL. Values in \textbf{bold} indicate the best performance for a column. Only one experiment was performed for each setting.}
\vspace{1mm}
\small
\begin{tabular}{c|ccccc}
\textbf{} & \textbf{IL}    & \textbf{DDF1} & \textbf{DSHM}  & \textbf{PER}   & \textbf{GTPA}   \\ \hline
Uni       & 23.02                            & 94.17 & \textbf{74.14}                             & 97.72 & 99.76         \\ \hline
Shif2       & 22.82                            & \textbf{94.53}     & 74.04                & 98.35   & 99.80       \\ \hline
Casande   & \textbf{19.92}                   & 94.12 & 74.10                             & \textbf{98.80} & \textbf{99.81} \\ \hline
\end{tabular}
\label{tbl:scheduler}
\end{table}

\subsection{Impact of the severity reward weight}

\begin{figure}
    \centering
    \includegraphics[width=0.55\textwidth]{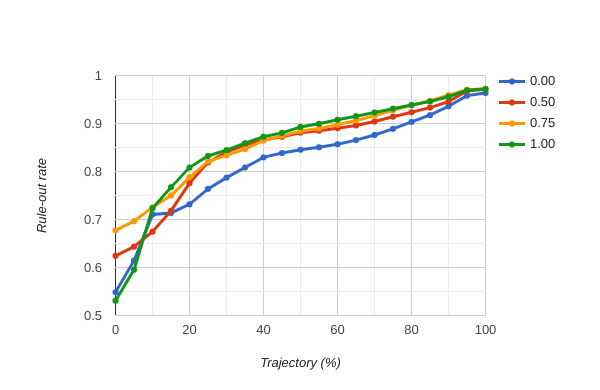}
    \caption{Evolution of the severe pathologies' rule-out rate for different values of  $\alpha_{\text{Sev}}$ on the validation set.}
    \label{fig:weightSevOut}
\end{figure}

Table~\ref{tbl:sevOutWeight} shows the performance of the agent when considering different values of $\alpha_{\text{Sev}}$ and Figure~\ref{fig:weightSevOut} depicts how the rule-out rate of severe pathologies evolves over time throughout the interactions. As $\alpha_{\text{Sev}}$ increases, DDF1 and DSHM improve up at to the point where $\alpha_{\text{Sev}}$ is equal to 0.75, after which both scores go down. This is likely due to the fact that the severity reward focuses on ruling out severe pathologies the patient is not experiencing but doesn't focus on ruling in the relevant severe pathologies (which are handled by the classification reward). $\alpha_{\text{Sev}}$ also affects the pace at which severe pathologies are ruled out from the differential predicted by the agent throughout the interaction process. Indeed, the higher $\alpha_{\text{Sev}}$, the quicker the severe pathologies are ruled out.

\begin{table}[tbh!]
\centering
\caption{Impact of the severity reward weight $\alpha_{\text{Sev}}$ on the differential prediction and the handling of severe pathologies, as measured on the validation set. Values are in \%.  Values in \textbf{bold} indicate the best performance for a row. Only one experiment was performed for each setting.}
\vspace{1mm}
\small
\begin{tabular}{c|cccc}
     & $\alpha_{\text{Sev}}=0.0$ & $\alpha_{\text{Sev}}=0.5$ & $\alpha_{\text{Sev}}=0.75$ & $\alpha_{\text{Sev}}=1.0$ \\ \hline
\textbf{DDF1} & 93.24                     & 93.76                     & \textbf{94.12}                      & 93.66                     \\ \hline
\textbf{DSHM} & 73.52                     & 73.78                     & \textbf{74.10}                      & 73.91                     \\ \hline
\end{tabular}
\label{tbl:sevOutWeight}
\end{table}

\subsection{Impact of the confirmation reward weight}
\begin{figure}
    \centering
    \includegraphics[width=0.55\textwidth]{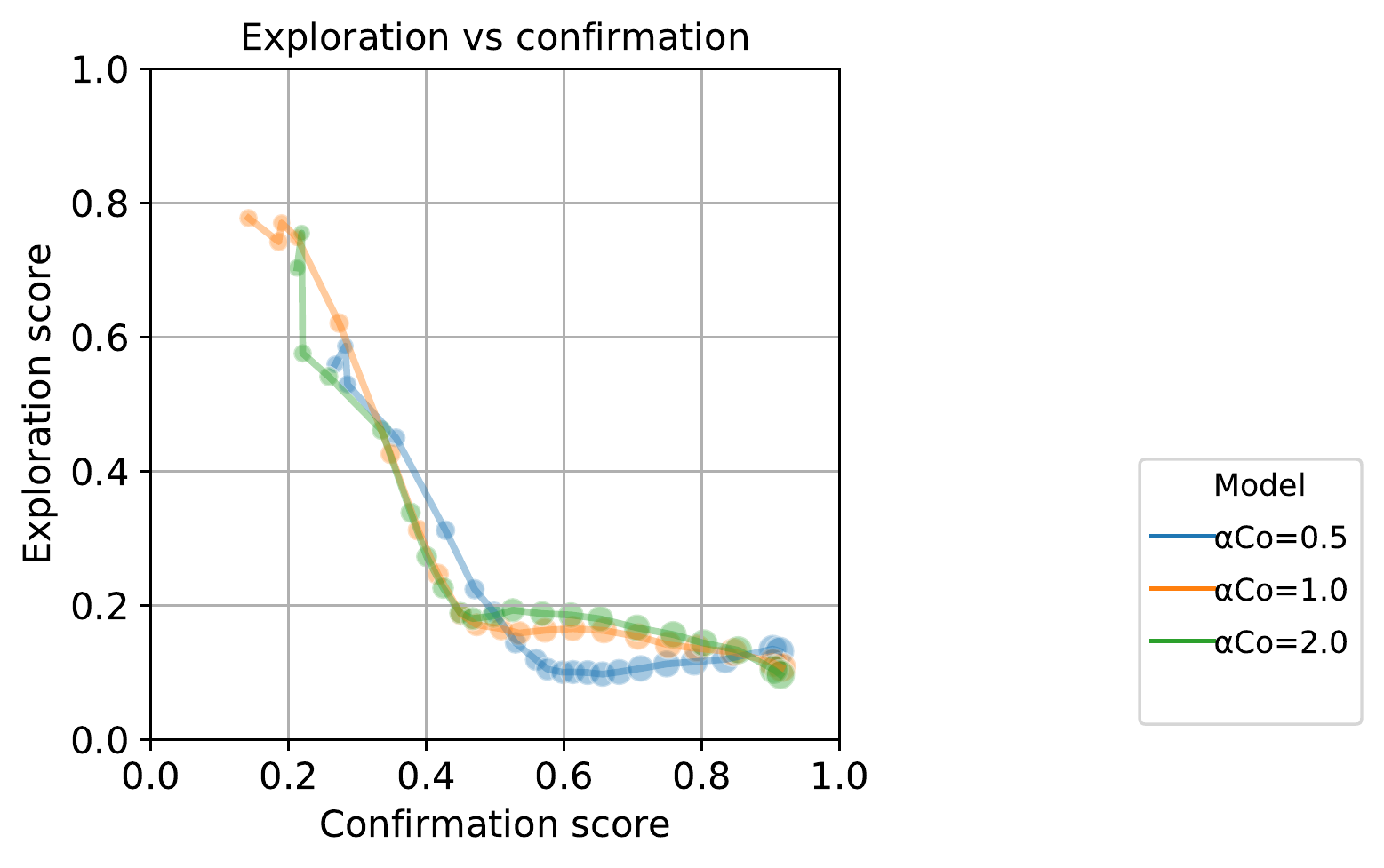}
    \caption{Plots of the confirmation score vs. exploration score for different values of $\alpha_{\text{Co}}$ on the validation set.}
    \label{fig:weightCE}
\end{figure}

Table~\ref{tbl:weightCE} and Figure~\ref{fig:weightCE} show the results obtained when varying the values of $\alpha_{\text{Co}}$. It is noticeable that a low value of $\alpha_{\text{Co}}$ tends to shorten the interaction length. Conversely, a high value of $\alpha_{\text{Co}}$ tends to increase the PER metric. 
Focusing on Figure~\ref{fig:weightCE}, we observe that, for  $\alpha_{\text{Co}} = 0.5$, the exploration score tends to move upward towards the end of the interaction, a behavior that is not desirable. On the other hand, for  $\alpha_{\text{Co}} = 1.0$ and $\alpha_{\text{Co}} = 2.0$, the resulting trajectories are similar and exhibit the desired properties as their exploration scores follow a downward trend at the end of the interaction.

\begin{table}[tbh!]
\centering
\caption{Impact of the confirmation reward weight $\alpha_{\text{Co}}$ on the performance metrics, as measured on the validation set. Values are in \% except for IL. Values in \textbf{bold} indicate the best performance for a column. Only one experiment was performed for each setting.}
\vspace{1mm}
\small
\begin{tabular}{c|cccccl}
\textbf{}                & \textbf{IL}    & \textbf{DDF1} & \textbf{DSHM}  & \textbf{PER} & \textbf{GTPA}   &  \\ \cline{1-6}
$\alpha_{\text{Co}}=0.5$ & \textbf{18.90} & 93.55 & 73.76         & 97.75 & 99.80          &  \\ \cline{1-6}
$\alpha_{\text{Co}}=1.0$ & 19.92          & \textbf{94.12} & \textbf{74.10} & \textbf{98.80} &  \textbf{99.81}        &  \\ \cline{1-6}
$\alpha_{\text{Co}}=2.0$ & 19.58          & 93.86 & 73.81         & {98.65} & \textbf{99.81} &  \\ \cline{1-6}
\end{tabular}
\label{tbl:weightCE}
\end{table}

\subsection{Impact of the exploration reward weight}
\begin{figure}
    \centering
    \includegraphics[width=0.55\textwidth]{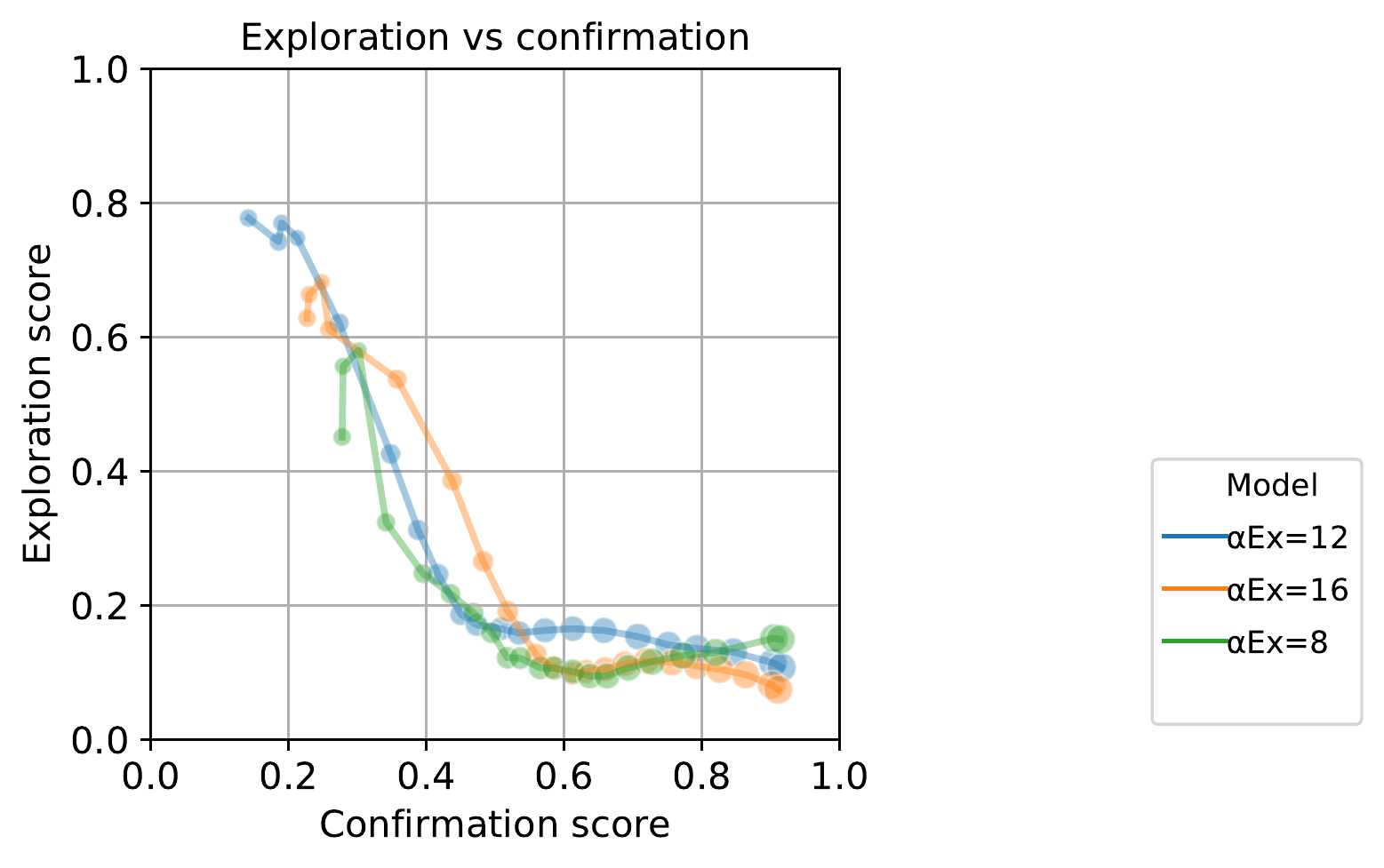}
    \caption{Plots of the confirmation score vs. exploration score for different values of $\alpha_{\text{Ex}}$ on the validation set.}
    \label{fig:weightJS}
\end{figure}

Table~\ref{tbl:weightJS} and Figure~\ref{fig:weightJS} show the results obtained when varying the values of $\alpha_{\text{Ex}}$. As expected, the interaction length tends to increase with the values of $\alpha_{\text{Ex}}$. Also, the higher the value of $\alpha_{\text{Ex}}$, the higher the exploration score is in the initial phase of the interaction with the patient (see Figure~\ref{fig:weightJS}).

\begin{table}[tbh!]
\centering
\caption{Impact of the exploration reward weight $\alpha_{\text{Ex}}$ on the performance metrics, as measured on the validation set. Values are in \% except for IL.  Values in \textbf{bold} indicate the best performance for a column. Only one experiment was performed for each setting.}
\vspace{1mm}
\small
\begin{tabular}{c|ccccc}
\textbf{}               & \textbf{IL}    & \textbf{DDF1} & \textbf{DSHM}  & \textbf{PER} & \textbf{GTPA}   \\ \hline
$\alpha_{\text{Ex}}=8$  & \textbf{18.69} & 93.90 & 73.89         & 97.80 & 99.81          \\ \hline
$\alpha_{\text{Ex}}=16$ & 19.67          & 93.60  & 73.91        & 98.33 & \textbf{99.85}          \\ \hline
$\alpha_{\text{Ex}}=12$ & 19.92          & \textbf{94.12} & \textbf{74.10} & \textbf{98.80} & 99.81 \\ \hline
\end{tabular}
\label{tbl:weightJS}
\end{table}

\subsection{Disabling of the reward functions}

\begin{figure}
    \centering
    \includegraphics[width=\textwidth]{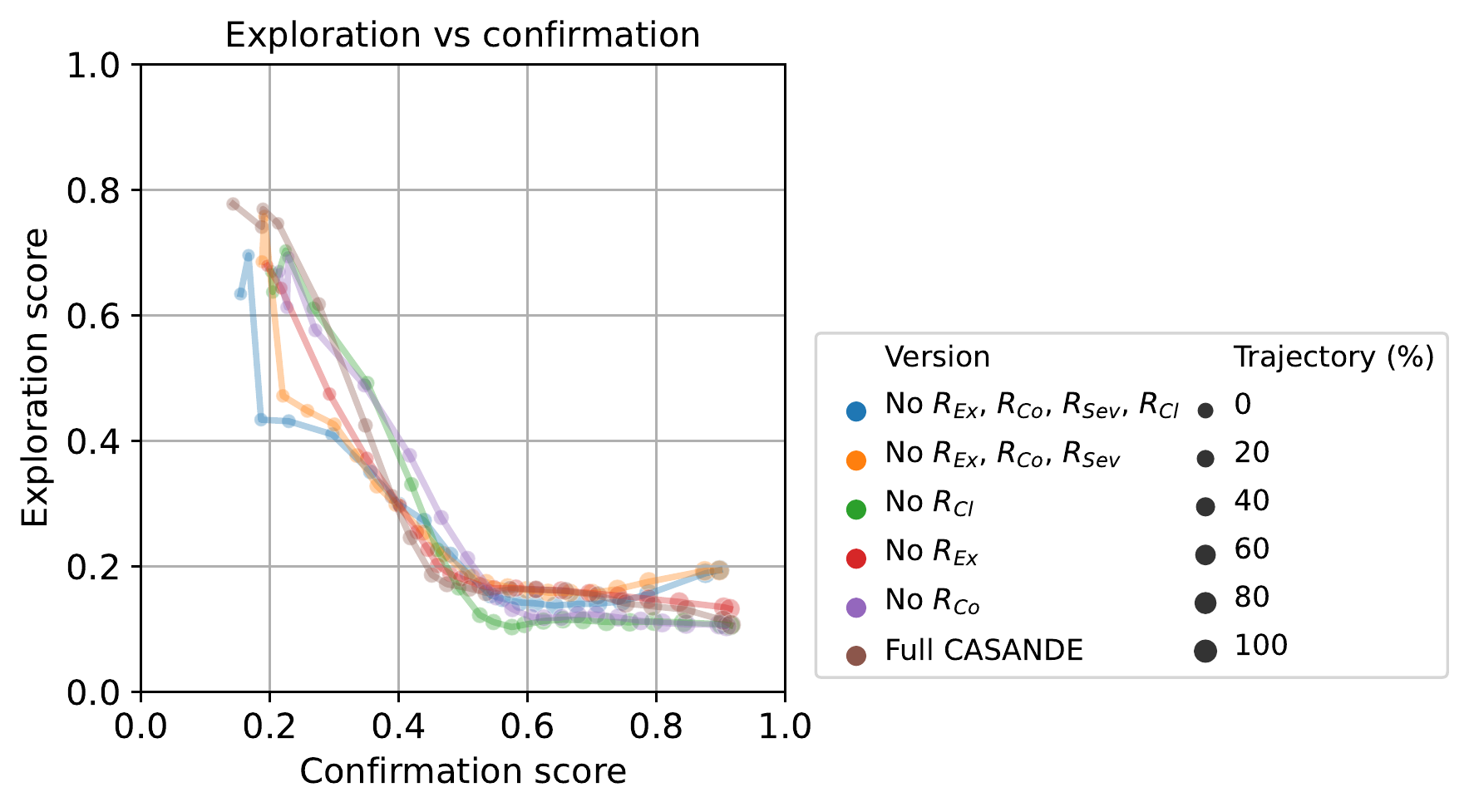}
    \caption{Exploration score vs confirmation score on the validation set for the ablation study presented in Table \ref{tbl:ablations}. The scores correspond to averages over 3 runs using different random seeds.}
    \label{fig:ablation_explore_confirm}
\end{figure}

We presented in Table~\ref{tbl:ablations} the results of an ablation study when different subsets of the exploration, confirmation, severity, and classification reward functions are disabled, and described the impact on DDF1, DSHM and PER. We now analyze the impact of those ablations on the exploration-confirmation score trajectories. Those trajectories are depicted in Figure~\ref{fig:ablation_explore_confirm}. We observe several patterns:
\begin{itemize}
    \item When using all rewards (brown curve), the  trajectory starts with the highest exploration score and a small confirmation score and slowly shifts towards the lower right corner with a small exploration score and a high confirmation score. This trajectory corresponds to the desired behavior.
    \item When the exploration reward is disabled (blue, orange, and red curves), the agent quickly reduces the amount of exploration it does (with the third point on each curve getting close to 40\% of exploration while the other curves are at higher exploration scores at the same stage in the interactions).
    \item When the confirmation reward is disabled (blue, orange, and purples curves), the agent is not as strongly constrained towards the end of the interaction to consolidating its belief and can instead also decide to increase its exploration of differentials. 
    \item When only the classification reward is disabled (green curve), the agent manages to confirm the differential at the end of the interaction thanks to the confirmation reward. 
    \item When all the reward components are disabled (blue curve), we observe good confirmation because the agent is trained to recover the ground truth differential by construction (as part of the classifier training). As for the exploration, the agent starts from some initial distribution and naturally moves toward the ground truth differential. 
\end{itemize}

\section{Qualitative evaluation}
\label{sec:qualEval}

We asked the doctor supporting us to qualitatively evaluate the trajectories generated by CASANDE. The doctor defined the following evaluation criteria, with a score on a 5-point Likert scale:

\begin{itemize}
    \item Q1: The agent asks relevant questions.
    \item Q2: The questions asked allow me to establish a differential diagnosis.
    \item Q3: The agent asks enough questions to make a differential diagnosis.
    \item Q4: The questions asked are similar to what I would have asked.
    \item Q5: The information collected is useful for me to continue assessing the patient.
    \item Q6: The sequence of questions seems logical to me.
\end{itemize}

The Likert scale is defined as follows:
\begin{enumerate}
    \item strongly disagree,
    \item disagree,
    \item neutral,
    \item agree,
    \item strongly agree.
\end{enumerate}

Table~\ref{tab:qualEval} shows the scores on 20 patients that were randomly sampled from the test set. Those scores are next commented by the doctor.

\begin{table}[hbt!]
\centering
\caption{Qualitative evaluation of CASANDE on 20 randomly selected patients from the test set. The evaluation is based on 6 questions whose scores are defined on a 5-point Likert scale. The last column of the table shows the average score with the standard deviation, in brackets, for each question.}
\label{tab:qualEval}
\vspace{1mm}
\resizebox{\textwidth}{!}{%
\begin{tabular}{c|cccccccccccccccccccc|c}
\hline
\multirow{2}{*}{\textbf{Questions}} & \multicolumn{20}{c|}{\textbf{Patient IDs}}                         & \multirow{2}{*}{\textbf{Average}} \\ \cline{2-21}
                                    & \multicolumn{1}{c}{\textbf{1}} & \multicolumn{1}{c}{\textbf{2}} & \multicolumn{1}{c}{\textbf{3}} & \multicolumn{1}{c}{\textbf{4}} & \multicolumn{1}{c}{\textbf{5}} & \multicolumn{1}{c}{\textbf{6}} & \multicolumn{1}{c}{\textbf{7}} & \multicolumn{1}{c}{\textbf{8}} & \multicolumn{1}{c}{\textbf{9}} & \multicolumn{1}{c}{\textbf{10}} & \multicolumn{1}{c}{\textbf{11}} & \multicolumn{1}{c}{\textbf{12}} & \multicolumn{1}{c}{\textbf{13}} & \multicolumn{1}{c}{\textbf{14}} & \multicolumn{1}{c}{\textbf{15}} & \multicolumn{1}{c}{\textbf{16}} & \multicolumn{1}{c}{\textbf{17}} & \multicolumn{1}{c}{\textbf{18}} & \multicolumn{1}{c}{\textbf{19}} & \textbf{20} &                                   \\ \hline \hline
\textbf{Q1}                         & \multicolumn{1}{c}{4}          & \multicolumn{1}{c}{4}          & \multicolumn{1}{c}{4}          & \multicolumn{1}{c}{5}          & \multicolumn{1}{c}{2}          & \multicolumn{1}{c}{5}          & \multicolumn{1}{c}{4}          & \multicolumn{1}{c}{4}          & \multicolumn{1}{c}{4}          & \multicolumn{1}{c}{4}           & \multicolumn{1}{c}{3}           & \multicolumn{1}{c}{4}           & \multicolumn{1}{c}{4}           & \multicolumn{1}{c}{4}           & \multicolumn{1}{c}{3}           & \multicolumn{1}{c}{4}           & \multicolumn{1}{c}{4}           & \multicolumn{1}{c}{4}           & \multicolumn{1}{c}{4}           & 4           & \textbf{3.90 ($\pm$ 0.64)}        \\ \hline
\textbf{Q2}                         & \multicolumn{1}{c}{4}          & \multicolumn{1}{c}{5}          & \multicolumn{1}{c}{4}          & \multicolumn{1}{c}{5}          & \multicolumn{1}{c}{3}          & \multicolumn{1}{c}{5}          & \multicolumn{1}{c}{4}          & \multicolumn{1}{c}{5}          & \multicolumn{1}{c}{4}          & \multicolumn{1}{c}{4}           & \multicolumn{1}{c}{3}           & \multicolumn{1}{c}{4}           & \multicolumn{1}{c}{5}           & \multicolumn{1}{c}{4}           & \multicolumn{1}{c}{4}           & \multicolumn{1}{c}{4}           & \multicolumn{1}{c}{5}           & \multicolumn{1}{c}{3}           & \multicolumn{1}{c}{4}           & 5           & \textbf{4.20 ($\pm$ 0.70)}        \\ \hline
\textbf{Q3}                         & \multicolumn{1}{c}{4}          & \multicolumn{1}{c}{5}          & \multicolumn{1}{c}{5}          & \multicolumn{1}{c}{5}          & \multicolumn{1}{c}{3}          & \multicolumn{1}{c}{5}          & \multicolumn{1}{c}{4}          & \multicolumn{1}{c}{5}          & \multicolumn{1}{c}{5}          & \multicolumn{1}{c}{5}           & \multicolumn{1}{c}{3}           & \multicolumn{1}{c}{4}           & \multicolumn{1}{c}{4}           & \multicolumn{1}{c}{4}           & \multicolumn{1}{c}{4}           & \multicolumn{1}{c}{5}           & \multicolumn{1}{c}{5}           & \multicolumn{1}{c}{3}           & \multicolumn{1}{c}{4}           & 5           & \textbf{4.35 ($\pm$ 0.75)}        \\ \hline
\textbf{Q4}                         & \multicolumn{1}{c}{3}          & \multicolumn{1}{c}{4}          & \multicolumn{1}{c}{4}          & \multicolumn{1}{c}{4}          & \multicolumn{1}{c}{2}          & \multicolumn{1}{c}{4}          & \multicolumn{1}{c}{3}          & \multicolumn{1}{c}{4}          & \multicolumn{1}{c}{4}          & \multicolumn{1}{c}{4}           & \multicolumn{1}{c}{2}           & \multicolumn{1}{c}{3}           & \multicolumn{1}{c}{4}           & \multicolumn{1}{c}{3}           & \multicolumn{1}{c}{3}           & \multicolumn{1}{c}{4}           & \multicolumn{1}{c}{4}           & \multicolumn{1}{c}{3}           & \multicolumn{1}{c}{4}           & 4           & \textbf{3.50 ($\pm$ 0.69)}        \\ \hline
\textbf{Q5}                         & \multicolumn{1}{c}{5}          & \multicolumn{1}{c}{5}          & \multicolumn{1}{c}{5}          & \multicolumn{1}{c}{5}          & \multicolumn{1}{c}{3}          & \multicolumn{1}{c}{5}          & \multicolumn{1}{c}{4}          & \multicolumn{1}{c}{5}          & \multicolumn{1}{c}{5}          & \multicolumn{1}{c}{4}           & \multicolumn{1}{c}{3}           & \multicolumn{1}{c}{4}           & \multicolumn{1}{c}{5}           & \multicolumn{1}{c}{5}           & \multicolumn{1}{c}{4}           & \multicolumn{1}{c}{5}           & \multicolumn{1}{c}{4}           & \multicolumn{1}{c}{4}           & \multicolumn{1}{c}{5}           & 5           & \textbf{4.50 ($\pm$ 0.69)}        \\ \hline
\textbf{Q6}                         & \multicolumn{1}{c}{4}          & \multicolumn{1}{c}{3}          & \multicolumn{1}{c}{3}          & \multicolumn{1}{c}{4}          & \multicolumn{1}{c}{1}          & \multicolumn{1}{c}{4}          & \multicolumn{1}{c}{4}          & \multicolumn{1}{c}{3}          & \multicolumn{1}{c}{3}          & \multicolumn{1}{c}{3}           & \multicolumn{1}{c}{2}           & \multicolumn{1}{c}{3}           & \multicolumn{1}{c}{3}           & \multicolumn{1}{c}{3}           & \multicolumn{1}{c}{3}           & \multicolumn{1}{c}{4}           & \multicolumn{1}{c}{4}           & \multicolumn{1}{c}{4}           & \multicolumn{1}{c}{4}           & 3           & \textbf{3.25 ($\pm$ 0.79)}        \\ \hline
\end{tabular}%
}
\end{table}

\textbf{Evaluating physician's comments}

The evaluation of the interactions was done as if the agent was trained on an extensive set that would include most of the pathologies. As a doctor, it would be very difficult for me to evaluate it otherwise: my training and clinical experience in acute care setting shaped how I optimize patient evaluation. It would be impractical to deconstruct and optimize my framework on a smaller set of diseases, as I would require to train this skill. It mostly explains the lower general scores attributed to Q1 and Q4.

That being formulated, we also have to keep in mind that a tool such as CASANDE has to be considered as a mean to improve the patient experience and improve clinical outcome, by providing the care team relevant medical information from which they can build on. 

The differential diagnosis approach is the one we use in practice when interacting with patients. The differential helps us explore potential pathologies and converge toward a most likely differential towards the end of the interaction. That differential will guide the next steps in the evaluation and help choose the best investigations and treatments. The clinical context influences the differential. In an acute care setting, more emphasis will be put on diseases carrying higher short term mortality and/or morbidity.

Regarding Q2, the questions asked by the agent generally allowed me to build a good differential that I would find useful in clinical practice. 

For Q3, the amount of questions seemed reasonable. Of course, since the agent is not trained on the same amount of pathologies than I am, I would naturally seek out a wider array of diseases and negative evidences in the initial exploration.

Q5 is somewhat correlated to Q2. I found the information relevant and inclusive, even seeking past medical history elements and risk factors, all of that useful in a clinical context. It would be very easy to go to that patient and complete the questionnaire.

Finally, for Q6, as mentioned earlier, the optimal strategy used by CASANDE is derived from the breadth of the dataset trained on. I did not expect the agent to use a similar sequence. The order is less relevant \textit{during} the patient evaluation in a clinical context, as the collected evidences are usually reordered by systems and sections in the patient's chart \textit{a posteriori} and then presented to the attending physician. Still with the questions asked, I could understand what differential was being evaluated at each step, which resembles what I would normally do.

The trajectories corresponding to those 20 patients are shown in what follows.

\subsection*{Patient 1}

\begin{lstlisting}
Sex: F, Age: 45
Initial evidence: 
	 I have recently had stools that were black (like coal)
Agent inquiries:
---------------
*Turn 1:
  Q: Do you have pain somewhere, related to your reason for consulting?
  A: Y
*Turn 2:
  Q: Does the pain radiate to another location?
  A:
	 * nowhere
*Turn 3:
  Q: Do you have swelling in one or more areas of your body?
  A: N
*Turn 4:
  Q: Characterize your pain:
  A:
	 * sensitive
	 * tugging
	 * a cramp
	 * exhausting
*Turn 5:
  Q: Do you feel pain somewhere?
  A:
	 * back of head
	 * forehead
*Turn 6:
  Q: How precisely is the pain located?
  A:
	 * 4
*Turn 7:
  Q: How intense is the pain?
  A:
	 * 1
*Turn 8:
  Q: How fast did the pain appear?
  A:
	 * 4
*Turn 9:
  Q: Are you taking any new oral anticoagulants ((NOACs)?
  A: Y
*Turn 10:
  Q: Have you ever had a diagnosis of anemia?
  A: Y
*Turn 11:
  Q: Have you noticed light red blood or blood clots in your stool?
  A: Y
*Turn 12:
  Q: Do you have chronic kidney failure?
  A: Y
*Turn 13:
  Q: Do you have a poor diet?
  A: N
*Turn 14:
  Q: Do you have any family members who have been diagnosed with anemia?
  A: Y
*Turn 15:
  Q: Is your BMI less than 18.5, or are you underweight?
  A: Y
*Turn 16:
  Q: Have you traveled out of the country in the last 4 weeks?
  A: N
*Turn 17:
  Q: Do you feel so tired that you are unable to do your usual activities or are you stuck in your bed all day long?
  A: Y
*Turn 18:
  Q: Do you think you are pregnant or are you currently pregnant?
  A: Y
*Turn 19:
  Q: Do you feel slightly dizzy or lightheaded?
  A: Y
*Turn 20:
  Q: Do you feel lightheaded and dizzy or do you feel like you are about to faint?
  A: Y
*Turn 21:
  Q: Are you experiencing shortness of breath or difficulty breathing in a significant way?
  A: N
*Turn 22:
  Q: Is your skin much paler than usual?
  A: Y
Predicted differential:
----------------------
Anemia, Atrial fibrillation, Cluster headache, HIV (initial infection), PSVT
Ground truth differential:
-------------------------
Anemia: 0.349, Atrial fibrillation: 0.243, Cluster headache: 0.153, HIV (initial infection): 0.152, PSVT: 0.103
\end{lstlisting}

\subsection*{Patient 2}

\begin{lstlisting}
Sex: F, Age: 19
Initial evidence: 
	 I have a sore throat
Agent inquiries:
---------------
*Turn 1:
  Q: Do you have swelling in one or more areas of your body?
  A: N
*Turn 2:
  Q: Do you have any lesions, redness or problems on your skin that you believe are related to the condition you are consulting for?
  A: N
*Turn 3:
  Q: Do you have pain somewhere, related to your reason for consulting?
  A: Y
*Turn 4:
  Q: Do you feel pain somewhere?
  A:
	 * top of the head
	 * forehead
	 * cheek(R)
	 * occiput
	 * temple(R)
*Turn 5:
  Q: Characterize your pain:
  A:
	 * sensitive
*Turn 6:
  Q: How precisely is the pain located?
  A:
	 * 1
*Turn 7:
  Q: How intense is the pain?
  A:
	 * 6
*Turn 8:
  Q: How fast did the pain appear?
  A:
	 * 3
*Turn 9:
  Q: Do you live with 4 or more people?
  A: Y
*Turn 10:
  Q: Do you have a cough that produces colored or more abundant sputum than usual?
  A: N
*Turn 11:
  Q: Do you have a cough?
  A: Y
*Turn 12:
  Q: Do you have nasal congestion or a clear runny nose?
  A: Y
*Turn 13:
  Q: Are you exposed to secondhand cigarette smoke on a daily basis?
  A: Y
*Turn 14:
  Q: Do you have a fever (either felt or measured with a thermometer)?
  A: N
*Turn 15:
  Q: Have you been in contact with a person with similar symptoms in the past 2 weeks?
  A: Y
*Turn 16:
  Q: Do you smoke cigarettes?
  A: Y
*Turn 17:
  Q: Do you attend or work in a daycare?
  A: Y
*Turn 18:
  Q: Have you had significantly increased sweating?
  A: N
*Turn 19:
  Q: Do you have diffuse (widespread) muscle pain?
  A: Y
*Turn 20:
  Q: Do you have any family members who have asthma?
  A: N
Predicted differential:
----------------------
URTI, Bronchitis, Cluster headache, Chronic rhinosinusitis, Acute rhinosinusitis, Influenza, Anemia, Chagas
Ground truth differential:
-------------------------
URTI: 0.230, Cluster headache: 0.174, Bronchitis: 0.167, Chronic rhinosinusitis: 0.141, Acute rhinosinusitis: 0.101, Chagas: 0.066, Anemia: 0.066, Influenza: 0.054
\end{lstlisting}

\subsection*{Patient 3}

\begin{lstlisting}
Sex: M, Age: 36
Initial evidence: 
	 I feel pain
Agent inquiries:
---------------
*Turn 1:
  Q: Do you have swelling in one or more areas of your body?
  A: N
*Turn 2:
  Q: Do you have any lesions, redness or problems on your skin that you believe are related to the condition you are consulting for?
  A: Y
*Turn 3:
  Q: What color is the rash?
  A:
	 * yellow
*Turn 4:
  Q: Characterize your pain:
  A:
	 * sensitive
	 * a cramp
*Turn 5:
  Q: Do you feel pain somewhere?
  A:
	 * flank(L)
*Turn 6:
  Q: Where is the affected region located?
  A:
	 * epigastric
*Turn 7:
  Q: How intense is the pain?
  A:
	 * 6
*Turn 8:
  Q: How precisely is the pain located?
  A:
	 * 5
*Turn 9:
  Q: How severe is the itching?
  A:
	 * 3
*Turn 10:
  Q: How intense is the pain caused by the rash?
  A:
	 * 1
*Turn 11:
  Q: Is the lesion (or are the lesions) larger than 1cm?
  A:
	 * Y
*Turn 12:
  Q: Is the rash swollen?
  A:
	 * 1
*Turn 13:
  Q: How fast did the pain appear?
  A:
	 * 2
*Turn 14:
  Q: Have you had an involuntary weight loss over the last 3 months?
  A: Y
*Turn 15:
  Q: Are there members of your family who have been diagnosed with pancreatic cancer?
  A: Y
*Turn 16:
  Q: Have you had diarrhea or an increase in stool frequency?
  A: Y
*Turn 17:
  Q: Do you have diabetes?
  A: N
*Turn 18:
  Q: Do you have chronic pancreatitis?
  A: Y
*Turn 19:
  Q: Are you significantly overweight compared to people of the same height as you?
  A: N
*Turn 20:
  Q: Does the pain radiate to another location?
  A:
	 * thoracic spine
*Turn 21:
  Q: Do you have pale stools and dark urine?
  A: Y
*Turn 22:
  Q: Do you constantly feel fatigued or do you have non-restful sleep?
  A: Y
*Turn 23:
  Q: Do you smoke cigarettes?
  A: Y
*Turn 24:
  Q: Are you feeling nauseous or do you feel like vomiting?
  A: N
*Turn 25:
  Q: Do you have a cough?
  A: Y
Predicted differential:
----------------------
Pancreatic neoplasm, Bronchitis, Anaphylaxis, Scombroid food poisoning, Chagas, Pulmonary neoplasm
Ground truth differential:
-------------------------
Pancreatic neoplasm: 0.265, Anaphylaxis: 0.213, Pulmonary neoplasm: 0.151, Scombroid food poisoning: 0.142, Chagas: 0.118, Bronchitis: 0.111
\end{lstlisting}

\subsection*{Patient 4}

\begin{lstlisting}
Sex: F, Age: 39
Initial evidence: 
	 I am coughing
Agent inquiries:
---------------
*Turn 1:
  Q: Do you have any lesions, redness or problems on your skin that you believe are related to the condition you are consulting for?
  A: N
*Turn 2:
  Q: Do you have pain somewhere, related to your reason for consulting?
  A: Y
*Turn 3:
  Q: Does the pain radiate to another location?
  A:
	 * posterior chest wall(R)
	 * posterior chest wall(L)
*Turn 4:
  Q: Do you feel pain somewhere?
  A:
	 * side of the chest(R)
	 * breast(R)
	 * breast(L)
	 * posterior chest wall(R)
	 * posterior chest wall(L)
*Turn 5:
  Q: Characterize your pain:
  A:
	 * heartbreaking
	 * haunting
	 * violent
	 * sharp
*Turn 6:
  Q: How precisely is the pain located?
  A:
	 * 10
*Turn 7:
  Q: Do you have pain that is increased when you breathe in deeply?
  A: Y
*Turn 8:
  Q: Do you have pain that is increased with movement?
  A: Y
*Turn 9:
  Q: How intense is the pain?
  A:
	 * 10
*Turn 10:
  Q: Are you being treated for osteoporosis?
  A: Y
*Turn 11:
  Q: How fast did the pain appear?
  A:
	 * 9
*Turn 12:
  Q: Do you drink alcohol excessively or do you have an addiction to alcohol?
  A: N
*Turn 13:
  Q: Do you have intense coughing fits?
  A: Y
*Turn 14:
  Q: Are you experiencing shortness of breath or difficulty breathing in a significant way?
  A: N
Predicted differential:
----------------------
Spontaneous rib fracture, Bronchitis, Possible NSTEMI / STEMI, Unstable angina, Pulmonary embolism, Whooping cough, Stable angina, Spontaneous pneumothorax
Ground truth differential:
-------------------------
Spontaneous rib fracture: 0.205, Bronchitis: 0.187, Whooping cough: 0.133, Pulmonary embolism: 0.128, Possible NSTEMI / STEMI: 0.126, Unstable angina: 0.119, Stable angina: 0.061, Spontaneous pneumothorax: 0.040
\end{lstlisting}

\subsection*{Patient 5}

\begin{lstlisting}
Sex: M, Age: 25
Initial evidence: 
	 I had an involuntary weight loss over the last 3 months
Agent inquiries:
---------------
*Turn 1:
  Q: Do you have pain somewhere, related to your reason for consulting?
  A: Y
*Turn 2:
  Q: Do you have any lesions, redness or problems on your skin that you believe are related to the condition you are consulting for?
  A: Y
*Turn 3:
  Q: What color is the rash?
  A:
	 * pale
*Turn 4:
  Q: Characterize your pain:
  A:
	 * a pulse
	 * exhausting
*Turn 5:
  Q: Do you feel pain somewhere?
  A:
	 * cervical spine
	 * forehead
	 * temple(R)
	 * temple(L)
*Turn 6:
  Q: Where is the affected region located?
  A:
	 * lower gum
	 * upper gum
	 * labia majora(R)
	 * labia majora(L)
	 * internal cheek(R)
*Turn 7:
  Q: How precisely is the pain located?
  A:
	 * 2
*Turn 8:
  Q: How intense is the pain caused by the rash?
  A:
	 * 7
*Turn 9:
  Q: How intense is the pain?
  A:
	 * 4
*Turn 10:
  Q: Is the rash swollen?
  A:
	 * 0
*Turn 11:
  Q: Is the lesion (or are the lesions) larger than 1cm?
  A:
	 * Y
*Turn 12:
  Q: How fast did the pain appear?
  A:
	 * 4
*Turn 13:
  Q: Do you have a fever (either felt or measured with a thermometer)?
  A: Y
*Turn 14:
  Q: Have you had sexual intercourse with an HIV-positive partner in the past 12 months?
  A: Y
*Turn 15:
  Q: Have you had unprotected sex with more than one partner in the last 6 months?
  A: Y
*Turn 16:
  Q: Do you feel so tired that you are unable to do your usual activities or are you stuck in your bed all day long?
  A: N
*Turn 17:
  Q: Have you ever had a sexually transmitted infection?
  A: Y
*Turn 18:
  Q: Are you currently using intravenous drugs?
  A: Y
*Turn 19:
  Q: Do you have swollen or painful lymph nodes?
  A: N
*Turn 20:
  Q: Have you had diarrhea or an increase in stool frequency?
  A: Y
*Turn 21:
  Q: Have you had significantly increased sweating?
  A: Y
*Turn 22:
  Q: Do you have diffuse (widespread) muscle pain?
  A: N
*Turn 23:
  Q: Do you have a sore throat?
  A: N
*Turn 24:
  Q: Are you feeling nauseous or do you feel like vomiting?
  A: N
*Turn 25:
  Q: Do you have painful mouth ulcers or sores?
  A: N
Predicted differential:
----------------------
HIV (initial infection), Chagas, Scombroid food poisoning
Ground truth differential:
-------------------------
HIV (initial infection): 0.544, Chagas: 0.353, Scombroid food poisoning: 0.103
\end{lstlisting}

\subsection*{Patient 6}

\begin{lstlisting}
Sex: F, Age: 56
Initial evidence: 
	 I am experiencing shortness of breath or difficulty breathing in a significant way
Agent inquiries:
---------------
*Turn 1:
  Q: Do you have any lesions, redness or problems on your skin that you believe are related to the condition you are consulting for?
  A: N
*Turn 2:
  Q: Do you have pain somewhere, related to your reason for consulting?
  A: Y
*Turn 3:
  Q: Does the pain radiate to another location?
  A:
	 * side of the chest(R)
	 * side of the chest(L)
	 * scapula(R)
	 * breast(R)
	 * posterior chest wall(R)
*Turn 4:
  Q: Do you feel pain somewhere?
  A:
	 * side of the chest(R)
	 * side of the chest(L)
	 * scapula(R)
	 * scapula(L)
	 * posterior chest wall(R)
*Turn 5:
  Q: Characterize your pain:
  A:
	 * sensitive
	 * a knife stroke
*Turn 6:
  Q: How precisely is the pain located?
  A:
	 * 8
*Turn 7:
  Q: How intense is the pain?
  A:
	 * 6
*Turn 8:
  Q: How fast did the pain appear?
  A:
	 * 4
*Turn 9:
  Q: Do you have swelling in one or more areas of your body?
  A: Y
*Turn 10:
  Q: Have you had surgery within the last month?
  A: Y
*Turn 11:
  Q: Do you have pain that is increased when you breathe in deeply?
  A: N
*Turn 12:
  Q: Have you been coughing up blood?
  A: Y
*Turn 13:
  Q: Did you lose consciousness?
  A: Y
*Turn 14:
  Q: Where is the swelling located?
  A:
	 * posterior aspect of the ankle(R)
	 * posterior aspect of the ankle(L)
	 * calf(L)
*Turn 15:
  Q: Have you ever had deep vein thrombosis (DVT)?
  A: N
*Turn 16:
  Q: Have you been unable to move or get up for more than 3 consecutive days within the last 4 weeks?
  A: Y
*Turn 17:
  Q: Have you traveled out of the country in the last 4 weeks?
  A: N
*Turn 18:
  Q: Do you currently take hormones?
  A: Y
*Turn 19:
  Q: Do you have an active cancer?
  A: N
Predicted differential:
----------------------
Pulmonary embolism, Acute pulmonary edema, Panic attack, Possible NSTEMI / STEMI, Unstable angina, Myocarditis, Stable angina, Pulmonary neoplasm, Guillain-Barre syndrome, Atrial fibrillation, Acute dystonic reactions, Myasthenia gravis, Anemia, Sarcoidosis, PSVT, Anaphylaxis
Ground truth differential:
-------------------------
Pulmonary embolism: 0.114, Acute pulmonary edema: 0.094, Panic attack: 0.090, Possible NSTEMI / STEMI: 0.084, Unstable angina: 0.080, Myocarditis: 0.070, Pulmonary neoplasm: 0.070, Stable angina: 0.065, Guillain-Barre syndrome: 0.063, Atrial fibrillation: 0.060, Acute dystonic reactions: 0.044, Myasthenia gravis: 0.044, Anemia: 0.044, Sarcoidosis: 0.036, PSVT: 0.024, Anaphylaxis: 0.016
\end{lstlisting}

\subsection*{Patient 7}

\begin{lstlisting}
Sex: F, Age: 45
Initial evidence: 
	 I feel pain
Agent inquiries:
---------------
*Turn 1:
  Q: Do you have swelling in one or more areas of your body?
  A: N
*Turn 2:
  Q: Do you have any lesions, redness or problems on your skin that you believe are related to the condition you are consulting for?
  A: N
*Turn 3:
  Q: Does the pain radiate to another location?
  A:
	 * nowhere
*Turn 4:
  Q: Do you feel pain somewhere?
  A:
	 * side of the chest(L)
	 * flank(R)
	 * hypochondrium(R)
	 * hypochondrium(L)
	 * belly
*Turn 5:
  Q: Characterize your pain:
  A:
	 * a cramp
	 * sharp
*Turn 6:
  Q: How precisely is the pain located?
  A:
	 * 6
*Turn 7:
  Q: How intense is the pain?
  A:
	 * 7
*Turn 8:
  Q: How fast did the pain appear?
  A:
	 * 7
*Turn 9:
  Q: Have you ever been diagnosed with gastroesophageal reflux?
  A: N
*Turn 10:
  Q: Have you ever felt like you were suffocating for a very short time associated with inability to breathe or speak?
  A: N
*Turn 11:
  Q: Are you infected with the human immunodeficiency virus (HIV)?
  A: N
*Turn 12:
  Q: Do you have a known severe food allergy?
  A: N
*Turn 13:
  Q: Have you ever had a spontaneous pneumothorax?
  A: N
*Turn 14:
  Q: Do you have a deviated nasal septum?
  A: N
*Turn 15:
  Q: Do you suffer from chronic anxiety?
  A: Y
*Turn 16:
  Q: Do you have diabetes?
  A: N
*Turn 17:
  Q: Have you had diarrhea or an increase in stool frequency?
  A: N
*Turn 18:
  Q: Have you ever had a head trauma?
  A: Y
*Turn 19:
  Q: Have you ever been diagnosed with depression?
  A: Y
*Turn 20:
  Q: Are you experiencing shortness of breath or difficulty breathing in a significant way?
  A: Y
*Turn 21:
  Q: Do you feel your heart is beating fast (racing), irregularly (missing a beat) or do you feel palpitations?
  A: Y
*Turn 22:
  Q: Have you recently had numbness, loss of sensation or tingling, in both arms and legs and around your mouth?
  A: Y
*Turn 23:
  Q: Do you suffer from fibromyalgia?
  A: N
*Turn 24:
  Q: Do you feel like you are dying or were you afraid that you were about do die?
  A: Y
*Turn 25:
  Q: Do you have asthma or have you ever had to use a bronchodilator in the past?
  A: N
*Turn 26:
  Q: Do you feel like you are detached from your own body or your surroundings?
  A: Y
*Turn 27:
  Q: Are you feeling nauseous or do you feel like vomiting?
  A: N
*Turn 28:
  Q: Do you feel anxious?
  A: Y
*Turn 29:
  Q: Have you ever had a migraine or is a member of your family known to have migraines?
  A: Y
*Turn 30:
  Q: Do you feel like you are (or were) choking or suffocating?
  A: N
Predicted differential:
----------------------
Panic attack, PSVT, Guillain-Barre syndrome, Spontaneous pneumothorax, Possible NSTEMI / STEMI, Atrial fibrillation, Unstable angina, Acute pulmonary edema, Anaphylaxis, Pulmonary embolism, Anemia, GERD, Stable angina, Scombroid food poisoning, Acute dystonic reactions, Myasthenia gravis, Sarcoidosis, Pancreatic neoplasm, SLE
Ground truth differential:
-------------------------
Panic attack: 0.101, PSVT: 0.085, Guillain-Barre syndrome: 0.077, Spontaneous pneumothorax: 0.067, Possible NSTEMI / STEMI: 0.063, Atrial fibrillation: 0.063, Anemia: 0.061, Anaphylaxis: 0.060, Pulmonary embolism: 0.055, Acute pulmonary edema: 0.054, Unstable angina: 0.051, Stable angina: 0.049, GERD: 0.042, Scombroid food poisoning: 0.039, Acute dystonic reactions: 0.033, Myasthenia gravis: 0.033, Sarcoidosis: 0.027, Pancreatic neoplasm: 0.023, SLE: 0.015
\end{lstlisting}

\subsection*{Patient 8}

\begin{lstlisting}
Sex: M, Age: 16
Initial evidence: 
	 I have nasal congestion
Agent inquiries:
---------------
*Turn 1:
  Q: Do you have any lesions, redness or problems on your skin that you believe are related to the condition you are consulting for?
  A: N
*Turn 2:
  Q: Do you have pain somewhere, related to your reason for consulting?
  A: Y
*Turn 3:
  Q: Characterize your pain:
  A:
	 * tedious
	 * sensitive
*Turn 4:
  Q: Do you feel pain somewhere?
  A:
	 * forehead
	 * cheek(R)
	 * temple(L)
*Turn 5:
  Q: How precisely is the pain located?
  A:
	 * 1
*Turn 6:
  Q: How intense is the pain?
  A:
	 * 4
*Turn 7:
  Q: How fast did the pain appear?
  A:
	 * 2
*Turn 8:
  Q: Do you have a cough?
  A: Y
*Turn 9:
  Q: Are you exposed to secondhand cigarette smoke on a daily basis?
  A: Y
*Turn 10:
  Q: Do you have a cough that produces colored or more abundant sputum than usual?
  A: N
*Turn 11:
  Q: Do you have a sore throat?
  A: N
*Turn 12:
  Q: Do you live with 4 or more people?
  A: Y
*Turn 13:
  Q: Do you have a fever (either felt or measured with a thermometer)?
  A: N
*Turn 14:
  Q: Have you been in contact with a person with similar symptoms in the past 2 weeks?
  A: Y
*Turn 15:
  Q: Do you have diffuse (widespread) muscle pain?
  A: N
*Turn 16:
  Q: Have you had significantly increased sweating?
  A: Y
*Turn 17:
  Q: Do you attend or work in a daycare?
  A: Y
*Turn 18:
  Q: Do you smoke cigarettes?
  A: N
Predicted differential:
----------------------
Bronchitis, URTI, Cluster headache, Chronic rhinosinusitis, Acute rhinosinusitis, Chagas, Anemia
Ground truth differential:
-------------------------
URTI: 0.230, Bronchitis: 0.223, Cluster headache: 0.186, Acute rhinosinusitis: 0.130, Chagas: 0.086, Anemia: 0.086, Chronic rhinosinusitis: 0.059
\end{lstlisting}

\subsection*{Patient 9}

\begin{lstlisting}
Sex: M, Age: 81
Initial evidence: 
	 I feel pain
Agent inquiries:
---------------
*Turn 1:
  Q: Do you have swelling in one or more areas of your body?
  A: N
*Turn 2:
  Q: Do you have any lesions, redness or problems on your skin that you believe are related to the condition you are consulting for?
  A: N
*Turn 3:
  Q: Do you feel pain somewhere?
  A:
	 * tonsil(R)
	 * tonsil(L)
	 * palace
	 * under the jaw
*Turn 4:
  Q: Characterize your pain:
  A:
	 * sensitive
	 * burning
*Turn 5:
  Q: How precisely is the pain located?
  A:
	 * 10
*Turn 6:
  Q: How intense is the pain?
  A:
	 * 5
*Turn 7:
  Q: How fast did the pain appear?
  A:
	 * 1
*Turn 8:
  Q: Do you have nasal congestion or a clear runny nose?
  A: Y
*Turn 9:
  Q: Have you been in contact with a person with similar symptoms in the past 2 weeks?
  A: Y
*Turn 10:
  Q: Do you have a fever (either felt or measured with a thermometer)?
  A: Y
*Turn 11:
  Q: Do you attend or work in a daycare?
  A: Y
*Turn 12:
  Q: Do you have a cough?
  A: N
*Turn 13:
  Q: Are you immunosuppressed?
  A: N
*Turn 14:
  Q: Do you live with 4 or more people?
  A: N
*Turn 15:
  Q: Have you been coughing up blood?
  A: N
*Turn 16:
  Q: Do you smoke cigarettes?
  A: Y
Predicted differential:
----------------------
Viral pharyngitis, Possible NSTEMI / STEMI, Epiglottitis, Unstable angina, Acute laryngitis, URTI, Chagas, Stable angina
Ground truth differential:
-------------------------
Viral pharyngitis: 0.254, Possible NSTEMI / STEMI: 0.158, Unstable angina: 0.148, Epiglottitis: 0.144, URTI: 0.090, Acute laryngitis: 0.082, Chagas: 0.070, Stable angina: 0.054
\end{lstlisting}

\subsection*{Patient 10}

\begin{lstlisting}
Sex: F, Age: 76
Initial evidence: 
	 I am coughing
Agent inquiries:
---------------
*Turn 1:
  Q: Do you have any lesions, redness or problems on your skin that you believe are related to the condition you are consulting for?
  A: N
*Turn 2:
  Q: Do you have pain somewhere, related to your reason for consulting?
  A: Y
*Turn 3:
  Q: Do you feel pain somewhere?
  A:
	 * posterior chest wall(L)
*Turn 4:
  Q: Characterize your pain:
  A:
	 * a knife stroke
	 * exhausting
*Turn 5:
  Q: How precisely is the pain located?
  A:
	 * 0
*Turn 6:
  Q: How intense is the pain?
  A:
	 * 1
*Turn 7:
  Q: How fast did the pain appear?
  A:
	 * 0
*Turn 8:
  Q: Have you been coughing up blood?
  A: Y
*Turn 9:
  Q: Have you had an involuntary weight loss over the last 3 months?
  A: Y
*Turn 10:
  Q: Are you a former smoker?
  A: Y
*Turn 11:
  Q: Do you smoke cigarettes?
  A: Y
*Turn 12:
  Q: Are you exposed to secondhand cigarette smoke on a daily basis?
  A: N
*Turn 13:
  Q: Do you have family members who have had lung cancer?
  A: Y
*Turn 14:
  Q: Do you constantly feel fatigued or do you have non-restful sleep?
  A: Y
*Turn 15:
  Q: Are you experiencing shortness of breath or difficulty breathing in a significant way?
  A: Y
*Turn 16:
  Q: Do you have an active cancer?
  A: N
*Turn 17:
  Q: Have you recently had a loss of appetite or do you get full more quickly then usually?
  A: N
*Turn 18:
  Q: In the last month, have you been in contact with anyone infected with the Ebola virus?
  A: N
*Turn 19:
  Q: Do you have swelling in one or more areas of your body?
  A: N
*Turn 20:
  Q: Do you feel slightly dizzy or lightheaded?
  A: N
Predicted differential:
----------------------
Pulmonary neoplasm, Tuberculosis, Acute pulmonary edema, Possible NSTEMI / STEMI, Bronchitis, Pneumonia, Pulmonary embolism, Unstable angina, Myocarditis, Anemia, Bronchiectasis, Bronchospasm / acute asthma exacerbation, Stable angina, Atrial fibrillation, Guillain-Barre syndrome, Spontaneous rib fracture, Pancreatic neoplasm, Myasthenia gravis, Acute dystonic reactions
Ground truth differential:
-------------------------
Pulmonary neoplasm: 0.091, Tuberculosis: 0.074, Bronchitis: 0.069, Acute pulmonary edema: 0.069, Pneumonia: 0.066, Pulmonary embolism: 0.064, Possible NSTEMI / STEMI: 0.059, Unstable angina: 0.057, Bronchiectasis: 0.050, Myocarditis: 0.050, Anemia: 0.048, Stable angina: 0.046, Guillain-Barre syndrome: 0.044, Bronchospasm / acute asthma exacerbation: 0.043, Atrial fibrillation: 0.042, Spontaneous rib fracture: 0.038, Acute dystonic reactions: 0.031, Myasthenia gravis: 0.031, Pancreatic neoplasm: 0.027
\end{lstlisting}

\subsection*{Patient 11}

\begin{lstlisting}
Sex: M, Age: 49
Initial evidence: 
	 I feel pain
Agent inquiries:
---------------
*Turn 1:
  Q: Do you have swelling in one or more areas of your body?
  A: N
*Turn 2:
  Q: Do you have any lesions, redness or problems on your skin that you believe are related to the condition you are consulting for?
  A: Y
*Turn 3:
  Q: What color is the rash?
  A:
	 * yellow
*Turn 4:
  Q: Characterize your pain:
  A:
	 * a cramp
*Turn 5:
  Q: Where is the affected region located?
  A:
	 * epigastric
*Turn 6:
  Q: Do you feel pain somewhere?
  A:
	 * flank(L)
	 * hypochondrium(L)
	 * epigastric
*Turn 7:
  Q: How intense is the pain?
  A:
	 * 4
*Turn 8:
  Q: How precisely is the pain located?
  A:
	 * 5
*Turn 9:
  Q: How severe is the itching?
  A:
	 * 3
*Turn 10:
  Q: How intense is the pain caused by the rash?
  A:
	 * 0
*Turn 11:
  Q: Is the rash swollen?
  A:
	 * 2
*Turn 12:
  Q: Is the lesion (or are the lesions) larger than 1cm?
  A:
	 * Y
*Turn 13:
  Q: How fast did the pain appear?
  A:
	 * 1
*Turn 14:
  Q: Are there members of your family who have been diagnosed with pancreatic cancer?
  A: Y
*Turn 15:
  Q: Does the pain radiate to another location?
  A:
	 * thoracic spine
*Turn 16:
  Q: Do you have chronic pancreatitis?
  A: Y
*Turn 17:
  Q: Are you significantly overweight compared to people of the same height as you?
  A: Y
*Turn 18:
  Q: Do you have diabetes?
  A: Y
*Turn 19:
  Q: Do you constantly feel fatigued or do you have non-restful sleep?
  A: Y
*Turn 20:
  Q: Do you smoke cigarettes?
  A: N
*Turn 21:
  Q: Do you have pale stools and dark urine?
  A: Y
*Turn 22:
  Q: Do you have a cough?
  A: N
*Turn 23:
  Q: Have you had an involuntary weight loss over the last 3 months?
  A: N
*Turn 24:
  Q: Are you feeling nauseous or do you feel like vomiting?
  A: Y
*Turn 25:
  Q: Have you had diarrhea or an increase in stool frequency?
  A: Y
Predicted differential:
----------------------
Boerhaave, Pancreatic neoplasm, Anaphylaxis, Scombroid food poisoning, Possible NSTEMI / STEMI, Unstable angina, GERD, Chagas
Ground truth differential:
-------------------------
Pancreatic neoplasm: 0.190, Unstable angina: 0.150, Anaphylaxis: 0.132, Boerhaave: 0.131, Scombroid food poisoning: 0.126, Possible NSTEMI / STEMI: 0.096, Chagas: 0.090, GERD: 0.084
\end{lstlisting}

\subsection*{Patient 12}

\begin{lstlisting}
Sex: M, Age: 62
Initial evidence: 
	 I feel pain
Agent inquiries:
---------------
*Turn 1:
  Q: Do you have swelling in one or more areas of your body?
  A: N
*Turn 2:
  Q: Do you have any lesions, redness or problems on your skin that you believe are related to the condition you are consulting for?
  A: N
*Turn 3:
  Q: Does the pain radiate to another location?
  A:
	 * nowhere
*Turn 4:
  Q: Do you feel pain somewhere?
  A:
	 * tonsil(L)
	 * thyroid cartilage
	 * palace
	 * under the jaw
*Turn 5:
  Q: Characterize your pain:
  A:
	 * sensitive
	 * burning
*Turn 6:
  Q: How precisely is the pain located?
  A:
	 * 10
*Turn 7:
  Q: How intense is the pain?
  A:
	 * 6
*Turn 8:
  Q: How fast did the pain appear?
  A:
	 * 2
*Turn 9:
  Q: Do you attend or work in a daycare?
  A: Y
*Turn 10:
  Q: Do you live with 4 or more people?
  A: Y
*Turn 11:
  Q: Have you been in contact with a person with similar symptoms in the past 2 weeks?
  A: Y
*Turn 12:
  Q: Do you have a fever (either felt or measured with a thermometer)?
  A: Y
*Turn 13:
  Q: Do you have a cough?
  A: Y
*Turn 14:
  Q: Are you immunosuppressed?
  A: N
*Turn 15:
  Q: Do you have nasal congestion or a clear runny nose?
  A: N
*Turn 16:
  Q: Have you been coughing up blood?
  A: Y
*Turn 17:
  Q: Do you smoke cigarettes?
  A: N
Predicted differential:
----------------------
Bronchitis, Tuberculosis, Viral pharyngitis, Acute laryngitis, Possible NSTEMI / STEMI, Unstable angina, Pneumonia, Influenza, Stable angina
Ground truth differential:
-------------------------
Bronchitis: 0.197, Tuberculosis: 0.183, Ebola: 0.152, Acute laryngitis: 0.131, Viral pharyngitis: 0.130, Possible NSTEMI / STEMI: 0.107, Unstable angina: 0.100
\end{lstlisting}

\subsection*{Patient 13}

\begin{lstlisting}
Sex: F, Age: 60
Initial evidence: 
	 I have a sore throat
Agent inquiries:
---------------
*Turn 1:
  Q: Do you have any lesions, redness or problems on your skin that you believe are related to the condition you are consulting for?
  A: N
*Turn 2:
  Q: Do you have pain somewhere, related to your reason for consulting?
  A: Y
*Turn 3:
  Q: Does the pain radiate to another location?
  A:
	 * nowhere
*Turn 4:
  Q: Do you feel pain somewhere?
  A:
	 * top of the head
	 * forehead
	 * cheek(R)
	 * occiput
	 * temple(R)
*Turn 5:
  Q: Characterize your pain:
  A:
	 * sensitive
	 * heavy
*Turn 6:
  Q: How precisely is the pain located?
  A:
	 * 3
*Turn 7:
  Q: How intense is the pain?
  A:
	 * 8
*Turn 8:
  Q: How fast did the pain appear?
  A:
	 * 3
*Turn 9:
  Q: Do you have nasal congestion or a clear runny nose?
  A: N
*Turn 10:
  Q: Do you live with 4 or more people?
  A: Y
*Turn 11:
  Q: Do you attend or work in a daycare?
  A: Y
*Turn 12:
  Q: Do you have a cough?
  A: Y
*Turn 13:
  Q: Do you have a cough that produces colored or more abundant sputum than usual?
  A: N
*Turn 14:
  Q: Do you have a fever (either felt or measured with a thermometer)?
  A: N
*Turn 15:
  Q: Do you have diffuse (widespread) muscle pain?
  A: Y
*Turn 16:
  Q: Have you been in contact with a person with similar symptoms in the past 2 weeks?
  A: N
*Turn 17:
  Q: Are you exposed to secondhand cigarette smoke on a daily basis?
  A: Y
*Turn 18:
  Q: Have you had significantly increased sweating?
  A: Y
*Turn 19:
  Q: Do you smoke cigarettes?
  A: Y
Predicted differential:
----------------------
Bronchitis, URTI, Influenza, Cluster headache, Anemia, Chagas
Ground truth differential:
-------------------------
Bronchitis: 0.260, URTI: 0.255, Cluster headache: 0.153, Influenza: 0.144, Chagas: 0.094, Anemia: 0.094
\end{lstlisting}

\subsection*{Patient 14}

\begin{lstlisting}
Sex: F, Age: 63
Initial evidence: 
	 I have nasal congestion
Agent inquiries:
---------------
*Turn 1:
  Q: Do you have pain somewhere, related to your reason for consulting?
  A: Y
*Turn 2:
  Q: Do you have swelling in one or more areas of your body?
  A: N
*Turn 3:
  Q: Do you have any lesions, redness or problems on your skin that you believe are related to the condition you are consulting for?
  A: N
*Turn 4:
  Q: Do you feel pain somewhere?
  A:
	 * top of the head
	 * forehead
	 * cheek(L)
	 * occiput
	 * temple(L)
*Turn 5:
  Q: Characterize your pain:
  A:
	 * tedious
*Turn 6:
  Q: How precisely is the pain located?
  A:
	 * 6
*Turn 7:
  Q: How intense is the pain?
  A:
	 * 4
*Turn 8:
  Q: How fast did the pain appear?
  A:
	 * 3
*Turn 9:
  Q: Do you live with 4 or more people?
  A: Y
*Turn 10:
  Q: Do you have a sore throat?
  A: Y
*Turn 11:
  Q: Do you have a cough?
  A: Y
*Turn 12:
  Q: Are you exposed to secondhand cigarette smoke on a daily basis?
  A: Y
*Turn 13:
  Q: Have you been in contact with a person with similar symptoms in the past 2 weeks?
  A: Y
*Turn 14:
  Q: Do you smoke cigarettes?
  A: Y
*Turn 15:
  Q: Do you attend or work in a daycare?
  A: Y
*Turn 16:
  Q: Do you have a cough that produces colored or more abundant sputum than usual?
  A: N
*Turn 17:
  Q: Do you have a fever (either felt or measured with a thermometer)?
  A: Y
*Turn 18:
  Q: Have you had significantly increased sweating?
  A: Y
*Turn 19:
  Q: Do you have diffuse (widespread) muscle pain?
  A: N
Predicted differential:
----------------------
URTI, Influenza, HIV (initial infection), Bronchitis, Tuberculosis, Chronic rhinosinusitis, Chagas, Acute rhinosinusitis
Ground truth differential:
-------------------------
URTI: 0.218, Influenza: 0.174, HIV (initial infection): 0.164, Bronchitis: 0.155, Tuberculosis: 0.119, Chronic rhinosinusitis: 0.071, Chagas: 0.052, Acute rhinosinusitis: 0.046
\end{lstlisting}

\subsection*{Patient 15}

\begin{lstlisting}
Sex: F, Age: 14
Initial evidence: 
	 I have diffuse muscle pain
Agent inquiries:
---------------
*Turn 1:
  Q: Do you have any lesions, redness or problems on your skin that you believe are related to the condition you are consulting for?
  A: N
*Turn 2:
  Q: Do you have pain somewhere, related to your reason for consulting?
  A: Y
*Turn 3:
  Q: Does the pain radiate to another location?
  A:
	 * nowhere
*Turn 4:
  Q: Characterize your pain:
  A:
	 * tedious
	 * sensitive
	 * heavy
*Turn 5:
  Q: How precisely is the pain located?
  A:
	 * 5
*Turn 6:
  Q: Do you feel pain somewhere?
  A:
	 * top of the head
	 * forehead
	 * cheek(R)
	 * cheek(L)
	 * occiput
*Turn 7:
  Q: How intense is the pain?
  A:
	 * 7
*Turn 8:
  Q: How fast did the pain appear?
  A:
	 * 5
*Turn 9:
  Q: Do you have a cough that produces colored or more abundant sputum than usual?
  A: N
*Turn 10:
  Q: Do you have a cough?
  A: Y
*Turn 11:
  Q: Do you have a fever (either felt or measured with a thermometer)?
  A: N
*Turn 12:
  Q: Do you live with 4 or more people?
  A: N
*Turn 13:
  Q: Do you have a sore throat?
  A: Y
*Turn 14:
  Q: Do you smoke cigarettes?
  A: Y
*Turn 15:
  Q: Have you been in contact with a person with similar symptoms in the past 2 weeks?
  A: N
*Turn 16:
  Q: Are you exposed to secondhand cigarette smoke on a daily basis?
  A: N
*Turn 17:
  Q: Do you have nasal congestion or a clear runny nose?
  A: Y
*Turn 18:
  Q: Do you attend or work in a daycare?
  A: Y
*Turn 19:
  Q: Have you had significantly increased sweating?
  A: N
Predicted differential:
----------------------
URTI, Bronchitis, Influenza, Cluster headache, Chronic rhinosinusitis, Chagas, Anemia, Acute rhinosinusitis, Pneumonia
Ground truth differential:
-------------------------
URTI: 0.230, Cluster headache: 0.174, Bronchitis: 0.167, Chronic rhinosinusitis: 0.141, Acute rhinosinusitis: 0.101, Chagas: 0.066, Anemia: 0.066, Influenza: 0.054
\end{lstlisting}

\subsection*{Patient 16}

\begin{lstlisting}
Sex: M, Age: 49
Initial evidence: 
	 I have nasal congestion
Agent inquiries:
---------------
*Turn 1:
  Q: Do you have swelling in one or more areas of your body?
  A: N
*Turn 2:
  Q: Do you have any lesions, redness or problems on your skin that you believe are related to the condition you are consulting for?
  A: N
*Turn 3:
  Q: Do you have pain somewhere, related to your reason for consulting?
  A: Y
*Turn 4:
  Q: Does the pain radiate to another location?
  A:
	 * nowhere
*Turn 5:
  Q: Characterize your pain:
  A:
	 * sensitive
*Turn 6:
  Q: Do you feel pain somewhere?
  A:
	 * top of the head
	 * forehead
	 * cheek(R)
	 * occiput
	 * temple(R)
*Turn 7:
  Q: How precisely is the pain located?
  A:
	 * 7
*Turn 8:
  Q: How intense is the pain?
  A:
	 * 2
*Turn 9:
  Q: How fast did the pain appear?
  A:
	 * 5
*Turn 10:
  Q: Do you have a sore throat?
  A: N
*Turn 11:
  Q: Do you live with 4 or more people?
  A: N
*Turn 12:
  Q: Do you attend or work in a daycare?
  A: Y
*Turn 13:
  Q: Do you have a cough that produces colored or more abundant sputum than usual?
  A: N
*Turn 14:
  Q: Do you have a cough?
  A: Y
*Turn 15:
  Q: Have you been in contact with a person with similar symptoms in the past 2 weeks?
  A: Y
*Turn 16:
  Q: Are you exposed to secondhand cigarette smoke on a daily basis?
  A: N
*Turn 17:
  Q: Do you have a fever (either felt or measured with a thermometer)?
  A: Y
*Turn 18:
  Q: Do you have diffuse (widespread) muscle pain?
  A: N
*Turn 19:
  Q: Have you had significantly increased sweating?
  A: Y
*Turn 20:
  Q: Do you smoke cigarettes?
  A: N
Predicted differential:
----------------------
URTI, Bronchitis, Influenza, Tuberculosis, Chronic rhinosinusitis, Chagas, Acute rhinosinusitis
Ground truth differential:
-------------------------
URTI: 0.236, Influenza: 0.202, Bronchitis: 0.201, Tuberculosis: 0.149, Chronic rhinosinusitis: 0.089, Chagas: 0.065, Acute rhinosinusitis: 0.058
\end{lstlisting}

\subsection*{Patient 17}

\begin{lstlisting}
Sex: F, Age: 34
Initial evidence: 
	 I feel pain
Agent inquiries:
---------------
*Turn 1:
  Q: Do you have any lesions, redness or problems on your skin that you believe are related to the condition you are consulting for?
  A: N
*Turn 2:
  Q: Does the pain radiate to another location?
  A:
	 * nowhere
*Turn 3:
  Q: Do you feel pain somewhere?
  A:
	 * lower chest
	 * side of the chest(L)
	 * upper chest
	 * breast(L)
	 * posterior chest wall(R)
*Turn 4:
  Q: Characterize your pain:
  A:
	 * sensitive
*Turn 5:
  Q: How precisely is the pain located?
  A:
	 * 7
*Turn 6:
  Q: How intense is the pain?
  A:
	 * 5
*Turn 7:
  Q: Are you experiencing shortness of breath or difficulty breathing in a significant way?
  A: N
*Turn 8:
  Q: How fast did the pain appear?
  A:
	 * 0
*Turn 9:
  Q: Have you ever had pneumonia?
  A: N
*Turn 10:
  Q: Do you smoke cigarettes?
  A: Y
*Turn 11:
  Q: Do you have a cough?
  A: Y
*Turn 12:
  Q: Do you have a chronic obstructive pulmonary disease (COPD)?
  A: Y
*Turn 13:
  Q: Do you have nasal congestion or a clear runny nose?
  A: Y
*Turn 14:
  Q: Do you attend or work in a daycare?
  A: N
*Turn 15:
  Q: Do you have a cough that produces colored or more abundant sputum than usual?
  A: Y
*Turn 16:
  Q: Do you have a fever (either felt or measured with a thermometer)?
  A: N
*Turn 17:
  Q: Are your symptoms more prominent at night?
  A: N
*Turn 18:
  Q: Do you have a sore throat?
  A: N
*Turn 19:
  Q: Have you noticed a wheezing sound when you exhale?
  A: Y
*Turn 20:
  Q: Are your vaccinations up to date?
  A: N
Predicted differential:
----------------------
Bronchitis, Acute COPD exacerbation / infection, GERD, Pneumonia, Bronchiectasis, Bronchospasm / acute asthma exacerbation, Spontaneous rib fracture, Possible NSTEMI / STEMI, Unstable angina, Boerhaave, URTI, Pericarditis, Stable angina, Allergic sinusitis, Scombroid food poisoning
Ground truth differential:
-------------------------
Acute COPD exacerbation / infection: 0.121, Bronchitis: 0.118, GERD: 0.092, Pneumonia: 0.090, Bronchospasm / acute asthma exacerbation: 0.084, Bronchiectasis: 0.079, Spontaneous rib fracture: 0.074, Possible NSTEMI / STEMI: 0.070, Unstable angina: 0.065, Boerhaave: 0.062, URTI: 0.045, Pericarditis: 0.043, Stable angina: 0.036, Scombroid food poisoning: 0.020
\end{lstlisting}

\subsection*{Patient 18}

\begin{lstlisting}
Sex: F, Age: 18
Initial evidence: 
	 I feel pain
Agent inquiries:
---------------
*Turn 1:
  Q: Do you have any lesions, redness or problems on your skin that you believe are related to the condition you are consulting for?
  A: N
*Turn 2:
  Q: Does the pain radiate to another location?
  A:
	 * nowhere
*Turn 3:
  Q: Do you feel pain somewhere?
  A:
	 * lower chest
	 * side of the chest(L)
	 * breast(L)
*Turn 4:
  Q: Characterize your pain:
  A:
	 * heartbreaking
	 * a knife stroke
*Turn 5:
  Q: How precisely is the pain located?
  A:
	 * 6
*Turn 6:
  Q: How intense is the pain?
  A:
	 * 8
*Turn 7:
  Q: Do you have pain that is increased when you breathe in deeply?
  A: N
*Turn 8:
  Q: Are you experiencing shortness of breath or difficulty breathing in a significant way?
  A: N
*Turn 9:
  Q: Have you ever had a spontaneous pneumothorax?
  A: Y
*Turn 10:
  Q: How fast did the pain appear?
  A:
	 * 7
*Turn 11:
  Q: Do you smoke cigarettes?
  A: Y
*Turn 12:
  Q: Do you have nasal congestion or a clear runny nose?
  A: N
*Turn 13:
  Q: Do you have symptoms that are increased with physical exertion but alleviated with rest?
  A: Y
*Turn 14:
  Q: Do you have chest pain even at rest?
  A: N
*Turn 15:
  Q: Do you feel your heart is beating fast (racing), irregularly (missing a beat) or do you feel palpitations?
  A: N
Predicted differential:
----------------------
Unstable angina, Stable angina, Possible NSTEMI / STEMI, Spontaneous pneumothorax, Atrial fibrillation, Pericarditis
Ground truth differential:
-------------------------
Unstable angina: 0.272, Stable angina: 0.243, Possible NSTEMI / STEMI: 0.199, Pericarditis: 0.124, Atrial fibrillation: 0.097, Spontaneous pneumothorax: 0.064
\end{lstlisting}

\subsection*{Patient 19}

\begin{lstlisting}
Sex: F, Age: 35
Initial evidence: 
	 I feel pain
Agent inquiries:
---------------
*Turn 1:
  Q: Do you have any lesions, redness or problems on your skin that you believe are related to the condition you are consulting for?
  A: N
*Turn 2:
  Q: Does the pain radiate to another location?
  A:
	 * lower chest
*Turn 3:
  Q: Characterize your pain:
  A:
	 * haunting
	 * sensitive
	 * a knife stroke
	 * tugging
	 * burning
	 * sickening
*Turn 4:
  Q: Do you feel pain somewhere?
  A:
	 * lower chest
	 * upper chest
	 * hypochondrium(R)
	 * hypochondrium(L)
	 * epigastric
*Turn 5:
  Q: How precisely is the pain located?
  A:
	 * 6
*Turn 6:
  Q: How intense is the pain?
  A:
	 * 3
*Turn 7:
  Q: Are your symptoms worse when lying down and alleviated while sitting up?
  A: Y
*Turn 8:
  Q: Do you have a cough?
  A: N
*Turn 9:
  Q: Do you have a hiatal hernia?
  A: N
*Turn 10:
  Q: Do you have symptoms that get worse after eating?
  A: N
*Turn 11:
  Q: Do you have a burning sensation that starts in your stomach then goes up into your throat, and can be associated with a bitter taste in your mouth?
  A: Y
*Turn 12:
  Q: How fast did the pain appear?
  A:
	 * 0
*Turn 13:
  Q: Do you have asthma or have you ever had to use a bronchodilator in the past?
  A: N
*Turn 14:
  Q: Do you smoke cigarettes?
  A: Y
*Turn 15:
  Q: Are you experiencing shortness of breath or difficulty breathing in a significant way?
  A: N
*Turn 16:
  Q: Do you think you are pregnant or are you currently pregnant?
  A: Y
*Turn 17:
  Q: Have you recently thrown up blood or something resembling coffee beans?
  A: N
*Turn 18:
  Q: Do you drink alcohol excessively or do you have an addiction to alcohol?
  A: Y
*Turn 19:
  Q: Are you significantly overweight compared to people of the same height as you?
  A: N
*Turn 20:
  Q: Have you recently had stools that were black (like coal)?
  A: N
Predicted differential:
----------------------
GERD, Possible NSTEMI / STEMI, Pericarditis, Unstable angina, Stable angina, Pancreatic neoplasm, Anaphylaxis
Ground truth differential:
-------------------------
GERD: 0.297, Possible NSTEMI / STEMI: 0.221, Pericarditis: 0.151, Unstable angina: 0.145, Stable angina: 0.116, Anaphylaxis: 0.070
\end{lstlisting}

\subsection*{Patient 20}

\begin{lstlisting}
Sex: F, Age: 43
Initial evidence: 
	 I have had significantly increased sweating
Agent inquiries:
---------------
*Turn 1:
  Q: Do you have any lesions, redness or problems on your skin that you believe are related to the condition you are consulting for?
  A: N
*Turn 2:
  Q: Do you have pain somewhere, related to your reason for consulting?
  A: Y
*Turn 3:
  Q: Does the pain radiate to another location?
  A:
	 * biceps(L)
	 * thyroid cartilage
	 * thoracic spine
	 * shoulder(R)
	 * shoulder(L)
*Turn 4:
  Q: Do you feel pain somewhere?
  A:
	 * biceps(R)
	 * thoracic spine
	 * side of the chest(L)
	 * upper chest
	 * epigastric
*Turn 5:
  Q: Characterize your pain:
  A:
	 * tedious
	 * heavy
*Turn 6:
  Q: How precisely is the pain located?
  A:
	 * 4
*Turn 7:
  Q: How intense is the pain?
  A:
	 * 5
*Turn 8:
  Q: How fast did the pain appear?
  A:
	 * 5
*Turn 9:
  Q: Have you ever had a heart attack or do you have angina (chest pain)?
  A: Y
*Turn 10:
  Q: Do you have diabetes?
  A: Y
*Turn 11:
  Q: Are you experiencing shortness of breath or difficulty breathing in a significant way?
  A: Y
*Turn 12:
  Q: Do you have close family members who had a cardiovascular disease problem before the age of 50?
  A: Y
*Turn 13:
  Q: Do you have high blood pressure or do you take medications to treat high blood pressure?
  A: Y
*Turn 14:
  Q: Are you significantly overweight compared to people of the same height as you?
  A: Y
*Turn 15:
  Q: Do you have high cholesterol or do you take medications to treat high cholesterol?
  A: Y
*Turn 16:
  Q: Do you have swelling in one or more areas of your body?
  A: N
*Turn 17:
  Q: Do you smoke cigarettes?
  A: Y
*Turn 18:
  Q: Do you exercise regularly, 4 times per week or more?
  A: Y
*Turn 19:
  Q: Do you drink alcohol excessively or do you have an addiction to alcohol?
  A: N
*Turn 20:
  Q: Do you have symptoms that are increased with physical exertion but alleviated with rest?
  A: N
*Turn 21:
  Q: Are you feeling nauseous or do you feel like vomiting?
  A: N
*Turn 22:
  Q: Do you constantly feel fatigued or do you have non-restful sleep?
  A: N
*Turn 23:
  Q: Do you have chest pain even at rest?
  A: N
*Turn 24:
  Q: Do you find that your symptoms have worsened over the last 2 weeks and that progressively less effort is required to cause the symptoms?
  A: N
*Turn 25:
  Q: Are you infected with the human immunodeficiency virus (HIV)?
  A: N
Predicted differential:
----------------------
Possible NSTEMI / STEMI, SLE, GERD, Guillain-Barre syndrome, Atrial fibrillation, Myocarditis, Anemia, Acute dystonic reactions, Pericarditis, Panic attack, Acute pulmonary edema, Myasthenia gravis, Anaphylaxis, Pancreatic neoplasm, Scombroid food poisoning, Pulmonary neoplasm, PSVT, Chagas
Ground truth differential:
-------------------------
Possible NSTEMI / STEMI: 0.127, SLE: 0.087, GERD: 0.086, Guillain-Barre syndrome: 0.081, Atrial fibrillation: 0.077, Panic attack: 0.065, Myocarditis: 0.063, Pericarditis: 0.058, Acute dystonic reactions: 0.057, Myasthenia gravis: 0.057, Anemia: 0.057, Acute pulmonary edema: 0.053, Pancreatic neoplasm: 0.043, Scombroid food poisoning: 0.036, PSVT: 0.030, Pulmonary neoplasm: 0.013, Unstable angina: 0.009
\end{lstlisting}

\section{Example interactions}
\label{sec:exampleInteractions}

We showcase three example interactions of CASANDE with synthetic patients from the DDXPlus test set. For each example, we show the sex, age and initial symptom of the patient, followed by the questions asked by CASANDE, the answers given by the patient, and the differential diagnosis at each turn. Finally, we show the ground truth differential diagnosis. Diseases that are severe are marked with \texttt{(severe)} after their names.

In all three examples, we observe a common pattern: the predicted differential diagnosis tends to be larger and changes more frequently at the beginning of the interaction than at the end. We also see several questions at the end with the same differential. This is aligned with the desired exploration-confirmation behavior. Notably, there are often severe pathologies present in the predicted differential at some turns which are later excluded. This could indicate the agent is actively considering those diseases and gathering evidence to rule them out.

\subsection*{Example 1}

\begin{lstlisting}
Sex: F, age: 31
Initial evidence: I feel slightly dizzy or lightheaded.
Differential diagnosis: ['Anemia', 'Atrial fibrillation', 'Cluster headache', 'HIV (initial infection)']
----------
Turn 1:
Q: Have you traveled out of the country in the last 4 weeks?
A: North America
Differential diagnosis: ['Anemia', 'Atrial fibrillation', 'Cluster headache', 'Pulmonary embolism (severe)', 'Anaphylaxis (severe)', 'Guillain-Barre syndrome (severe)', 'Acute dystonic reactions (severe)', 'Acute pulmonary edema (severe)', 'HIV (initial infection)', 'Myocarditis (severe)', 'Myasthenia gravis', 'PSVT (severe)', 'Chagas', 'Scombroid food poisoning (severe)', 'Pulmonary neoplasm', 'SLE']

Turn 2:
Q: Do you have pain somewhere, related to your reason for consulting?
A: No
Differential diagnosis: ['Atrial fibrillation', 'Panic attack', 'Cluster headache', 'Myocarditis (severe)', 'Anemia', 'Pulmonary embolism (severe)', 'PSVT (severe)', 'Guillain-Barre syndrome (severe)', 'Acute pulmonary edema (severe)', 'Anaphylaxis (severe)', 'Acute dystonic reactions (severe)', 'Myasthenia gravis', 'Scombroid food poisoning (severe)', 'SLE', 'HIV (initial infection)', 'Bronchospasm / acute asthma exacerbation', 'Pulmonary neoplasm', 'Acute COPD exacerbation / infection']

Turn 3:
Q: Do you feel your heart is beating very irregularly or in a disorganized pattern?
A: Yes
Differential diagnosis: ['Atrial fibrillation', 'PSVT (severe)', 'Pulmonary embolism (severe)', 'Anemia', 'Anaphylaxis (severe)', 'Acute dystonic reactions (severe)', 'Acute pulmonary edema (severe)', 'Scombroid food poisoning (severe)', 'Guillain-Barre syndrome (severe)', 'Myasthenia gravis', 'Myocarditis (severe)', 'Bronchospasm / acute asthma exacerbation', 'Acute COPD exacerbation / infection', 'Panic attack']

Turn 4:
Q: Do you have any lesions, redness or problems on your skin that you believe are related to the condition you are consulting for?
A: No
Differential diagnosis: ['Atrial fibrillation', 'PSVT (severe)', 'Anemia', 'Scombroid food poisoning (severe)']

Turn 5:
Q: Are you experiencing shortness of breath or difficulty breathing in a significant way?
A: No
Differential diagnosis: ['Atrial fibrillation', 'PSVT (severe)', 'Anemia', 'Scombroid food poisoning (severe)']

Turn 6:
Q: Do you have severe Chronic Obstructive Pulmonary Disease (COPD)?
A: Yes
Differential diagnosis: ['Atrial fibrillation', 'PSVT (severe)', 'Anemia', 'Scombroid food poisoning (severe)']

Turn 7:
Q: Do you have high blood pressure or do you take medications to treat high blood pressure?
A: No
Differential diagnosis: ['Atrial fibrillation', 'PSVT (severe)', 'Anemia', 'Scombroid food poisoning (severe)']

Turn 8:
Q: Do you have a known issue with one of your heart valves?
A: Yes
Differential diagnosis: ['Atrial fibrillation', 'PSVT (severe)', 'Anemia', 'Scombroid food poisoning (severe)']

Turn 9:
Q: Have you ever had a heart attack or do you have angina (chest pain)?
A: Yes
Differential diagnosis: ['Atrial fibrillation', 'PSVT (severe)', 'Anemia', 'Scombroid food poisoning (severe)']

Turn 10:
Q: Have you been diagnosed with hyperthyroidism?
A: Yes
Differential diagnosis: ['Atrial fibrillation', 'PSVT (severe)', 'Anemia', 'Scombroid food poisoning (severe)']

Turn 11:
Q: Have you ever had a pericarditis?
A: No
Differential diagnosis: ['Atrial fibrillation', 'PSVT (severe)', 'Anemia', 'Scombroid food poisoning (severe)']

Turn 12:
Q: Are you significantly overweight compared to people of the same height as you?
A: No
Differential diagnosis: ['Atrial fibrillation', 'PSVT (severe)', 'Anemia', 'Scombroid food poisoning (severe)']

Turn 13:
Q: Do you feel your heart is beating fast (racing), irregularly (missing a beat) or do you feel palpitations?
A: Yes
Differential diagnosis: ['Atrial fibrillation', 'PSVT (severe)', 'Anemia', 'Scombroid food poisoning (severe)']

Turn 14:
Q: Do you have a known heart defect?
A: Yes
Differential diagnosis: ['Atrial fibrillation', 'PSVT (severe)', 'Anemia', 'Scombroid food poisoning (severe)']

Turn 15:
Q: Do you have diabetes?
A: Yes
Differential diagnosis: ['Atrial fibrillation', 'PSVT (severe)', 'Anemia', 'Scombroid food poisoning (severe)']

Turn 16:
Q: Do you have symptoms that are increased with physical exertion but alleviated with rest?
A: No
Differential diagnosis: ['Atrial fibrillation', 'PSVT (severe)', 'Anemia', 'Scombroid food poisoning (severe)']

----------
Ground truth differential diagnosis: ['Atrial fibrillation: 0.416', 'PSVT (severe): 0.331', 'Anemia: 0.165', 'Scombroid food poisoning (severe): 0.088']


\end{lstlisting}

\subsection*{Example 2}

\begin{lstlisting}
Sex: M, age: 20
Initial evidence: I have pain that is increased when I breathe in deeply.
Differential diagnosis: ['Pulmonary embolism (severe)', 'Spontaneous pneumothorax (severe)', 'Panic attack', 'Pericarditis', 'Guillain-Barre syndrome (severe)', 'Boerhaave (severe)', 'Myocarditis (severe)', 'Possible NSTEMI / STEMI (severe)', 'Anaphylaxis (severe)', 'GERD', 'Atrial fibrillation', 'Myasthenia gravis', 'Acute pulmonary edema (severe)', 'Anemia', 'Acute dystonic reactions (severe)', 'Scombroid food poisoning (severe)', 'Sarcoidosis', 'HIV (initial infection)', 'Chagas', 'Pancreatic neoplasm', 'PSVT (severe)', 'SLE', 'Unstable angina (severe)']
----------
Turn 1:
Q: Do you have any lesions, redness or problems on your skin that you believe are related to the condition you are consulting for?
A: No
Differential diagnosis: ['Localized edema', 'Possible NSTEMI / STEMI (severe)', 'GERD', 'Pulmonary embolism (severe)', 'Unstable angina (severe)', 'Panic attack', 'Pericarditis', 'Spontaneous pneumothorax (severe)', 'Boerhaave (severe)', 'Cluster headache', 'Guillain-Barre syndrome (severe)', 'Myocarditis (severe)', 'Acute pulmonary edema (severe)', 'Anaphylaxis (severe)', 'Atrial fibrillation', 'Stable angina (severe)', 'Anemia', 'Scombroid food poisoning (severe)', 'Acute dystonic reactions (severe)', 'Bronchospasm / acute asthma exacerbation', 'Sarcoidosis', 'PSVT (severe)', 'HIV (initial infection)', 'Bronchitis', 'Myasthenia gravis', 'Chagas', 'Pneumonia', 'Pancreatic neoplasm', 'Bronchiectasis', 'Larygospasm (severe)', 'SLE', 'Spontaneous rib fracture']

Turn 2:
Q: Do you have pain somewhere, related to your reason for consulting?
A: Yes
Differential diagnosis: ['Possible NSTEMI / STEMI (severe)', 'Localized edema', 'Pulmonary embolism (severe)', 'Unstable angina (severe)', 'Spontaneous pneumothorax (severe)', 'Anaphylaxis (severe)', 'Stable angina (severe)', 'Pericarditis', 'Scombroid food poisoning (severe)', 'GERD', 'Acute COPD exacerbation / infection', 'Chagas', 'Boerhaave (severe)', 'Sarcoidosis']

Turn 3:
Q: Does the pain radiate to another location?
A: nowhere
Differential diagnosis: ['Pancreatic neoplasm', 'Scombroid food poisoning (severe)', 'Chagas', 'Anaphylaxis (severe)', 'Anemia']

Turn 4:
Q: Do you have swelling in one or more areas of your body?
A: No
Differential diagnosis: ['Cluster headache', 'Anemia', 'Possible NSTEMI / STEMI (severe)', 'Pancreatic neoplasm', 'GERD', 'Pulmonary embolism (severe)', 'Chagas', 'Unstable angina (severe)', 'Stable angina (severe)', 'Scombroid food poisoning (severe)']

Turn 5:
Q: Do you feel pain somewhere?
A: {'lower chest', 'upper chest', 'breast(R)'}
Differential diagnosis: ['Bronchitis', 'Possible NSTEMI / STEMI (severe)', 'Panic attack', 'Pulmonary embolism (severe)', 'Unstable angina (severe)', 'Pericarditis', 'Pneumonia', 'Spontaneous pneumothorax (severe)', 'Acute pulmonary edema (severe)', 'Spontaneous rib fracture', 'GERD', 'Bronchiectasis', 'Boerhaave (severe)', 'Guillain-Barre syndrome (severe)', 'Atrial fibrillation', 'Myocarditis (severe)', 'Cluster headache', 'Whooping cough', 'Stable angina (severe)', 'Anaphylaxis (severe)', 'Acute COPD exacerbation / infection', 'Bronchospasm / acute asthma exacerbation', 'Acute dystonic reactions (severe)', 'Anemia', 'Scombroid food poisoning (severe)', 'PSVT (severe)', 'Sarcoidosis', 'Myasthenia gravis', 'Acute otitis media', 'HIV (initial infection)', 'Pulmonary neoplasm', 'URTI', 'Chagas']

Turn 6:
Q: Characterize your pain:
A: {'a knife stroke', 'heartbreaking', 'violent'}
Differential diagnosis: ['Unstable angina (severe)', 'Possible NSTEMI / STEMI (severe)', 'Pulmonary embolism (severe)', 'Spontaneous pneumothorax (severe)', 'Stable angina (severe)', 'Guillain-Barre syndrome (severe)', 'Pericarditis', 'GERD', 'Myocarditis (severe)', 'Acute pulmonary edema (severe)', 'Panic attack', 'Atrial fibrillation', 'Acute dystonic reactions (severe)', 'Sarcoidosis', 'Anemia', 'Myasthenia gravis', 'Boerhaave (severe)', 'Scombroid food poisoning (severe)']

Turn 7:
Q: How precisely is the pain located?
A: 5
Differential diagnosis: ['Unstable angina (severe)', 'Possible NSTEMI / STEMI (severe)', 'Spontaneous pneumothorax (severe)', 'Pulmonary embolism (severe)', 'Stable angina (severe)', 'Guillain-Barre syndrome (severe)', 'Pericarditis', 'GERD', 'Myocarditis (severe)', 'Acute pulmonary edema (severe)', 'Panic attack', 'Atrial fibrillation', 'Acute dystonic reactions (severe)', 'Sarcoidosis', 'Anemia', 'Myasthenia gravis', 'Scombroid food poisoning (severe)', 'Boerhaave (severe)', 'PSVT (severe)']

Turn 8:
Q: How intense is the pain?
A: 5
Differential diagnosis: ['Pulmonary embolism (severe)', 'Spontaneous pneumothorax (severe)', 'Possible NSTEMI / STEMI (severe)', 'Unstable angina (severe)', 'Pericarditis', 'Stable angina (severe)', 'Guillain-Barre syndrome (severe)', 'Panic attack', 'Acute pulmonary edema (severe)', 'GERD', 'Myocarditis (severe)', 'Boerhaave (severe)', 'Atrial fibrillation', 'Acute dystonic reactions (severe)', 'PSVT (severe)', 'Sarcoidosis', 'Anemia', 'Myasthenia gravis', 'Scombroid food poisoning (severe)', 'Pneumonia']

Turn 9:
Q: How fast did the pain appear?
A: 5
Differential diagnosis: ['Pulmonary embolism (severe)', 'Spontaneous pneumothorax (severe)', 'Possible NSTEMI / STEMI (severe)', 'Unstable angina (severe)', 'Stable angina (severe)', 'Pericarditis', 'Guillain-Barre syndrome (severe)', 'Panic attack', 'Acute pulmonary edema (severe)', 'GERD', 'Myocarditis (severe)', 'Atrial fibrillation', 'Boerhaave (severe)', 'Acute dystonic reactions (severe)', 'PSVT (severe)', 'Sarcoidosis', 'Anemia', 'Myasthenia gravis', 'Scombroid food poisoning (severe)']

Turn 10:
Q: Have you ever had a spontaneous pneumothorax?
A: Yes
Differential diagnosis: ['Pulmonary embolism (severe)', 'Spontaneous pneumothorax (severe)', 'Possible NSTEMI / STEMI (severe)', 'Unstable angina (severe)', 'Pericarditis', 'Stable angina (severe)', 'Guillain-Barre syndrome (severe)', 'Panic attack', 'Acute pulmonary edema (severe)', 'GERD', 'Myocarditis (severe)', 'Atrial fibrillation', 'Boerhaave (severe)', 'Acute dystonic reactions (severe)', 'PSVT (severe)', 'Sarcoidosis', 'Anemia', 'Myasthenia gravis', 'Scombroid food poisoning (severe)']

Turn 11:
Q: Have any of your family members ever had a pneumothorax?
A: Yes
Differential diagnosis: ['Spontaneous pneumothorax (severe)', 'Pulmonary embolism (severe)', 'Possible NSTEMI / STEMI (severe)', 'Unstable angina (severe)', 'Pericarditis', 'Stable angina (severe)', 'Guillain-Barre syndrome (severe)', 'Panic attack', 'GERD', 'Myocarditis (severe)', 'Atrial fibrillation', 'Boerhaave (severe)', 'Acute pulmonary edema (severe)', 'Acute dystonic reactions (severe)', 'PSVT (severe)', 'Sarcoidosis', 'Anemia', 'Myasthenia gravis', 'Scombroid food poisoning (severe)']

Turn 12:
Q: Do you have a chronic obstructive pulmonary disease (COPD)?
A: Yes
Differential diagnosis: ['Spontaneous pneumothorax (severe)', 'Pulmonary embolism (severe)', 'Possible NSTEMI / STEMI (severe)', 'Unstable angina (severe)', 'Pericarditis', 'Stable angina (severe)', 'Guillain-Barre syndrome (severe)', 'Panic attack', 'Acute pulmonary edema (severe)', 'GERD', 'Myocarditis (severe)', 'Atrial fibrillation', 'Boerhaave (severe)', 'Acute dystonic reactions (severe)', 'PSVT (severe)', 'Sarcoidosis', 'Anemia', 'Myasthenia gravis', 'Scombroid food poisoning (severe)']

Turn 13:
Q: Do you have symptoms that are increased with physical exertion but alleviated with rest?
A: No
Differential diagnosis: ['Pulmonary embolism (severe)', 'Spontaneous pneumothorax (severe)', 'Possible NSTEMI / STEMI (severe)', 'Guillain-Barre syndrome (severe)', 'GERD', 'Boerhaave (severe)', 'Pericarditis', 'Atrial fibrillation', 'Myocarditis (severe)', 'Panic attack', 'Sarcoidosis', 'Acute dystonic reactions (severe)', 'Unstable angina (severe)', 'Anemia', 'Myasthenia gravis', 'Scombroid food poisoning (severe)', 'Stable angina (severe)', 'PSVT (severe)']

Turn 14:
Q: Do you smoke cigarettes?
A: No
Differential diagnosis: ['Pulmonary embolism (severe)', 'Spontaneous pneumothorax (severe)', 'Possible NSTEMI / STEMI (severe)', 'Guillain-Barre syndrome (severe)', 'GERD', 'Boerhaave (severe)', 'Myocarditis (severe)', 'Atrial fibrillation', 'Unstable angina (severe)', 'Pericarditis', 'Sarcoidosis', 'Anemia', 'Acute dystonic reactions (severe)', 'Panic attack', 'Myasthenia gravis', 'Scombroid food poisoning (severe)', 'Stable angina (severe)', 'PSVT (severe)']

Turn 15:
Q: Do you have chest pain even at rest?
A: No
Differential diagnosis: ['Possible NSTEMI / STEMI (severe)', 'Spontaneous pneumothorax (severe)', 'Pulmonary embolism (severe)', 'GERD', 'Guillain-Barre syndrome (severe)', 'Myocarditis (severe)', 'Pulmonary neoplasm', 'Boerhaave (severe)', 'Atrial fibrillation', 'Sarcoidosis', 'Pericarditis', 'Anemia', 'Scombroid food poisoning (severe)', 'Acute dystonic reactions (severe)', 'Myasthenia gravis', 'Panic attack', 'Unstable angina (severe)', 'PSVT (severe)', 'Stable angina (severe)']

Turn 16:
Q: Are you experiencing shortness of breath or difficulty breathing in a significant way?
A: No
Differential diagnosis: ['Possible NSTEMI / STEMI (severe)', 'Pulmonary embolism (severe)', 'Spontaneous pneumothorax (severe)']

----------
Ground truth differential diagnosis: ['Possible NSTEMI / STEMI (severe): 0.395', 'Pulmonary embolism (severe): 0.381', 'Spontaneous pneumothorax (severe): 0.222']

\end{lstlisting}

\subsection*{Example 3}

\begin{lstlisting}
Sex: M, age: 49
Initial evidence: I have felt like I was suffocating for a very short time associated with inability to breathe or speak.
Differential diagnosis: ['Bronchitis', 'Possible NSTEMI / STEMI (severe)', 'Panic attack', 'Pneumonia', 'Inguinal hernia', 'GERD', 'Pericarditis', 'Pulmonary embolism (severe)', 'Acute pulmonary edema (severe)', 'Unstable angina (severe)', 'Myocarditis (severe)', 'Spontaneous pneumothorax (severe)', 'Atrial fibrillation', 'Boerhaave (severe)', 'Guillain-Barre syndrome (severe)', 'Bronchiectasis', 'Spontaneous rib fracture', 'Whooping cough', 'Anaphylaxis (severe)', 'URTI', 'Anemia', 'Acute COPD exacerbation / infection', 'Bronchospasm / acute asthma exacerbation', 'Localized edema', 'Acute dystonic reactions (severe)', 'Stable angina (severe)', 'Pancreatic neoplasm', 'Myasthenia gravis', 'PSVT (severe)', 'HIV (initial infection)', 'Scombroid food poisoning (severe)', 'Acute otitis media', 'Sarcoidosis', 'Chagas', 'Cluster headache', 'SLE', 'Pulmonary neoplasm', 'Larygospasm (severe)', 'Tuberculosis']
----------
Turn 1:
Q: Have you traveled out of the country in the last 4 weeks?
A: N
Differential diagnosis: ['Acute otitis media', 'URTI', 'Localized edema', 'Influenza', 'Anemia', 'Cluster headache', 'HIV (initial infection)', 'GERD', 'Bronchitis', 'Pancreatic neoplasm', 'Chagas', 'Boerhaave (severe)', 'Panic attack', 'Pneumonia', 'Bronchiectasis', 'SLE', 'Tuberculosis', 'Myocarditis (severe)', 'Guillain-Barre syndrome (severe)', 'Possible NSTEMI / STEMI (severe)', 'Atrial fibrillation', 'Pulmonary embolism (severe)', 'Acute pulmonary edema (severe)', 'Unstable angina (severe)']

Turn 2:
Q: Do you have pain somewhere, related to your reason for consulting?
A: No
Differential diagnosis: ['Acute dystonic reactions (severe)', 'Pulmonary embolism (severe)', 'Atrial fibrillation', 'Pulmonary neoplasm', 'Anemia', 'Anaphylaxis (severe)', 'Scombroid food poisoning (severe)', 'Bronchitis', 'PSVT (severe)', 'Myasthenia gravis', 'Possible NSTEMI / STEMI (severe)', 'Acute COPD exacerbation / infection', 'Chagas', 'Bronchospasm / acute asthma exacerbation', 'Tuberculosis', 'Guillain-Barre syndrome (severe)', 'Pneumonia', 'Acute pulmonary edema (severe)', 'Myocarditis (severe)', 'Whooping cough']

Turn 3:
Q: Do you have a hard time opening/raising one or both eyelids?
A: Yes
Differential diagnosis: ['Acute dystonic reactions (severe)', 'Pulmonary embolism (severe)', 'Myocarditis (severe)', 'Myasthenia gravis', 'Acute pulmonary edema (severe)', 'Atrial fibrillation', 'Anaphylaxis (severe)', 'Anemia', 'Pulmonary neoplasm', 'Scombroid food poisoning (severe)', 'SLE', 'Possible NSTEMI / STEMI (severe)', 'PSVT (severe)', 'Guillain-Barre syndrome (severe)', 'Chagas']

Turn 4:
Q: Do you feel that muscle spasms or soreness in your neck are keeping you from turning your head to one side?
A: No
Differential diagnosis: ['Acute dystonic reactions (severe)', 'Pulmonary neoplasm', 'Atrial fibrillation', 'Anemia', 'Scombroid food poisoning (severe)', 'Myasthenia gravis', 'Myocarditis (severe)', 'Possible NSTEMI / STEMI (severe)', 'Anaphylaxis (severe)']

Turn 5:
Q: Do you regularly take stimulant drugs?
A: No
Differential diagnosis: ['Acute dystonic reactions (severe)', 'Atrial fibrillation', 'Pulmonary neoplasm', 'Myasthenia gravis', 'Anemia', 'Scombroid food poisoning (severe)', 'Anaphylaxis (severe)', 'Possible NSTEMI / STEMI (severe)', 'Myocarditis (severe)', 'PSVT (severe)', 'Guillain-Barre syndrome (severe)', 'Chagas', 'Acute pulmonary edema (severe)']

Turn 6:
Q: Do you have difficulty articulating words/speaking?
A: No
Differential diagnosis: ['Acute dystonic reactions (severe)', 'Myasthenia gravis', 'Anemia']

Turn 7:
Q: Have you been treated in hospital recently for nausea, agitation, intoxication or aggressive behavior and received medication via an intravenous or intramuscular route?
A: Yes
Differential diagnosis: ['Acute dystonic reactions (severe)', 'Myasthenia gravis', 'Atrial fibrillation', 'Anemia', 'Pulmonary neoplasm', 'Guillain-Barre syndrome (severe)', 'Scombroid food poisoning (severe)', 'Myocarditis (severe)', 'Possible NSTEMI / STEMI (severe)']

Turn 8:
Q: Have you started or taken any antipsychotic medication within the last 7 days?
A: Yes
Differential diagnosis: ['Acute dystonic reactions (severe)', 'Atrial fibrillation', 'Anemia', 'Pulmonary neoplasm', 'Myasthenia gravis', 'Myocarditis (severe)', 'Scombroid food poisoning (severe)', 'Guillain-Barre syndrome (severe)', 'Possible NSTEMI / STEMI (severe)', 'PSVT (severe)']

Turn 9:
Q: Do you have trouble keeping your tongue in your mouth?
A: Yes
Differential diagnosis: ['Acute dystonic reactions (severe)', 'Myasthenia gravis', 'Guillain-Barre syndrome (severe)', 'Myocarditis (severe)', 'Atrial fibrillation', 'Anemia', 'Scombroid food poisoning (severe)', 'Pulmonary neoplasm', 'Pulmonary embolism (severe)', 'SLE', 'PSVT (severe)', 'Chagas', 'Possible NSTEMI / STEMI (severe)']

Turn 10:
Q: Do you have annoying muscle spasms in your face, neck or any other part of your body?
A: Yes
Differential diagnosis: ['Acute dystonic reactions (severe)', 'Acute pulmonary edema (severe)', 'Pulmonary embolism (severe)', 'Atrial fibrillation', 'Anaphylaxis (severe)', 'Myocarditis (severe)', 'Bronchospasm / acute asthma exacerbation', 'Guillain-Barre syndrome (severe)', 'Bronchitis', 'Myasthenia gravis', 'Pulmonary neoplasm', 'Anemia', 'Scombroid food poisoning (severe)', 'Acute COPD exacerbation / infection', 'SLE', 'Pneumonia', 'Possible NSTEMI / STEMI (severe)', 'PSVT (severe)', 'Bronchiectasis', 'Tuberculosis', 'Panic attack', 'Influenza']

Turn 11:
Q: Are you experiencing shortness of breath or difficulty breathing in a significant way?
A: No
Differential diagnosis: ['Acute dystonic reactions (severe)', 'Myasthenia gravis']

Turn 12:
Q: Do you suddenly have difficulty or an inability to open your mouth or have jaw pain when opening it?
A: No
Differential diagnosis: ['Acute dystonic reactions (severe)', 'Myasthenia gravis']

Turn 13:
Q: Are you unable to control the direction of your eyes?
A: No
Differential diagnosis: ['Acute dystonic reactions (severe)', 'Myasthenia gravis']

----------
Ground truth differential diagnosis: ['Acute dystonic reactions (severe): 0.627', 'Myasthenia gravis: 0.373']


\end{lstlisting}

\end{document}